\documentclass[11pt]{article}
\usepackage{amsfonts,amsmath,amssymb,amsthm,enumerate,graphicx,color,sidecap,url}

\newtheorem{theorem}{Theorem}
\newtheorem*{theorem*}{Theorem}
\newtheorem*{prop*}{Proposition}

\newtheorem{fact}[theorem]{Fact}
\newtheorem{lemma}[theorem]{Lemma}
\newtheorem{claim}[theorem]{Claim}

\newtheorem{definition}{Definition}
\newtheorem{corollary}[theorem]{Corollary}

\newcommand{\ignore}[1]{}

\newenvironment{theorem1}[2]{\noindent \textbf{{{#1}~\ref{#2}.}}}{}

\newenvironment{prevproof}[2]{\noindent {\em {Proof of
{#1}~\ref{#2}:}}}{$\Box$\vskip \belowdisplayskip}

\newtheorem{proposition}       [theorem]   {Proposition}

\def\EX{\mathrm{SA}}
\def\E{\mathrm{E}}
\def\P{\mathrm{P}}
\def\bS{\mathbb{S}}

\def\cD{\mathcal{D}}
\def\cE{\mathcal{E}}

\def\cN{\mathcal{N}}

\def\cS{\mathcal{S}}
\def\cW{\mathcal{W}}

\def\01{\{0,1\}}
\def\r01{[0,1]}
\def\reals{{\bf R}}

\def\eps{\epsilon}

\def\poly{\mathrm{poly}}

\newcommand{\FFF}{\mathcal{F}}

\newcommand{\R}{\mathbb{R}}

\newtheorem{Alg}{Algorithm}
\renewcommand{\P}{\mathrm{Pr}}

\newcommand{\myalg}[3]{
\medskip
\footnotesize{
\fbox{
\parbox{5.5in}{
\begin{Alg}\label{#1}{\sc #2}\\
{\tt #3}
\end{Alg}
}}
\medskip
}}

\usepackage{fullpage}

\title{Settling the Polynomial Learnability of Mixtures of Gaussians}
\author{ Ankur Moitra \and Gregory Valiant}

\begin{document}
\maketitle
\thispagestyle{empty}

\abstract{Given data drawn from a mixture of multivariate Gaussians, a basic problem is to accurately estimate the mixture parameters. We give an algorithm for this problem that has a running time, and data requirement polynomial in the dimension and the inverse of the desired accuracy, with provably minimal assumptions on the Gaussians. As simple consequences of our learning algorithm, we can perform near-optimal clustering of the sample points and density estimation for mixtures of $k$ Gaussians, efficiently.

The building blocks of our algorithm are based on the work (Kalai \emph{et al}, STOC 2010)~\cite{2Gs} that gives an efficient algorithm for learning mixtures of two Gaussians by considering a series of projections down to one dimension, and applying the \emph{method of moments} to each univariate projection. A major technical hurdle in~\cite{2Gs} is showing that one can efficiently learn \emph{univariate} mixtures of two Gaussians.  In contrast, because pathological scenarios can arise when considering univariate projections of mixtures of more than two Gaussians, the bulk of the work in this paper concerns how to leverage an algorithm for learning univariate mixtures (of many Gaussians) to yield an efficient algorithm for learning in high dimensions. Our algorithm employs \emph{hierarchical clustering} and rescaling, together with delicate methods for backtracking and recovering from failures that can occur in our univariate algorithm.

Finally, while the running time and data requirements of our algorithm depend exponentially on the number of Gaussians in the mixture, we prove that such a dependence is necessary.
}

\newpage

\setcounter{page}{1}
\section{Introduction}
Given access to random samples generated from a mixture of (multivariate) Gaussians, the algorithmic problem of learning the parameters of the underlying distribution is of fundamental importance in physics, biology, geology, social sciences -- any area in which such finite mixture models arise~\cite{Mc,Tit}.  Starting with Dasgupta~\cite{D}, a series of work in theoretical computer science has sought to find (or disprove the existence of) an efficient algorithm for this task ~\cite{AK, DS, VW, AM, BV, BS}.  In this paper, we settle the polynomial-time learnability of mixtures of Gaussians, giving an algorithm that uses a polynomial amount of data and estimates the components at an inverse polynomial rate under provably minimal assumptions on the mixture (specifically, that the mixing weights and the statistical distance between the components are bounded away from zero).  As a corollary, our efficient learning algorithm can be employed to yield the first provably efficient algorithm for near-optimal clustering and density estimation, without \emph{any} restrictions on the Gaussian mixture.  Finally, we note that the runtime and data requirements of our algorithm are exponential in the number of Gaussian components; however, as we show in Section~\ref{sec:expDep}, this exponential dependence is necessary.  In the remainder of this section, we briefly summarize previous work on this problem, formally state our main result, and then discuss the differences between learning mixtures of $2$ Gaussians, and mixtures of many Gaussians, which motivates the high-level outline of our algorithm presented in Section~\ref{sec:highlevel}.  We first define a Gaussian Mixture Model (GMM).

Consider a set of $k$ \emph{different} multinormal distributions, with each distribution being defined by a mean $\mu_i \in \R^n,$ and covariance matrix $\Sigma_i \in \R^{n \times n}.$  Given a vector of $k$ nonnegative weights, $w$, summing to one, we define the associated Gaussian Mixture Model (GMM) to be the distribution yielded by, for each $i=1,\ldots,k$, taking a sample from $\cN(\mu_i,\Sigma_i)$ with probability $w_i$.  Letting $F_i$ denote the multinormal density function of the $i^{th}$ component, $\cN(\mu_i,\Sigma_i),$ the density function of the mixture is: $F = \sum_{i=1}^k w_i F_i.$

\subsection{A Brief History}

 The most popular solution for recovering reasonable estimates of the components of GMMs in practice is the EM algorithm given by Dempster, Laird and Rubin \cite{DLR}. This algorithm is a local-search heuristic that converges to a set of parameters that locally maximizes the probability of generated the observed samples. However, the EM algorithm is a heuristic only, and makes no guarantees about converging to an estimate that is close to the true parameters. Worse still, the EM algorithm (even for univariate mixtures of just two Gaussians) has been observed to converge very slowly (see Redner and Walker for a thorough treatment \cite{RW}).

In order to even hope for an algorithm (not necessarily even polynomial time), we would need a uniqueness property -- that two distinct mixtures of Gaussians must have different probability density functions. Teicher \cite{Teicher} demonstrated that a mixture of Gaussians can be uniquely identified (up to a relabeling components) by considering the probability density function at points sufficiently far from the centers (in the tails). However, such a result sheds little light on the {\em rate} of convergence of an estimator: If distinguishing Gaussian mixtures really required analyzing the tails of the distribution, then we would require an enormous number of data samples!

Dasgupta \cite{D} introduced theoretical computer science to the algorithmic problem of {\em provably} recovering good estimates for the parameters in polynomial time (and a polynomial number of samples). His technique is based on projecting data down to a randomly chosen low-dimensional subspace, finding an accurate clustering. Given enough accurately clustered points, the empirical means and co-variances of these points will be a good estimate for the actual parameters. Arora and Kannan \cite{AK} extended these ideas to work in the much more general setting in which the co-variances of each Gaussian component could be arbitrary, and not necessarily almost spherical as in \cite{D}. Yet both of these techniques are based on the concentration of distances (under random projections), and consequently required that the centers of the components be separated by at least $\sqrt{n}$ times the largest variance. Vempala and Wang \cite{VW} and Achlioptas and McSherry \cite{AM} introduced the use of spectral techniques, and were able to overcome this barrier (of relying on distance concentration) by choosing a subspace on which to project based on large principle components. Brubaker and Vempala \cite{BV} later gave the first affine-invariant algorithm for learning mixtures of Gaussians, and these ideas proved to be central in subsequent work \cite{2Gs}.

Yet all of these approaches for {\em provably} learning good estimates require, at the very least, that the statistical overlap (i.e. one minus the statistical distance) between each pair of components be at least smaller than some constant (in some cases, it is even required that the statistical overlap be exponentially small). Recently, Felman {\em et al}~\cite{FSO} gave a polynomial time algorithm for the related problem of density estimation (without any separation condition) for the special case of axis-aligned GMMs (GMMs where each component has principle coordinates aligned with the coordinate axes).  Also without any separation requirements, Belkin and Sinha\cite{BS} showed that one can efficiently learn GMMs in the special case that all components are identical spherical Gaussians.  Most similar to the present work is the recent work of Kalai {\em et al}~\cite{2Gs}, that gave a learning algorithm for the case of mixtures of two arbitrary Gaussians with provably minimal assumptions.

\subsection{Main Results}

In this section we state our main results.  To motivate these results, we first state three obvious lower bounds for recovering the parameters of  a GMM $F = \sum_{i=1}^k w_i F_i,$ which motivate our defintion of $\eps$\emph{-statistically learnable}. We provide a formal definition of statistical distance in Section~\ref{sec:defs}.

\begin{enumerate}
\item Permuting the order of the components does not change the resulting density, thus at best the hope is to recover the parameter \emph{set}, $\{(w_1,\mu_1,\Sigma_1),\ldots,(w_k,\mu_k,\Sigma_k)\}.$

\item We require at least $\Omega(1/\min_i(w_i))$ samples to estimate the parameters, since we require this number of samples to ensure that we have seen, with reasonable probability, any sample from each component.

\item If $F_i=F_j$, then it is impossible to accurately estimate $w_i,$ and in general we require at least $\Omega(1/D(F_i,F_j))$ samples to estimate $w_i$, where $D(F_i,F_j)$ denotes the statistical distance between the two distributions.
\end{enumerate}

\begin{definition}
We call a GMM $F = \sum_i w_i F_i$ $\epsilon$-statistically learnable if $\min_i w_i \geq \epsilon$ and $\min_{i \neq j} D(F_i, F_j) \geq \epsilon$.
\end{definition}

We now consider what it means to ``accurately recover the mixture components''.

\begin{definition}
Given two $n$-dimensional GMMs of $k$ Gaussians, $F = \sum_i w_i \cN(\mu_i, \Sigma_i)$ and $\hat{F} = \sum_i \hat{w}_i \cN(\hat{\mu}_i, \hat{\Sigma}_i)$, we call $\hat{F}$ an $\eps$-close estimate for $F$ if there is permutation function $\pi: [k] \rightarrow [k]$ such that for all $i \in [k]$

\begin{enumerate}

\item $ |w_i - \hat{w}_{\pi(i)}| \leq \eps$

\item $D(\cN(\mu_i, \Sigma_i), \cN(\hat{\mu}_{\pi(i)}, \hat{\Sigma}_{\pi(i)})) \leq \eps$

\end{enumerate}
\end{definition}

Note that the above definition of an $\eps$-close estimate is affine invariant.  This is more natural than defining a good estimate in terms of additive errors, since in general, even estimating the mean of an arbitrary Gaussian to some fixed additive precision is impossible without restrictions on the covariance, as scaling the data will scale the error linearly.  We can now state our main theorem:

\begin{theorem}~\label{thm:main}
Given any $n$ dimensional mixture of $k$ Gaussians $F$ that is $\eps$-statistically learnable, we can output an $\eps$-close estimate $\hat{F}$ and the running time and data requirements of our algorithm (for any fixed $k$) are polynomial in $n$, and $\frac{1}{\eps}$.
\end{theorem}

The guarantee in the main theorem implies that the estimated
parameters are off by an additive $O(\epsilon \sigma^2_{max})$, where
$\sigma^2_{max}$ is the largest (projected) variance of any
Gaussian in any direction.

Throughout this paper, we favor clarity of proof and exposition above optimization of runtime.  Since our main goal is show that these problems can be solved in polynomial time, we make very little effort to optimize the exponent.  Our algorithms are polynomial in the dimension, inverse of the success probability, and inverse of the target accuracy for any fixed number of Gaussians, $k$.  The dependency on $k$, however, is severe: the \emph{degree} of our polynomials are linear in $k$.  In Section~\ref{sec:expDep}, we give a natural construction of two GMMs $F,F'$ of $k$ components that are each $1/k$-statistically learnable, satisfy $D(F,F') \le e^{-k}$, but $F$ is not even a $1/4$-close estimate of $F$.  Thus we require an exponential in $k$ number of samples to even distinguish these two mixtures, demonstrating that the exponential dependency on $k$ in our learning algorithms is inevitable.

\begin{prop*}
There exists two GMMs $F,F'$ of $k$ components each that satisfies the following properties:
\begin{itemize}
\item $D(F,F') \le O(e^{-k/30}).$
\item $F,F'$ are $1/k$-statistically learnable.
\item $F$ is not a $1/4$-close estimate of $F'$.
\end{itemize}
\end{prop*}

\subsection{Applications}

We can leverage our main theorem to show that we can efficiently perform density estimation for \emph{arbitrary} GMMs.  For density estimation---as opposed to parameter recovery---we only care to recover a distribution that is similar to the GMM, without worrying about matching each component; in particular, if the true weight of one of the components is negligible, we can simply disregard that component with negligible effect on the statistical distance; if two components are nearly identical in statistical distance, we can simply regard them as being merged into one component.  For these reasons, we can perform density estimation efficiently without the restriction to $\eps$-statistically learnable distributions, that was required for Theorem~\ref{thm:main}.

\begin{corollary}~\label{cor:density}
For any $n\ge 1,$ $\eps,\delta >0,$ and any $n$-dimensional GMM $F=\sum_{i=1}^k w_i F_i,$ given access to independent samples from $F$, there is an algorithm that outputs $\hat{F}=\sum_{i=1}^k \hat{w_i}\hat{F_i}$ such that with probability at least $1-\delta$ over the randomization in the algorithm and in selecting the samples, $D(F,\hat{F}) \le \eps.$  Additionally, the runtime and number of samples is bounded by $poly(n,1/\eps,1/\delta).$
\end{corollary}

The proof of this corollary follows immediately from combining our main theorem, with the arguments in Appendix~\ref{sec:agenuni}. In fact, an almost identical approach to how we construct the {\sc General Univariate Algorithm} from the {\sc Basic Univariate Algorithm} (again in Appendix~\ref{sec:agenuni}) will work because we can run our main algorithm with many different parameter ranges so that most estimates are correct, and determine a consensus among the estimate so that we can recover a good statistical approximation to $F$ \emph{without any assumptions on the mixture} - not even $\eps$-statistical learnability.

The second corollary that we obtain from Theorem~\ref{thm:main} is for clustering. To define the problem of clustering, suppose that during the data sampling process, for each point $x_i\in \R^n,$ a hidden label $y_i \in \{1,\ldots,k\}$ called the ground truth, is generated based
upon which Gaussian was used for sampling. A clustering algorithm takes as input $m$ points and outputs a \emph{classifier} $C : \R^n \rightarrow \{1,\ldots,k\}$. The error of a classifier is the minimum, over all label permutations, of the probability that the permuted label agrees with the ground truth.  Given the mixture parameters, it is easy to see that the optimal clustering algorithm will simply assign labels based on the Gaussian component with largest posterior probability.  

\begin{corollary}~\label{cor:clustering}
For any $n\ge 1,$ $\eps,\delta >0,$ and any $n$-dimensional $\eps$-statistically learnable GMM $F=\sum_{i=1}^k w_i F_i,$ given access to independent samples from $F$, there is an algorithm that outputs a classifier $C'_F$ such that with probability at least $1-\delta$ over the randomization in the algorithm and in selecting the samples, the error of $C_F$ is at most $\eps$ larger than the error of any classifier $C'$.  Additionally, the runtime and number of samples used is bounded by $poly(n,1/\eps,1/\delta).$
\end{corollary}

The proof of this corollary follows immediately from our main theorem (yet here we need the assumption of $\eps$-statistical learnability in this case). 

\subsection{Comparing Learning Two Gaussians to Learning Many}\label{sec:obs}
This work leverages several key ideas initially presented in~\cite{2Gs} which were used to show that learning mixtures of two arbitrary Gaussians can be done efficiently.  Nevertheless, additional high-level insights, and technical details were required to extend the previous work to give an efficient learning algorithm for an arbitrary mixture of many Gaussians.  In this section we briefly summarize the algorithm for learning mixtures of two Gaussians given in~\cite{2Gs}, and then describe the hurdles to extending it to the general case.  This discussion will provide insights and motivate the high-level structure of the algorithm presented in this paper, as well as clarify which components of the proof are new, and which are straight-forward adaptations of ideas from~\cite{2Gs}.

Throughout this discussion, it will be helpful to refer to parameters $\eps_1,\eps_2,\eps_3,$ which are polynomially related to each other, and satisfy $\eps_1 << \eps_2 << \eps_3$.

There are three key components to the proof that mixtures of two Gaussians can be learned efficiently:  the 1-d Learnability Lemma, the Random Projection Lemma, and the Parameter Recovery Lemma.  The 1-d Learnability Lemma states that given a mixture of two univariate Gaussians whose two components have nonnegligible statistical distance, one can efficiently recover accurate estimates of the parameters of the mixture.  It is worth noting that in the univariate case, saying that the statistical distance between two Gaussians is non-negligible is roughly equivalent (polynomially related) to saying that the two sets of parameters are non-negligibly different, ie. the parameter distance, $|\mu-\mu'|+|\sigma^2 - \sigma'^2|,$ is non-negligible.   The Random Projection Lemma states that, given an $n$-dimensional mixture of two Gaussians which is in isotropic position and whose components have nonnegligible statistical distance, with high probability over the choice of a random unit vector $r,$ the projection of the mixture onto $r$ will yield a univariate mixture of two Gaussians that have nonnegligible statistical distance (say $\eps_3$).  The final component---the Parameter Recovery Lemma---states that, given a Gaussian $G$ in $n$ dimensions, if one has extremely accurate estimates (say to within some $\epsilon_1$) of the mean and variance of $G$ projected onto $n^2$ sufficiently distinct directions (directions that differ by at least $\epsilon_2 >> \epsilon_1$) one can accurately recover the parameters of $G$.

Given these three pieces, the high-level algorithm for learning mixtures of two Gaussians is straight-forward:
\begin{enumerate}
\item{Pick a random unit vector $r$.}
\item{Pick $n^2$ vectors $r_1,\ldots,r_{n^2},$ that are ``close'' to $r$, say $|r_i-r| \approx \epsilon_2.$}
\item{For each $i=1,\ldots,n^2,$ learn extremely accurate (to accuracy $\epsilon_1 << \epsilon_2$) univariate parameters $w_i, \mu_i,\sigma_i,\mu'_i,\sigma'_i$ for the projection of the mixture onto the vector $r_i$.}
\item{Since $|r_i-r_j| \approx \epsilon_2,$ it is not hard to show that with high probability, $|\mu_i-\mu_j| << \eps_3, |\sigma_i - \sigma_j| << \eps_3$ and by the Random Projection Lemma, $||(\mu_i,\sigma_i)- (\mu'_i,\sigma'_i)|| >> \epsilon_3$  thus it will be easy to accurately match up which parameters come from which component in the different projections, and we can apply the Parameter Recovery Lemma to each of the two components.}
\end{enumerate}

Some of the above ideas are immediately applicable to the problem of learning mixtures of many Gaussians: we can clearly use the Parameter Recovery Lemma without modification.  Additionally, we  prove a generalization of the 1-d Learnability Lemma for mixtures of arbitrary numbers of Gaussians, provided each component has non-negligible statistical distance (which, while technically tedious, employs the key idea from~\cite{2Gs} of ``deconvolving'' by a suitably chosen Gaussian---see Appendix~\ref{appendix:robustId}).  Given this extension, if we were given a mixture of $k$ Gaussians in isotropic position, and were guaranteed that the projection onto some vector $r$ resulted in a univariate mixture of Gaussians for which all pairs of components either had reasonably different means or reasonably different variances, then we could piece together the parts more-or-less as in the 2-Gaussians case.

Unfortunately, however, the Random Projection Lemma, ceases to hold in the general setting. There exist mixtures of just three Gaussians with significant pairwise statistical distances, that are in isotropic position, but have the property that with extremely high probability over choices of random unit vector $r$, the projection of the mixture onto $r$ yields a distribution that is extremely close to a univariate mixture of \emph{two} Gaussians.  This observation would foil the approach employed in the case of just two Gaussians!  Another difficulty is that if we take $n^2$ slightly different projections of our mixture of $k$ Gaussians, then it is possible that in some of the projections we see what looks like a mixture of $k'<k$ univariate Gaussians, and in some other projections we see what looks like a mixture of $k''$ univariate Gaussians. How do we match up estimates from projections onto different directions when the number of Gaussians in the estimate can differ? Or what if each projection results in an estimate that is a mixture of $k' < k$ Gaussians. Then how can we recover an $n$-dimensional estimate that is a mixture of $k$ Gaussians?

\section{Outline and Definitions}~\label{sec:highlevel}
We now discuss the high-level structure of our learning algorithm, building from the intuition given in the preceding section. At the highest level, our learning algorithm has the following form:\\ Given access to samples from a mixture of $k$ Gaussians,
\begin{enumerate}
  \item{Learn the parameters of some mixture of $k' \le k$ Gaussians, where each learned Gaussian component roughly corresponds to one or more of the Gaussians in the original mixture.}
  \item{If $k'<k$, for each of the $k'$ components recovered in the previous step, examine it closely and figure out whether it corresponds to a single Gaussian component of the original mixture, or whether it is a mixture of several of the original components (in which case we will then need to learn the parameters of these sub-components).}
\end{enumerate}

To accomplish the first step, we will require accurate parameters of the projection of each of the $k'$ ``clusters'' of components, onto $n^2$ univariate projections.  To do this, we employ a \emph{robust univariate algorithm} which, given access to samples from a univariate GMM, essentially searches for some target resolution window $(w_1,w_2)$ with $w_1 << w_2,$ such that the GMM is very close ($w_1$-close) to a GMM of $k'\le k$ statistically very distinct components (each pair of components is at least $w_2$ far apart).

Given our robust univariate algorithm, we embark on a \emph{partition pursuit} where we try to find $n^2$ vectors that yield consistent and compatible univariate parameter sets--in particular, we require that each of the $n^2$ univariate projections yields parameters that satisfy three conditions: 1) they have the same number of components, 2) the recovered parameters are much more precise than the distances between the $n^2$ projections, and 3) that the distance between the components is large enough so as to ensure an accurate matching of the components in the different projections.

Finally, given the ability to accurately recover $k' \le k$ high-dimensional Gaussians, where each learned Gaussian component roughly corresponds to one or more of the Gaussians in the original mixture, we want to be able to examine each recovered component, and determine whether it corresponds to a single component of the original mixture, or a set of original components.  We first claim that, with high probability, the only way a subset of original components will end up being grouped into a single recovered component is if the covariance of the mixture of that subset of components has a very small minimum eigenvalue.  The existence of such an eigenvalue implies that we can accurately cluster the given sample points (whose covariance, recall, is roughly 1).  Thus, given a recovered set of $k' < k$ parameters, we examine one of these $k'$ components; if the minimum eigenvalue is sufficiently small, we project the set of data samples onto the corresponding eigenvector, and then partition the sample points into two clusters (provided the eigenvalue is sufficiently small, since the overall mixture is in roughly isotropic position, we cluster so as to almost exactly respect some partition of the original components).  Given the set of sample points corresponding (roughly) to the recovered component that had small eigenvalue, we simply re-scale the data so that this subsample is now in isotropic position, and recursively run the entire algorithm on this rescaled subsample of the data, which, as we argue, consists of a mixture of $k'' < k$ components of the original mixture, with high probability.  We call this clustering step \emph{hierarchical clustering.}

We give a detailed summary in Appendix~\ref{sec:aoutline} of the main elements of each of these three main components: the robust univariate algorithm, partition pursuit, and hierarchical clustering.

\subsection{Definitions}\label{sec:defs}

\begin{definition}
Given two probability distributions $f(x), g(x)$ on $\Re^n$ we can define the statistical distance between these distributions as $$D(f(x), g(x)) = \frac{1}{2} \int_{\Re^n} |f(x) - g(x)| dx$$
\end{definition}

We will also be interested in a related notion of the parameter distance between two univariate Gaussians:

\begin{definition}
Given two univariate Gaussians, $F_1 = \cN(\mu_1, \sigma_1^2), F_2 = \cN(\mu_2, \sigma_2^2)$ we define the parameter distance as $$D_p(F_1, F_2) = |\mu_1 - \mu_2| + |\sigma_1^2 - \sigma_2^2 |$$
\end{definition}

In general, the parameter distance and the statistical distance between two univariate Gaussians can be unrelated. There are pairs of univariate Gaussians with arbitrarily small parameter distance, and yet statistical distance close to $1$, and there are pairs of univariate Gaussians with arbitrarily small statistical distance, and yet arbitrarily large parameter distances. But these scenarios can only occur if the variances can be arbitrarily small or arbitrarily large. In many instances in this paper, we will have reasonable upper and lower bounds on the variances and this will allow us to move back and forth from statistical distance and parameter distance, but we will highlight when we are doing so and note why we are able to assume an upper and lower bound on variance in that particular situation.

As we noted, there are $\eps$-statistically learnable mixtures of three Gaussians that are in isotropic position, but for which with overwhelming probability over a random direction $r$, in the projection onto $r$, there will be some pair of univariate Gaussians that are arbitrarily close in parameter distance. In these cases, our univariate algorithm may not return an estimate with three components, but will return a mixture which has only two components but is still a good estimate for the parameters of the projected mixture. To formalize this notion, we introduce what we call an $\eps$-correct sub-division.

\begin{definition}
Given a GMM of $k$ Gaussians, $F = \sum_i w_i \cN(\mu_i, \sigma_i^2)$ and a GMM of $k' \le k$ Gaussians $\hat{F} = \sum_i \hat{w}_i \cN(\hat{\mu}_i, \hat{\sigma}_i^2)$, we call $\hat{F}$ an $\eps$-correct subdivision of $F$ if there is a function $\pi: [k] \rightarrow [k']$ that is onto and

\begin{enumerate}

\item $\forall_{j \in [k']} |\sum_{i | \pi(i) = j} w_i -  \hat{w}_j | \leq \eps$

\item $\forall_{i \in [k]} D_p(F_i, \hat{F}_{\pi(i)}) \leq \eps$

\end{enumerate}

When considering high-dimensional mixtures, we replace the above parameter distance by $\|\mu_i - \hat{\mu}_{\pi(i)}\| + \|\Sigma_i - \hat{\Sigma}_{\pi(i)}\|_F \leq \eps,$ where $\|_F$ denotes the Frobenius norm.

Notationally, we will write $(\hat{F}, \pi) \in \cD_{\eps}(F)$ as shorthand for the statement that $\hat{F}$ is an $\eps$-correct subdivision for $F$ and $\pi$ is the (onto) function from $k$ to $k'$ that groups $F$ into $\hat{F}$ as above.
\end{definition}

Note that this definition, unlike the definition for $\eps$-close estimate, uses parameter distance as opposed to statistical distance. This is critical because our univariate algorithm will only be able to return an estimate that is an $\eps$-correct subdivision when the notion of ``close'' is in parameter distance, and not statistical distance because in general there could be a component of the univariate mixture of arbitrarily small variance, and we will only be able to match this to an additive guarantee and this implies nothing about the statistical distance between our estimate and the actual component.

\section{A Robust Univariate Algorithm}

In this section, we give a learning algorithm for univariate mixtures of Gaussians that will be the building block for our learning algorithm in $n$-dimensions. Unlike in the case of \cite{2Gs}, our univariate algorithm will not necessarily be given a mixture of Gaussians for which all pairwise parameter distances are reasonably large. Instead, it could happen that we are given a mixture of (say) three Gaussians so that some pair has arbitrarily small parameter distance.

In the case in which we are guaranteed that all pairwise parameter distances are reasonably large, we can iterate the technical ideas in \cite{2Gs} to give an inductive proof that a simple brute force search algorithm will return good estimates. We call this algorithm the {\sc Basic Univariate Algorithm}. From this, we build a {\sc General Univariate Algorithm} that will return a good estimate regardless of the parameter distances, although in order to do so we will need to relax the notion of a good estimate to something weaker: the algorithm return an $\eps$-correct subdivision.

\subsection{Polynomially Robust Identifiability}~\label{sec:PolyRobustId}

In this section, we show that we can efficiently learn the parameters of univariate mixtures of Gaussians, provided that the components of the mixture have nonnegligible pairwise parameter distances.  We refer to this algorithm as the {\sc Basic Univariate Algorithm}.  Such an algorithm will follow easily from Theorem~\ref{thm:identifiable}---the polynomially robust identifiability of univariate mixtures.  Throughout this section we will consider two univariate mixtures of Gaussians:

$$F(x) = \sum_{i=1}^n w_i \cN(\mu_i,\sigma_i^2,x)\mbox{, and }F'(x) = \sum_{i=1}^k w_i' \cN(\mu_i',\sigma_i'^2,x).$$

\begin{definition}~\label{def:epsStandard}
We will call the pair $F, F'$ $\epsilon$-standard if $\sigma_i^2, \sigma_i'^2 \leq 1$ and if $\epsilon$ satisfies:
\begin{enumerate} \itemsep 0pt
    \item{$w_i, w'_i \in [\epsilon, 1]$}
    \item{$|\mu_i|, |\mu'_i| \leq \frac{1}{\epsilon}$}
    \item{$|\mu_i-\mu_j| + |\sigma_i^2 - \sigma_j^2| \geq  \epsilon$ and  $|\mu'_i-\mu'_j| + |\sigma_i'^2 - \sigma_j'^2| \geq  \epsilon$ for all $i \neq j$}
    \item{$ \epsilon \le \min_{\pi} \sum_i \left( |w_i-w_{\pi(i)}'|+  |\mu_i-\mu_{\pi(i)}'|+|\sigma_i^2-\sigma_{\pi(i)}'^2|\right)$, \\where the minimization is taken over all mappings $\pi: \{1,\ldots,n\} \rightarrow \{1,\ldots,k\}.$}
\end{enumerate}
\end{definition}

\begin{theorem}~\label{thm:identifiable}
There is a constant $c>0$ such that, for any $\epsilon$-standard $F, F'$ and any $\epsilon<c$,
$$\max_{i \leq 2(n+k-1)} |M_i (F) - M_i(F')| \geq \epsilon^{O(k)}$$
\end{theorem}

While the dependency on $k$ in Theorem~\ref{thm:identifiable} is very bad, as we show in Section~\ref{sec:expDep}, this exponential dependency on $k$ is necessary.  Specifically, we give a construction of two $1/k$-standard distributions whose statistical distance is $O(e^{-k}).$

Given the polynomially robust identifiability guaranteed by the above theorem, and simple concentration bounds on the $i^{th}$ sample moment, it is easy to see that a brute-force search over a set of candidate parameter sets will yield an efficient algorithm that recovers the parameters for a univariate mixtures of Gaussians whose components have pairwise parameter distance at least $\eps$: roughly, the Basic Univariate Algorithm will take a polynomial number of samples, compute the first $4k-2$ sample moments, and compare those with the first $4k-2$ moments of each of the candidate parameter sets. The algorithm then returns the parameter set whose moments most closely match the sample moments. Theorem~\ref{thm:identifiable} guarantees that if the first $4k-2$ sample moments closely match those of the chosen parameter set, then the parameter set must be nearly accurate. To conclude the proof, we argue that a polynomial-sized set of candidate parameters suffices to guarantee that at least one set of parameters will yield moments sufficiently close to the sample moments, which, by simple concentration bounds, will be close to the true moments of the GMM.
We state the corollary below, and defer the details of the algorithm and the proof of its correctness to Appendix~\ref{appendix:basicUnivariate}.

\begin{corollary}\label{cor:basicuni}
Suppose we are given access to independent samples from a GMM $\sum_{i=1}^k w_i \cN(\mu_i,\sigma_i^2,x)$ with mean 0 and variance in the interval $[1/2,2],$ where $w_i \geq \epsilon$, and $|\mu_i - \mu_j| + |\sigma_i^2 - \sigma_j^2| \ge \epsilon$.  There exists a Basic Univariate Algorithm that, for any fixed $k$, has runtime at most $poly(\frac{1}{\epsilon},\frac{1}{\delta})$ samples and with probability at least $1 - \delta$ will output mixture parameters $\hat{w}_i,\hat{\mu}_i,\hat{\sigma_i}^2$, so that there is a permutation $\pi:[k] \rightarrow [k]$ and
$$|w_i - \hat{w}_{\pi(i)}|\leq \eps, \quad |\mu_i-\hat{\mu}_{\pi(i)}| \leq \eps, \quad |\sigma_i^2 - \hat\sigma_{\pi(i)}^2| \leq \eps \text{ for each $i=1,\ldots,k$ }.$$
\end{corollary}

\subsection{The {\sc General Univariate Algorithm}}

In this section we seek to extend the Basic Univariate Algorithm of Corollary~\ref{cor:basicuni} to the general setting of a univariate mixture of $k$ Gaussians without any requirements that the components have significant pair-wise parameter distances.  In particular, given some target accuracy $\epsilon,$ and access to independent samples from a mixture $F$ of $k$ univariate Gaussians, we want to efficiently compute a mixture $F'$ of $k' \le k$ Gaussians that is an $\epsilon$-correct subdivision of $F.$

\begin{proposition}~\label{prop:uniwolower}
There is a General Univariate Algorithm which, given $\eps,\delta>0$, and access to a GMM of $k$ Gaussians, $F = \sum_i w_i \cN(\mu_i, \sigma_i^2)$ that is in near isotropic position and satisfies $w_i \geq \eps$, will run in time polynomial in $1/\eps$ and $1/\delta,$ and will return with probability at least $1-\delta$ a GMM of $k' \le k$ Gaussians $\hat{F}$ that is an $\eps$-correct subdivision of $F$.
\end{proposition}

The critical insight in building up such a {\sc General Univariate Algorithm} is that if two components are actually close enough (in statistical distance), then the {\sc Basic Univariate Algorithm} could never tell these two components apart from a single (appropriately) chosen Gaussian, because this algorithm only requires a polynomial number of samples. So given a target precision $\eps_1$ for the {\sc Basic Univariate Algorithm}, there is some window that describes whether or not the algorithm will work correctly. If all pairwise parameter distances are either sufficiently large or sufficiently small, then the {\sc Basic Univariate Algorithm} will function as if it were given sample access to a mixture that actually does meet the requirements of the algorithm. So as long as no parameter distance falls inside a particular window (which characterizes whether or not the algorithm will behave properly), the algorithm will return a correct computation.

However, when there is some parameter distance that falls inside the {\sc Basic Univariate Algorithm}'s window, we are not guaranteed that the {\sc Basic Univariate Algorithm} will fail safely. The idea, then, is to use many disjoint windows (each of which corresponds to running the {\sc Basic Univariate Algorithm} with some target precision). If we choose enough such windows, each pairwise parameter distance can only corrupt a single run of the {\sc Basic Univariate Algorithm} so a majority of the computations will be correct. We will never know which computations resulted from cases when no parameter distance fell inside the corresponding window, but we will be able to define a notion of consensus among these different runs of the {\sc Basic Univariate Algorithm} so that a majority of the runs will agree, and any run which agrees with some computation that was correct will also be close to correct.

We defer the algorithm and proof of correctness to Appendix~\ref{sec:agenuni}

\section{Partition Pursuit}

\subsection{Outline}

In this section we demonstrate how to use the {\sc General Univariate Algorithm} to obtain good additive approximations in $n$-dimensions. Roughly, we will project the $n$-dimensional mixture $F$ onto many close-by directions, and run the {\sc General Univariate Algorithm} on each projection. This is also how the algorithm in \cite{2Gs} is able to recover good additive estimates in $n$-dimensions. However we will have to cope with the additional complication that our univariate algorithm (the {\sc General Univariate Algorithm}) does not necessarily return an estimate that is a mixture of $k$ Gaussians.

We explain in detail how the algorithm in \cite{2Gs} is able to obtain additive approximation guarantees in $n$-dimensions, building on a univariate algorithm for learning mixtures of two Gaussians. Let $\eps_3 >> \eps_2 >> \eps_1$. Given any $\eps$-statistically learnable mixture of two Gaussians in $n$-dimensions, with high probability, for a direction $r$ chosen uniformly at random the parameter distance between the two Gaussians in $P_r[F]$ will be at least $\eps_3$. Then given such a direction $r$, we can choose $n^2$ different directions $r_{x, y}$ each of which are $\eps_2$-close to $r$ (i.e. $\|r - r_{x,y}\| \approx \eps_2$). The mean and variance of a component in $P_u[F]$ change continuously as we vary the direction $u$ from $r$ to $r_{x, y}$, and this implies that for $\eps_2 << \eps_3$, we will be able to consistently pair up estimates recovered from each projection, so that for each Gaussian we have $n^2$ different estimates in different directions of the projected mean and variance. Each of these estimates are accurate to within $\eps_1$ (i.e. this is the target precision that is given to the univariate algorithm). For any Gaussian, an estimate for the projected mean and the projected variance for a direction $r$ gives a linear constraint on the mean vector $\mu$ and the co-variance matrix $\Sigma$. As a result, if $\eps_1 << \eps_2$ then the precision is much finer than the condition number of this system of linear constraints on $\mu, \Sigma$ and this yields an accurate estimate in $n$-dimensions.

\begin{lemma}\label{lemma:solve} \cite{2Gs}
Let $\eps_2,\eps_1>0$.  Suppose $|m^0-\mu\cdot r|$,$|{m}^{ij}-\mu\cdot r^{ij}|$, $|v^0-r^T \Sigma r|$,$|v^{ij}-(r^{ij})^T\Sigma r^{ij}|$ are all at most $\eps_1$.  Then {\sc Solve} outputs $\hat\mu \in \reals^n$ and $\hat\Sigma \in \reals^{n \times n}$ such that $\|\hat\mu-\mu\|<\frac{\eps_1 \sqrt{n}}{\eps_2}$, and $\|\hat{\Sigma}-\Sigma\|_F \leq \frac{6n \eps_1}{\eps_2^2}$.  Furthermore, $\hat\Sigma \succeq 0$ and $\hat\Sigma$ is symmetric.
\end{lemma}

The algorithm to which this lemma refers is given in Appendix~\ref{sec:arec}

However, the {\sc General Univariate Algorithm} does not always return a mixture of $k$ Gaussians, and can in fact return a mixture $\hat{F}^u$ of $k' < k$ Gaussians provided that this mixture is still an $\eps_1$-correct subdivision of $P_u[F]$ (for some direction $u$). But then what happens if we consider two close-by directions, $u$ and $v$ and the number of Gaussians in the estimate $\hat{F}^u$ is different from the number of Gaussians in the estimate $\hat{F}^v$?

The key insight is that if we choose some direction $r$, and close-by directions $r_{x, y}$, if any estimate returned for $r_{x, y}$ has more components than the estimate returned for the direction $r$, then we have made progress because we have identified another Gaussian in the original mixture $F$. So here, rather than trying to use this estimate for $r_{x, y}$, we just start the algorithm over using $r_{x, y}$ as the original direction, and considering $n^2$ close-by directions.

The additional complication is that we must make sure every time we see a different number of components, that we've made progress. We can do so by maintaining a Window from $\eps_1$ to $\eps_3$, and we say that a Window is satisfied if the estimate $\hat{F}^r$ returned for some direction $r$ has all pairs of Gaussians either at parameter distance at least $\eps_3$, or at most the precision $\eps_1$ of the  {\sc General Univariate Algorithm}. Then if we consider close-by directions $r_{x, y}$ (that are $\eps_2$-close to $r$, for $\eps_1 << \eps_2 << \eps_3$), we can ensure that whenever we see a different number of components in the estimate corresponding to some direction $r_{x, y}$, there are \emph{more} components. When we see more components, we may need to shift the Window $W$ to a Window $W'$ so that in this new direction $r_{x, y}$, the Window $W'$ is satisfied. We take $r_{x, y}$ as the new base direction. But we have made progress because we have identified a new component in the mixture.

We state our main theorem in this section, and defer the algorithm and proof to Appendix~\ref{sec:apartitionpursuit}

\begin{theorem}~\label{thm:partitionpursuit}
Given an $\eps$-statistically learnable GMM $F$ in isotropic position, the {\sc Partition Pursuit Algorithm} will recover an $\eps$-correct sub-division $\hat{F}$ and if $F$ has more than one component, $\hat{F}$ also has more than one component.
\end{theorem}

\section{Clustering and Recursion}\label{sec:clustering}

\subsection{Outline}

In this section, we give an efficient algorithm for learning an estimate $\hat{F}$ that is $\eps$-close to the actual mixture $F$. {\sc Partition Pursuit} assumes that the mixture $F$ is in isotropic position, and even though $F$ is not necessarily in isotropic position, we will be able to get around this hurdle by first taking enough samples to compute a transformation that places the mixture $F$ in nearly isotropic position and then applying this transformation to each sample from the oracle. The main technical challenge in this section is actually what to do when the mixture $\hat{F}$ returned by {\sc Partition Pursuit} is a good additive approximation to $F$ (i.e. it is an $\eps_1$-correct subdivision with $\eps_1 << \eps$), but is not $\eps$-close to the mixture $F$. This can only happen if there is a component in $F$ that has a very small variance in some direction. Consider for example, two Gaussians in one dimension $\cN(0, \gamma)$ and $\cN(0, \gamma + \eps_1)$. Even if $\eps_1$ is very small, if $\gamma$ is much smaller, then the statistical distance between these two Gaussians can be arbitrarily close to $1$.

So the high-level idea is that if the estimate $\hat{F}$ returned by {\sc Partition Pursuit} is not $\eps$-close to $F$ (but $\hat{F}$ is an $\eps_1$-correct subdivision of $F$ for $\eps_1 << \eps$), then it must be the case that some component $F_i$ of $F$ has a co-variance matrix $\Sigma_i$ so that for some direction $v$, $v^T \Sigma_i v$ is very small. Then we can use this direction $v$ to still make progress: If we project the mixture $F$ onto $v$, we will be able to cluster accurately. There will be some partition of the Gaussians in $F$ into two disjoint, non-empty sets of components $S, T$ and some clustering scheme that can accurately clusters points sampled from $F$ into points that originated from a component in $S$ and points that originated from a component in $T$. So we can hope to accurately cluster enough points sampled from $F$ into sets of points that originated from $S$ and sets of points that originated from $T$, and then we can run our learning algorithm (with a smaller maximum of at most $k-1$ components) on each set of points. By induction, this learning algorithm will return close estimates, and if we take a convex combination of these estimates we obtain a new estimate $\hat{F}'$ that is $\eps$-close to $F$. The main technical challenge is in showing that if there is some component of $F$ with a small enough variance in some direction $v$, then we can accurately cluster points sampled from $F$. Given this, our main result follows almost immediately from an inductive argument.

\subsection{How to Cluster}

Here we give formalize the notion of a clustering scheme. Additionally, we state the key lemmas that will be useful in showing that if $\hat{F}$ is not an $\eps$-close estimate to $F$, then we can use $\hat{F}$ to construct a good clustering scheme that makes progress on our learning problem.

\begin{definition}
We will call $A, B \subset \Re^n$ a clustering scheme if $A \cap B = \emptyset$
\end{definition}

\begin{definition}
For $A \subset \Re^n$, we will write $P[F_i, A]$ to denote $Pr_{x \sim F_i}[x \in A]$ - i.e. the probability that a randomly chosen sample from $F_i$ is in the set $A$.
\end{definition}

If we have a direction $v$ and some component $\hat{F}_i$ which has small variance in direction $v$, we want to use this direction to cluster accurately. The intuition is clearest in the case of mixtures of two Gaussians: Suppose one of the components, say $\hat{F}_1$, had small variance on direction $v$. If the entire mixture is in isotropic position, then the variance of the mixture when projected onto direction $v$ is $1$. This can only happen if either the difference in projected means $| v^T(\hat{\mu}_1 - \hat{\mu}_2)|$ is large or the variance of $\hat{F}_2$ on direction $v$ is large. In the first case, we can choose an interval around each projected (estimate) mean $v^T \hat{\mu}_1$ and $v^T \hat{\mu}_2$ so that with high probability, any point sampled from $F_1$ is contained in the interval around $v^T \hat{\mu}_1$ and similarly for $F_2$. If, instead, the variance of $F_2$ when projected onto $v$ is large, then again a small interval around the point $v^T \hat{\mu}_1$ will contain most samples from $F_1$, but because the maximum density of $v^T F_2$ is never large and the interval around $v^T \hat{\mu}_1$ is not too large either, most samples from $F_2$ will not be contained in the interval. This idea is the basis of our clustering lemmas, although there will be additional complications when the mixture contains more than two Gaussians, the intuition is close to the same.

Let $(\hat{F}, \pi) \in \cD_{\eps_1}(F)$. Suppose also that $\hat{F}$ is a mixture of $k'$ components.

\begin{lemma}~\label{lemma:line}
Suppose that for some direction $v$, for all $i$: $v^T \hat{\Sigma}_i v \leq \eps_2$, for $\eps_1 \leq \frac{\sqrt{\eps_2}}{2 \eps_3}$. If there is some bi-partition $S \subset [k']$ s.t. $\forall_{i \in S, j \in [k'] - S} |v^T \hat{\mu}_i - v^T \hat{\mu}_j| \geq  \frac{3\sqrt{\eps_2}}{\eps_3}$ then there is a clustering scheme $(A, B)$ (based only on $\hat{F}$) so that for all $i \in S, j \in \pi^{-1}(i)$, $P[F_i, A] \geq 1 - \eps_3$ and for all $i \notin S, j \in \pi^{-1}(i)$, $Pr[F_i, B] \geq 1 - \eps_3$.
\end{lemma}

This lemma corresponds to the first case in the above thought exercise when there is some bi-partition of the components so that all pairs of projected means across the bi-partition are reasonably separated.

\begin{lemma}~\label{lemma:line2}
Suppose that for some direction $v$ and some $i \in [k']$ such that: $v^T \hat{\Sigma}_i v \leq \eps_m$, for $\eps_m >> \eps_1$. If there is some bi-partition $S \subset [k']$ s.t. $$\frac{\min_{i \in S} v^T \hat{\Sigma}_i v}{\max(\max_{j \notin S} v^T \hat{\Sigma}_j v, \eps_m)} \geq \frac{1}{\eps_t}$$ (and $\eps_t << \eps_3^3$) then there is a clustering scheme $A, B$ such that  for all $i \in S, j \in \pi^{-1}(i)$, $P[F_i, A] \geq 1 - \eps_3$ and for all $i \notin S, j \in \pi^{-1}(i)$, $Pr[F_i, B] \geq 1 - \eps_3$.
\end{lemma}

This lemma corresponds to the second case to the second case, when there is some bi-partition of the components so that one side of the bi-partition has projected variances that are much larger than the other.

The proofs of these lemmas, along with additional technical details are given in Appendix~\ref{sec:ahtc}

\subsection{Making Progress when there is a Small Variance}

We state a lemma from \cite{2Gs} which formalizes the intuition that if there is no component in $\hat{F}$ with small variance in any direction, the $\hat{F}$ is a good statistical estimate to $F$:

\begin{lemma}~\label{lemma:dens} \cite{2Gs}
Suppose $\|\hat{\mu}_i - \mu_i\| \leq \eps_1$, $\|\hat{\Sigma}_i - \Sigma_i\|_F \leq \eps_1$, and $| \hat{w}_i - w_i| \leq \eps_1$, if either $\|\Sigma^{-1}_i\|_2 \leq \frac{1}{2\eps_m}$ or $\|\hat{\Sigma}^{-1}_i\|_2 \leq \frac{1}{2\eps_m}$ then $$D(\hat{F}_i, F_i)^2 \leq \frac{2n \eps_1}{\eps_m} + \frac{\eps_1^2}{2 \eps_m}$$
\end{lemma}

We will use this lemma as a building block to prove:

\begin{theorem}~\label{thm:clustering}
The {\sc Hierarchical Clustering Algorithm} either returns an $\eps$-close statistical estimate $\hat{F}$ for $F$, or returns a clustering scheme $A, B$ such that there is some bipartition $S \subset [k]$ such that for all $i \in S, j \in \pi^{-1}(i)$, $P[F_i, A] \geq 1 - \eps_3$ and for all $i \notin S, j \in \pi^{-1}(i)$, $Pr[F_i, B] \geq 1 - \eps_3$. And also $S, [k] -S$ are both non-emtpy.
\end{theorem}

We defer the algorithm and the proof of correctness to Appendix~\ref{sec:aclustering}.

\subsection{Recursion}

\begin{lemma}~\label{lemma:projsep} [Isotropic Projection Lemma]
Given a mixture of $k$ $n$-Dimensional Gaussians $F = \sum_i w_i F_i$ that is in isotropic position and is $\epsilon$-statistically learnable, with probability $\geq 1 - \delta$ over a randomly chosen direction $u$, there is some pair of Gaussians $F_i, F_j$ s.t. $D_p(P_u[F_i], P_u[F_j]) \geq \frac{\eps^5 \delta^2}{50 n^2}$.
\end{lemma}

We defer a proof of this lemma to Appendix~\ref{sec:aprojlem}

\begin{definition}
Let $H_{a}(\eps, \delta, k), H_{i}(\eps, \delta, k)$ be the inverse of the number of samples needed by the {\sc High Dimensional Anisotropic Algorithm} and the {\sc High Dimensional Isotropic Algorithm} respectively when given target precision $\eps$ (and access to an $\eps$-statisically learnable distribution), an upper bound $k$ on the number of Gaussians, and an error parameter $\delta$.
\end{definition}

We defer the algorithms to Appendix~\ref{sec:arecursion}

\begin{theorem}
Given $k, \eps$, and a mixture of at most $k$ Gaussians $F$ that is $\eps$-statistically learnable {\sc High Dimensional Anisotropic Algorithm} returns an estimate $\hat{F}$ that is $\eps$-close to the actual mixture $F$.
\end{theorem}

\begin{proof}
We prove this theorem by induction. Let $\eps_{k -1} = H_a(\frac{\eps}{2}, \delta, k-1)$.

We assume by induction that both the {\sc High Dimensional Isotropic Algorithm} and the {\sc High Dimensional Anisotropic Algorithm} return an $\eps$-close estimate for all values of $k' \leq k -1$. We then consider both algorithms for the case of $k$:

Consider the {\sc High Dimensional Isotropic Algorithm} which is given $k, \eps$, and a mixture of at most $k$ Gaussians $F$ that is $\eps$-statistically learnable and is in isotropic position:
We first run the {\sc Hierarchical Clustering Algorithm} with parameters $\eps, \delta, \eps_3, k$ where $\eps_3 =\frac{1}{2} \eps \eps_{k-1} \delta$. If this algorithm returns an estimate $\hat{F}$, we can return this estimate and it is guaranteed to be $\eps$-close to the actual mixture.

Note that if the number of components in $F$ is $1$, then the {\sc Hierarchical Clustering Algorithm} will necessarily return an estimate $\hat{F}$, because there is no partitioning scheme that partitions $F$ into two subsets of components that are both non-empty. This establishes the base case in the inductive argument.

Otherwise the output of the {\sc Hierarchical Clustering Algorithm} is a clustering scheme $(A, B)$ with the property that there is some partition $S, T$ of the Gaussians in $F$ ($S, T \neq \emptyset$) and for all $i \in S$, $Pr_{x \sim F_i}[x_i \in A] \geq 1 - \eps_3$, and $j \in T$, $Pr_{x \sim F_i}[x_i \in B] \geq 1 - \eps_3$. Let $F_S, F_T$ be the (re-weighted) mixtures that result from placing every component in $S$ from $F$ into $F_S$, and every component in $T$ from $F$ into $F_T$. Note that $F_S, F_T$ are still $\eps$-statistically learnable, but may not be in isotropic position any longer.

So we can take $m = \frac{\delta}{\eps_3} = \frac{2}{\eps \eps_{k -1}}$ total samples $x_1, x_2, ..., x_m$ from $\EX(F)$. With probability at least $1 - \delta$:

\begin{enumerate}

\item All samples $x_1, x_2, ..., x_m$ are either in $A$ or $B$

\item The number of samples in $A$ and the number of samples in $B$ will each be at least $\frac{1}{\eps_{k -1}}$

\item All samples are clustered correctly - i.e. if $x_i \in A$, then $x_i$ was generated by some Gaussian $F_j$ with $j \in S$ and if $x_i \in B$, then $x_i$ was generated some Gaussian $F_j$ with $j \in T$.

\end{enumerate}

Let $X_S, X_T$ be the samples from $x_1, x_2, ..., x_m$ that are in $A, B$ respectively. We can then run the the {\sc High Dimensional Anisotropic Algorithm} with parameters $\frac{\eps}{2}, \delta, k-1$ on each set $X_S$ and $X_T$. Let the algorithm return the mixtures $\hat{F}_A, \hat{F}_B$ respectively. We return a convex combination of these mixtures, $\hat{F} = \frac{|X_S|}{m} \hat{F}_A + \frac{|X_T|}{m} \hat{F}_B$. The estimates $\hat{F}_A, \hat{F}_B$ are $\eps$-close estimates to $F_S, F_T$ respectively. We can write $F = w_A F_A + w_B F_B$, and with high probability $\frac{|X_S|}{m}$, $\frac{|X_T|}{m}$ will be close to $w_A, w_B$ respectively. Then this implies that $\hat{F}$ is $\eps$-close to $F$. Thus by induction, the output of the {\sc High Dimensional Isotropic Algorithm} is an estimate $\hat{F}$ that is $\eps$-close to $F$.

We need to also verify by induction that the output of the {\sc High Dimensional Anisotropic Algorithm} is also an $\eps$-close estimate to $F$. So suppose that the input to the  {\sc High Dimensional Anisotropic Algorithm} is a mixture of at most $k$ Gaussians, that is $\eps$-statistically learnable and is not necessarily in isotropic position.

We let $\eps_{k } = \delta H_a(\frac{\eps}{2}, \delta, k)$. Then if we take $m = O(\frac{n^4 \ln \frac{k}{\delta}}{\eps_{k}^{3}})$ samples $x_1, x_2, ..., x_m$, compute the transformation $\hat{T}$ that places these samples in exactly isotropic position, and run the {\sc High Dimensional Isotropic Algorithm} with the sample oracle $\hat{T}(\EX(F))$, parameters $\frac{\eps}{2}, \delta, k$. Using the above section, and the induction hypothesis, {\sc High Dimensional Isotropic Algorithm} outputs an $\eps$-close estimate for all values of $k' \leq k$. The input sample oracle $\hat{T}(\EX(F))$ is not exactly in isotropic position, but there is another mixture $F'$ which is in exactly isotropic position, that is $\frac{\eps}{2}$-close to $F$ and for which $D(F, F') \leq \eps_{k }$ using Theorem~\ref{thm:neariso}. Since the  {\sc High Dimensional Isotropic Algorithm} will only take $H_a(\frac{\eps}{2}, \delta, k-1)$ samples, with probability at least $1 - \delta$ we can assume that all these samples come from $F'$, which implies (by induction) that the output will be an estimate $\hat{F}$ that is $\frac{\eps}{2}$-close to $F'$, which means that $\hat{F}$ is also $\eps$-close to $F$, as desired.
\end{proof}

\section{Exponential Dependence on $k$ is Inevitable}\label{sec:expDep}

In this section, we present a lower bound, showing that the inverse exponential dependency on the number of Gaussian components in each mixture is necessary, even for mixtures in just one dimension.  We show this by giving a simple construction of two 1-dimensional distributions, $D_1,D_2$ that are $1/(2m)$-standard.  Specifically, they are mixtures of at most $m$ Gaussians, such that the weights of all components of each mixture are at least $1/(2m)$, and the parameter distance between the pair of distributions is at least $1/(2m),$ but $||D_1-D_2||_1 \le e^{-m/30},$ for sufficiently large $m$.  The construction hinges on the inverse exponential (in $k \approx \sqrt{m}$) statistical distance between $\cN(0,2),$ and the mixtures of infinitely many Gaussians of unit variance whose components are centered at multiples of $1/k$, with the weight assigned to the component centered at $i/k$ being given by $N(0,1,i/k).$  Verifying that this is true is a straight-forward exercise in Fourier analysis.  The final construction truncates the mixture of infinitely many Gaussians by removing all the components with centers a distance greater than $k$ from 0.  This truncation clearly has negligibly small effect on the distribution.  Finally, we alter the pair of distributions by adding to both distributions, Gaussian components of equal weight with centers at $-k^2/k, (-k^2+1)/k,\ldots, k,$ which ensures that in the final pair of distributions, all components have significant weight.

\begin{proposition}~\label{prop:LB}
There exists a pair $D_1,D_2$ of $1/(4k^2+2)$-standard distributions that are each mixtures of $k^2+1$ Gaussians such that $$||D_1-D_2||_1 \le 11 k e^{-k^2/24}.$$
\end{proposition}

\begin{figure*}~\label{fig:fkn}
\begin{center}
\begin{tabular}{cc}
\includegraphics[scale = 0.45, angle = 0]{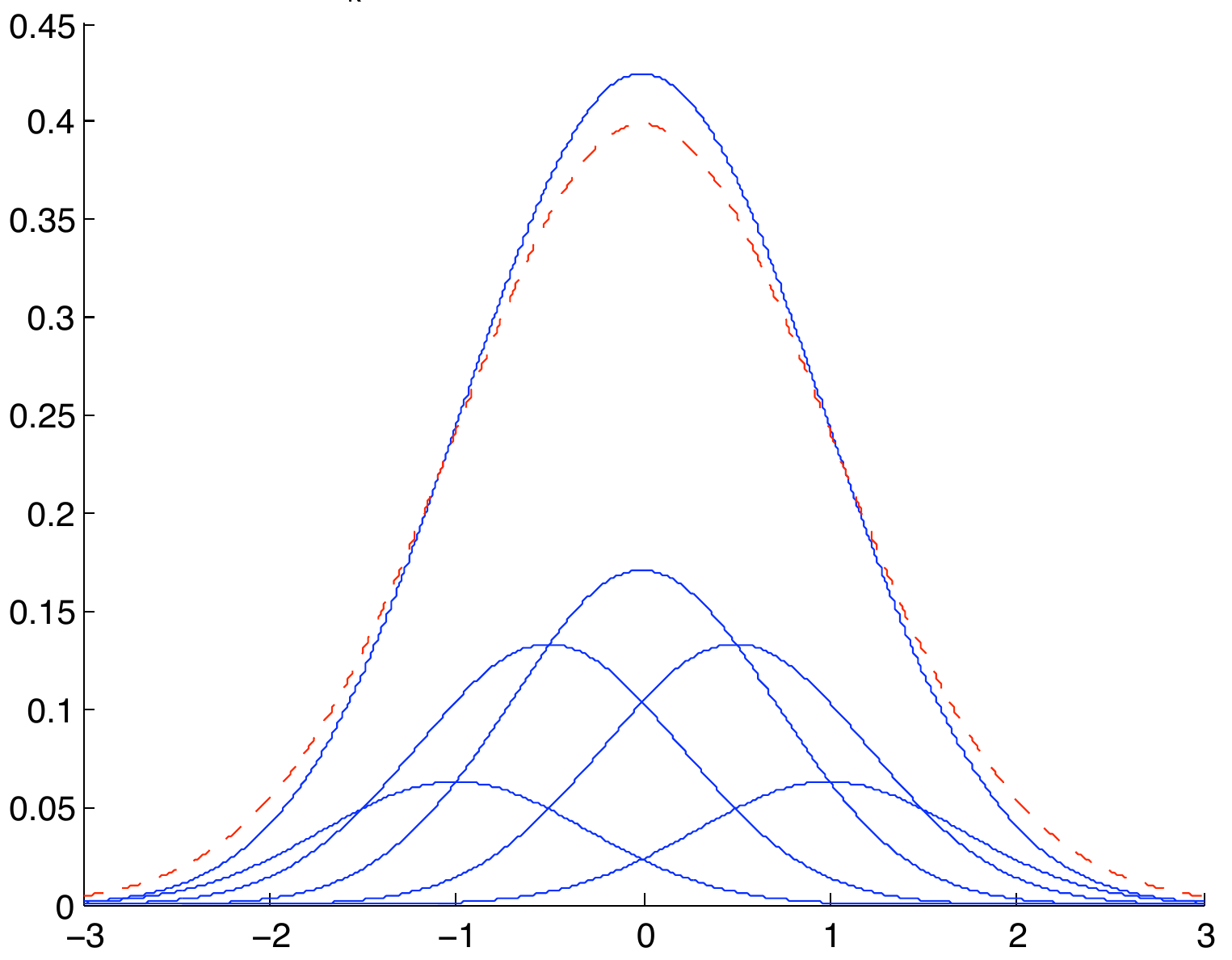} & \includegraphics[scale = 0.45, angle = 0]{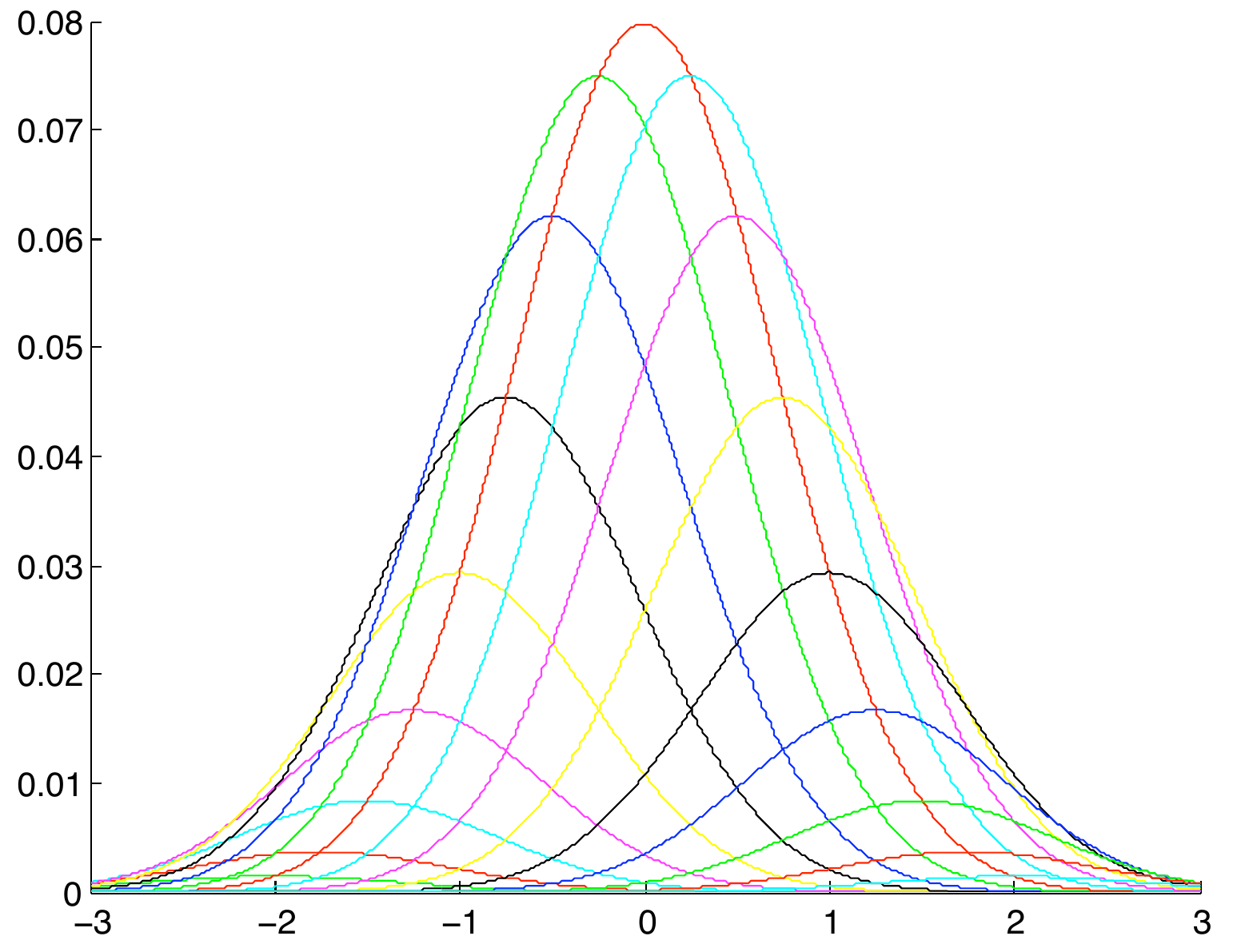}
\end{tabular}
\end{center}
\caption{a.  $F_k^N$ for $k = 2, N = 2$  (solid) and $\cN(0, 1)$ (dashed) b. $F_k^N$ for $k = 4, N = 8$}
\end{figure*}

We can define $F_k(x)^N = c_k \sum_{i=-N}^{N} \frac{1}{\sqrt{\pi}} e^{-(i/k)^2} \cN(i/k,1/2,x),$, and we give a plot of $F_k^N$ for $k = 2, N = 2$ in Figure~\ref{fig:fkn}a and the corresponding plot of each component, and in Figure~\ref{fig:fkn}b we give a plot of each component of $F_k^N$ for $k = 4, N = 8$.

We defer the details to Appendix~\ref{sec:aexpDep}

\section{Conclusions}

We give an estimator that converges to the true distribution at an inverse polynomial rate, and this result has implications for polynomial-time clustering and density estimation.  A natural question is: ``What is the \emph{optimal} rate of convergence?''  This question is wide open, and all we can say for certain is that the rate of convergence is at worst polynomial in the dimension and the inverse of the desired accuracy, and exponential in the number of components.   We made no attempt here to optimize the constants in the exponent of the rate of convergence and even if we had, the theoretical runtime would still be extremely impractical. This, however, raises the  practically relevant question of whether aspects of our algorithm can be combined with existing heuristics that seem to perform well in most applications.  For example, the brute-force-search component of our univariate algorithm is clearly expensive; perhaps employing existing heuristics (such as the EM algorithm) for the univariate problems, in conjunction with aspects of our dimension-reduction machinery might yield improved efficiency on real-world instances.

Additionally, we note that much of the machinery we developed---from the ``deconvolution'' argument for the polynomially robust identifiability, to the partition pursuit and hierarchical clustering for the dimension reduction arguments, seem to be relatively general and robust.  We suspect that such tools could be applied to yield corresponding results for other families of distributions.


\section{Acknowledgements}

We are grateful to Paul Valiant, for suggesting the lower-bound construction of Section~\ref{sec:expDep} and many helpful discussions throughout; and are indebted to Adam Tauman Kalai for introducing us to this beautiful line of research, and for all his guidance, encouragement and deep insights about mixtures of Gaussians.

\newpage

\appendix

\section{In-Depth Outline}\label{sec:aoutline}

\subsection{A Robust Univariate Algorithm}
To start, suppose that we are given access to independent samples from a mixture of $k$ Gaussians, and given a unit vector $r$ with the following promise: for each pair $G_1,G_2$ of components, in the projection of the mixture onto $r$, either the projections of $G_1$ and $G_2$ have reasonably different parameters ($> \epsilon_2$), or their projections are so close that our algorithm could never tell them apart from a single Gaussian (parameter distance at most $\epsilon_0<< \epsilon_1,$ where $\epsilon_1 << \epsilon_2$ is the desired accuracy of the 1-d parameter learning algorithm.  In this case, our 1-d parameter recovery algorithm will perform correctly, and return some $\eps_1$-accurate parameters for a mixture of $k' \le k$ components.

Thus in general, for a given desired accuracy $\epsilon_1$, there is some critical \emph{window}, namely $[\epsilon_0, \epsilon_2],$  associated with the 1-d learning algorithm that determines if it will function correctly. In a given projection, as long as no pair of components have parameter distances that fall within this window, then any pair of Gaussians is either reasonably different in parameters, or so close in parameters that the algorithm will never be able to tell the difference.

In this way, if an algorithm designer is told the parameters of a given mixture of $k$ Gaussians, he could construct an algorithm that would have been able to find some of these parameters. The algorithm would project onto a random direction $r$, and based on the pairwise parameter distances between the univariate Gaussians, there will be some window (i.e. some choice of a target precision with which to run the algorithm), bounded below by some polynomial in the desired output accuracy, so that the algorithm would function correctly.  The problem is that while there is always some window that would work for any mixture of $k$ univariate Gaussians, we don't know what window to use, and in general if we run the algorithm on a bad window, we aren't guaranteed that the algorithm will fail in a safe way.

To get around this, we run the 1-d Learning Algorithm algorithm many times on different windows that do not intersect. Because there are only $k$ univariate Gaussians, and thus at most $k^2$ different distances between component parameters in any given projection, at most $k^2$ of these windows can be corrupted.  If we choose sufficiently many windows (but still a polynomial number), a majority of the windows will yield correct parameters. Even though we can never determine which windows were good and which were bad, we can return the parameters generated by some window in consensus with a majority, and in this way, regardless of whether the window was good or bad, it is in consensus with a good window and must also be close to the correct parameters.

It is important to stress that even after the above consensus is conducted on a given projection, we still cannot be guaranteed that our univariate algorithm returns a mixture of $k$ Gaussians.  Instead, it will return some mixture of $k' < k$ Gaussians, where an element in the mixture might correspond to (say) a pair of Gaussians in the original mixture that were too close to differentiate in the given projection.

\subsection{Partition Pursuit}
This brings us to the second obstacle outlined in Section~\ref{sec:obs}: in order to recover the $n$-dimensional parameters, we will need estimates of the parameters of the Gaussians when projected on many different directions.  But, as mentioned above, the univariate algorithm will not necessarily return a mixture of $k$ Gaussians, and even if we choose a direction $r'$ that is sufficiently close to $r$ (but still $|r-r'| >> \epsilon_1,$ the accuracy of the 1-d algorithm), it may be the case that the univariate algorithm for direction $r$ returned $k'$ Gaussians, and the univariate algorithm for direction $r$ returned $k'' \neq k'$ Gaussians. How do we pair up these estimates in a consistent manner?

The key insight is that we are actually making progress if we see more Gaussians when projecting onto a different direction.  If we choose a new direction, and we see a mixture of Gaussians with more components, we should backtrack and start over as if this was the direction we originally chose. We may have to slide the \emph{window} corresponding to our 1-d algorithm and learn at a finer precision than what we chose previously, but this finer precision will still be polynomially bounded.  Effectively, we are clustering the Gaussian components into clusters with the property that the components of each cluster are indistinguishable in each of the one-dimensional projections that we have considered.   In order to make this idea work properly, we also need to ensure that we maintain a minimum parameter distance between all Gaussians clusters that we have seen (i.e. this distance is much larger than our 1-d accuracy $\epsilon_1$), so that when we choose a new direction $r'$ sufficiently close to $r,$  Gaussian component cannot switch clusters.  Thus at each stage, each cluster of Gaussians either continues to be a cluster, or it gets partitioned into several clusters of Gaussians.

\subsection{Hierarchical Clustering}
The final obstacle outlined in Section~\ref{sec:obs} can be addressed easily via an accurate clustering of the input samples together with a $k$-Gaussian analog of the Random Projection Lemma.  Intuitively, the only way that a set of high-dimensional Gaussians with significant statistical distance, when projected onto a random vector, will appear nearly identical is if the re-weighted mixture of the Gaussians in this set is very far from isotropic position.  This motivates the hope that if we have recovered some mixture of $k'<k$ components, then it must be the case that whichever of these components contains multiple original Gaussians has covariance matrix very far from isotropic.  Thus such a component must have at least one very small eigenvalue.  Given the eigenvector corresponding to such an eigenvalue, we should be able to very accurately cluster the sample points into some partition of the original Gaussians.  This motivates the following slightly more specific version of the high-level algorithm approach:\\
Given that we have recovered parameters for a mixture of $k'<k$ components:
\begin{enumerate}
  \item{Learn the parameters of some mixture of $k' \le k$ Gaussians, where each learned Gaussian component corresponds to one or more of the Gaussians in the original mixture.}
  \item{If $k'<k$, for each of the $k'$ components recovered in the previous step:
  \begin{itemize}
    \item{If the $i^{th}$ component has covariance matrix ``not too far'' from isotropic, then conclude that it corresponds to a single Gaussian in the original mixture.}
        \item{Else:
        \begin{enumerate}
          \item there is a very small eigenvalue of the covariance matrix, so project the sample points onto the corresponding eigenvector, and accurately cluster the sample points that come from this component
          \item Given the sample points corresponding to one of the components, rescale these data points so this component (which was very far from isotropic), is now in isotropic position, and repeat the entire algorithm on this sub-mixture
        \end{enumerate}}
  \end{itemize}}
\end{enumerate}

The final observation that guarantees that our algorithm will make progress with every iteration, and thus terminate after a polynomial number of steps is the following analog of the Random Projection Lemma for the $k$-Gaussians setting.  Given a mixture of $k$ Gaussians in isotropic position, with high probability over random unit vectors $r$, there will be some pair of projected Gaussians whose parameters are reasonably different.  Thus, in every projection, we will, with high probability, see what appears to be a mixture of at least two components.

\section{Polynomially Robust Identifiability}\label{appendix:robustId}

\subsection{Outline}

We now sketch the rough outline of the proof of Theorem~\ref{thm:identifiable}.  While there are considerable technical details, the main proof ideas are identical to those used in~\cite{2Gs} to prove the analogous theorem in the case that $n=k=2.$

Our proof will be via induction on $\max(n,k)$.  We start by considering the constituent Gaussian of minimal variance in the mixtures.  Assume without loss of generality that this minimum variance component is the first component of $F,$ and denote it by $N_1$.  If there is no component of $F'$ whose mean, variance, and mixing weight very closely matches those of $N_1$, then we argue that there is a significant disparity in the low order moments of $F$ and $F'$, no matter what the other Gaussian components are.  (This argument is rather involved, and we will give the high-level sketch in the next paragraph.)  If there \emph{is} a component $N_1'$ of $F'$ whose mean, variance, and mixture weight very closely matches those of $N_1$, then we argue that we can remove $N_1$ from $F$ and $N_1'$ from $F'$ with only negligible effect on the discrepancy in the low-order moments.  More formally, let $H$ be the mixture of $n-1$ Gaussians obtained by removing $N_1$ from $F$, and rescaling the weights so as to sum to one, and define $H',$ a mixture of $k-1$ Gaussians analogously.  Then, assuming that $N$ and $N'$ are very similar, the disparity in the low-order moments of $H$ and $H'$ is almost the same as the disparity in low-order moments of $F$ and$F'$.  We can then apply the induction hypothesis to the mixtures $H$ and $H'$.

We now return to the problem of showing that if the skinniest Gaussian in $F$ cannot be paired with a component of $F'$ with similar mean, variance, and weight,   that there must be a polynomially-significant discrepancy in the low-order moments of $F$ and $F'$. This step relies on 'deconvolving' by a Gaussian with an appropriately chosen variance (this corresponds to running the heat equation in reverse for a suitable amount of time).  We define the operation of deconvolving by a Gaussian of variance $\alpha$ as $\FFF_{\alpha}$; applying this operator to a mixture of Gaussians has a particularly simple effect: subtract $\alpha$ from the variance of each Gaussian in the mixture (assuming that each constituent Gaussian has variance at least $\alpha$).  If $\alpha$ is negative, this is just convolution.

\begin{definition}~\label{def:deconv}
Let $F(x) = \sum_{i=1}^n w_i \cN(\mu_i,\sigma_i^2,x)$ be the probability density function of a mixture of Gaussian distributions, and for any $\alpha < \min_i \sigma_i^2,$ define $$\FFF_{\alpha}(F)(x) = \sum_{i=1}^n w_i \cN(\mu_i,\sigma_i^2-\alpha,x).$$
\end{definition}

The key step will be to show that if the skinniest Gaussian in either of the two mixtures cannot be paired with a nearly identical Gaussian in the other mixture, then there is some $\alpha$ for which the resulting mixtures, after applying the operation $\FFF_{\alpha}$, have large statistical distance. Intuitively, this deconvolution operation allows us to isolate Gaussians in each mixture and then we can reason about the statistical distance between the two mixtures locally, without worrying about the other Gaussians in the mixture.

Given this statistical distance between the transformed pair of mixtures, we the fact that there are relatively few zero-crossings in the difference in probability density functions of two mixtures of Gaussians (Proposition~\ref{prop:number_zeros}) to show that this statistical distance gives rise to a discrepancy in at least one of the low-order moments of the pair of transformed distributions.  To complete the argument, we then show that applying this transform to a pair of distributions, while certainly not preserving statistical distance, roughly preserves the combined disparity between the low-order moments of the pair of distributions.  The complete proof can be found in Appendix~\ref{appendix:robustId}.

\subsection{Theorem~\ref{thm:identifiable}}

In this section we give the complete proof of the polynomially robust identifiability of univariate mixtures of $k$ Gaussians (Theorem~\ref{thm:identifiable}).  For convenience, we restate the theorem and all necessary definitions.  We make frequent reference to the simple properties of Gaussians and tail bounds provided in Appendix~\ref{sec:gaussian_properties}.  Throughout this section we will consider two univariate mixtures of Gaussians:

$$F(x) = \sum_{i=1}^n w_i \cN(\mu_i,\sigma_i^2,x)\mbox{, and }F'(x) = \sum_{i=1}^k w_i' \cN(\mu_i',\sigma_i'^2,x).$$

\begin{theorem1}{Definition}{def:epsStandard}
We will call the pair $F, F'$ $\epsilon$-standard if $\sigma_i^2, \sigma_i'^2 \leq 1$ and if $\epsilon$ satisfies:
\begin{enumerate} \itemsep 0pt
    \item{$w_i, w'_i \in [\epsilon, 1]$}
    \item{$|\mu_i|, |\mu'_i| \leq \frac{1}{\epsilon}$}
    \item{$|\mu_i-\mu_j| + |\sigma_i^2 - \sigma_j^2| \geq  \epsilon$ and  $|\mu'_i-\mu'_j| + |\sigma_i'^2 - \sigma_j'^2| \geq  \epsilon$ for all $i \neq j$}
    \item{$ \epsilon \le \min_{\pi} \sum_i \left( |w_i-w_{\pi(i)}'|+  |\mu_i-\mu_{\pi(i)}'|+|\sigma_i^2-\sigma_{\pi(i)}'^2|\right)$, \\where the minimization is taken over all mappings $\pi: \{1,\ldots,n\} \rightarrow \{1,\ldots,k\}.$}
\end{enumerate}
\end{theorem1}

\begin{theorem1}{Theorem}{thm:identifiable}
There is a constant $c>0$ such that, for any $\epsilon$-standard $F, F'$ and any $\epsilon<c$,
$$\max_{i \leq 2(n+k-1)} |M_i (F) - M_i(F')| \geq \epsilon^{O(k)}$$
\end{theorem1}

The following definition of the \emph{deconvolution} operation will be central to our proof of Theorem~\ref{thm:identifiable}:

\begin{theorem1}{Definition}{def:deconv}
Let $F(x) = \sum_{i=1}^n w_i \cN(\mu_i,\sigma_i^2,x)$ be the probability density function of a mixture of Gaussian distributions, and for any $\alpha < \min_i \sigma_i^2,$ define $$\FFF_{\alpha}(F)(x) = \sum_{i=1}^n w_i \cN(\mu_i,\sigma_i^2-\alpha,x).$$
\end{theorem1}

The following lemma argues that if the skinniest Gaussian in mixture $F$ can not be matched with a sufficiently similar component in the mixture $F'$, then there is some $\alpha$, possibly negative, such that $\max_x |\FFF_{\alpha}(F)(x)-\FFF_{\alpha}(F')(x)|$ is significant.  Furthermore, every component in the transformed mixtures have variances that are not too small.

\begin{lemma}~\label{prop:L1}
  Suppose $F, F'$ are $\epsilon$-standard. Suppose without loss of generality that the Gaussian of minimal variance is $\cN(\mu_1,\sigma_1^2),$ and there is some $\gamma$ satisfying $\eps/4 > \gamma >0$ such that for all $i,$ at least one of the following holds: \begin{itemize}
    \item{$|\mu_1-\mu_i'| > \gamma^8$}
    \item{$|\sigma_1^2 - \sigma_i'^2| > \gamma^8$}
    \item{$|w_1-w_i'| > \gamma.$}
   \end{itemize}
   Then there is some $\alpha$ such that either
   \begin{itemize} \item{$\max_x(|\FFF_{\alpha}(F)(x)-\FFF_{\alpha}(F')(x)|) \ge \frac{1}{2 \gamma \sqrt{ 2 \pi}}$ and the minimum variance in any component of $\FFF_{\alpha}(F)$ or $\FFF_{\alpha}(F')$ is at least $\gamma^4,$}
   \end{itemize} or
   \begin{itemize} \item{$\max_x(|\FFF_{\alpha}(F)(x)-\FFF_{\alpha}(F')(x)|) \ge \frac{1}{2 \gamma^8 \sqrt{ 2 \pi}}$ and the minimum variance in any component of $\FFF_{\alpha}(F)$ or $\FFF_{\alpha}(F')$ is at least $\gamma^{18}.$}
   \end{itemize}
\end{lemma}
\begin{proof}
  We start by considering the case when there is no Gaussian in $F'$ that matches both the mean and variance to within $\gamma^8.$    Consider applying $\FFF_{\sigma_1^2 - \gamma^{18}}$.
  $\FFF_{\sigma_1^2-\gamma^{18}}(F)(\mu_1) \ge \epsilon \cN(0,\gamma^{18},0)= \frac{\eps}{\gamma^9 \sqrt{2\pi}}.$  Next, by Corollary~\ref{cor:diffZ}, $$\FFF_{\sigma_1^2-\gamma^{18}}(F')(\mu_1) \le \frac{2}{\gamma^8 \sqrt{2 \pi e}}.$$ and thus $$\FFF_{\sigma_1^2-\gamma^{18}}(F)(\mu_1)-\FFF_{\sigma_1^2-\gamma^{18}}(F')(\mu_1) \ge \frac{2}{\gamma^8 \sqrt{2 \pi}}.$$

  Next, consider the case where we have at least one Gaussian component of $F'$ that matches both $\mu_1$ and $\sigma_1^2$ to within $\gamma^8,$ but whose weight differs from $w_1$ by at least $\gamma.$  By the definition of $\epsilon$-standard, there can be at most one such Gaussian component, say the $i^{th}$.  If $w_1 > w_i',$ then $\FFF_{\sigma_1^2-\gamma^4}(F)(\mu_1)-\FFF_{\sigma_1^2-\gamma^4}(F')(\mu_1) \ge \frac{1}{\gamma \sqrt{2 \pi}} +  \frac{2}{\epsilon \sqrt{2 \pi e}},$ where the second term is a bound on the contribution of the other Gaussian components, using the fact that $F,F'$ are $\epsilon$-standard and Corollary~\ref{cor:diffZ}.  Since $\gamma < \eps/4,$ this quantity is at least $\frac{1}{2 \gamma \sqrt{2 \pi}}.$

  If $w_1 \le w_i',$ then consider applying $\FFF_{\sigma_1^2-\gamma^4}$ to the pair of distributions.  Using the fact that $\frac{1}{\sqrt{x+x^2}}
   \ge \frac{1-x}{\sqrt{x}},$ we have
  \begin{eqnarray*}
        \FFF_{\sigma_1^2-\gamma^4}(F')(\mu_i')-\FFF_{\sigma_1^2-\gamma^4}(F)(\mu_i') & \ge & \frac{1}{\sqrt{\gamma^4 + \gamma^8} \sqrt{2 \pi}}(w_1+\gamma) - \frac{1}{\gamma^2 \sqrt{2 \pi}} w_1 - \frac{2}{\epsilon \sqrt{2 \pi e}}\\
        & \ge & \frac{1-\gamma^4}{\gamma^2 \sqrt{2 \pi}}(w_1+\gamma) - \frac{1}{\gamma^2 \sqrt{2 \pi}} w_1 - \frac{2}{\epsilon \sqrt{2 \pi e}}\\
        & \ge & \frac{1}{\gamma \sqrt{2 \pi}} -\frac{\gamma^2}{\sqrt{2 \pi}} - \frac{2}{\epsilon \sqrt{2 \pi e}}\\
        & \ge & \frac{1}{2 \gamma \sqrt{2 \pi}}.
  \end{eqnarray*}
\end{proof}

\begin{claim}~\label{claim:l}
Let $f(x^*) \geq M$ for some $x^* \in (0, r)$ and suppose that $f(x) \geq 0$ on $(0, r)$ and $f(0) = f(r) = 0$. Suppose also that $|f'(x)| \leq m$ everywhere. Then $\int_{0}^r f(x) dx \geq \frac{M^2}{m}$
\end{claim}

\begin{proof}
Consider the continuous function $g(x)$ that is defined to be $0$ for $x\in [0,x^*-M/m] \cup [x^*+M/m,r],$ and has slope $m$ on the interval $(x^*-M/m,x^*),$ and slope $-m$ on the interval $(x^*,x^*+M/m).$  Clearly $f(x) \ge g(x)$ for $x \in (0,r),$ and thus $$\int_0^r f(x) dx \ge \int_0^r g(x) dx = M^2m.$$
\end{proof}

The above claim together with Lemma~\ref{prop:L1} yields the following

\begin{corollary}
For $\alpha,\gamma$ as defined in Lemma~\ref{prop:L1}, $$D(\FFF_{\alpha}(F)(x), \FFF_{\alpha}(F')(x)) \geq \Omega(\gamma^{18}).$$
\end{corollary}

\begin{proof}
Let $f(x) = \FFF_{\alpha}(F)(x), \FFF_{\alpha}(F')(x)$, then $f(x*) \geq M$ for $M = \Omega(\frac{1}{\gamma})$ and for some $x^*$ contained in an interval $I$ in which $f(x)$ does not change sign. Similarly, because the minimum variance in any component of $\FFF_{\alpha}(F)$ or $\FFF_{\alpha}(F')$ is at least $\gamma^{18}$, this implies that $f'(x) = O(\frac{1}{\gamma^{18}}) = m$. So we can apply Claim~\ref{claim:l} using $m, M$ and get that $\int_I f(x) \geq \Omega(\gamma^{18})$ and this implies the corollary.
\end{proof}

We now show that the $\poly(\epsilon)$ statistical distance between $\FFF_{\alpha}(F)$ and $\FFF_{\alpha}(F')$ gives rise to a $\poly(\epsilon)$ disparity in one of the first $2(k+n-1)$ raw moments of the distributions.  To accomplish this, we show that there are at most $2(k+n-1)$ zero-crossings of the difference in densities, $f=\FFF_{\alpha} (F) - \FFF_{\alpha}(F')$, using properties of the evolution of the heat equation, and construct a degree $2(k+n-1)$ polynomial $p(x)$ that always has the same sign as $f(x)$, and when integrated against $f(x)$ is at least $\poly(\epsilon)$. We construct this polynomial so that the coefficients are bounded, and this implies that there is some raw moment $i$ (at most the degree of the polynomial) for which the difference between the $i^{th}$ raw moment of $\FFF_{\alpha}(F) $ and of $\FFF_{\alpha} (F')$ is large.

We use the following proposition from~\cite{2Gs} that shows that $\FFF_{\alpha}(D)(x)-\FFF_{\alpha}(D')(x)$ has few zeros.

\begin{proposition}~\label{prop:number_zeros}[Prop. 7 from~\cite{2Gs}.]
Given $f(x) = \sum_{i=1}^m a_i \cN(\mu_i,\sigma_i^2,x),$ the linear combination of $m$ one-dimensional Gaussian probability density functions, such that $\sigma_i^2 \neq \sigma_j^2$ for $i \neq j$, assuming that not all the $a_i$'s are zero, the number of solutions to $f(x) = 0$ is at most $2(m-1)$.
\end{proposition}

\begin{lemma}~\label{lemma:momDisp}
Suppose that $D(F, F') \geq \Omega(\gamma^{18})$ and that the minimum variance in any component of $F, F'$ is at least $\gamma^{18}$ and also let $F, F'$ be mixture of $n$ and $k$ Gaussians respectively, and the mean of each component of $F$ and $F'$ is at most $\frac{1}{\gamma}$. Then there is some moment $i \in [2 ( n + k -1)]$ s.t. $|E_{F}[x^i] - E_{F'}[x^i]| \geq \Omega(\gamma^c)$ for some constant $c = c(n, k)$ that depends on $n, k$.
\end{lemma}

\begin{proof}
Using Proposition~\ref{prop:number_zeros}, there are at most $2(n + k -1)$ zero crossings of the function $f(x) = F(x) - F'(x)$. Consider the interval $I = [\frac{-2}{\gamma}, \frac{2}{\gamma}]$. Using Corollary~\ref{cor:tailbd}, the contribution to $E_{F}[x^i]$ of $\Re - I$ is at most $O(\gamma^{-i} e^{-\frac{1}{8 \gamma^2}})$, and for sufficiently small $\gamma$, this is negligible.

Because $D(F, F') \geq \Omega(\gamma^{18})$ and the fact that there are at most $2(n + k -1)$ zero crossings of the function $f(x)$, there must be some interval $J$ for which $f(x)$ does not change signs and $\int_J |f(x)|dx \geq \Omega(\frac{\gamma^{18}}{2n + 2k})$. If we choose $p(x) = \pm \Pi_{z_i} (x - z_i)$ for all zeros $z_i \in I$. We can then choose signs so that $p(x)$ matches $f(x)$ on $J \cup I = J'$. Then $\int_{J'} p(x) |f(x)| dx \geq |\int_J p(x) f(x) dx| - \int_{\Re - I} |p(x) f(x)| dx \geq |\int_J p(x) f(x) dx| - O(\gamma^{-i - 2(n + k -1)} e^{-\frac{1}{8 \gamma^2}})$ because each coefficient in $p(x)$ is bounded by $\frac{1}{\gamma^{2(n + k -1)}}$. Let $J'' \subset J$ be the interval $[a - \delta, b + \delta] \subset J = [a, b]$.

Then $|\int_{J''} p(x) f(x)dx| \geq \delta^{2(n + k -1)} |\int_{J''} f(x) dx| $ and $ |\int_{J''} f(x) dx| \geq |\int_J f(x) dx | - O(\frac{\delta^2}{\gamma^{18}})$ because the derivative of $f(x)$ is bounded by $O(\frac{1}{\gamma^{18}})$, and $f(a) = f(b) = 0$. So choosing $\delta = O(\gamma^{19})$ yields that $|\int_{J''} f(x) dx| \geq \Omega(\gamma^{18})$ (where the constant hidden in $O()$ depends on $n, k$).

So this implies that $|\int_{J''} p(x) f(x)dx| \geq \Omega(\gamma^{c(n + k - 1)})$ for some constant $c$ (that does not depend on $n, k$. Using the fact that the coefficients of $p(x)$ are bounded by $\frac{1}{\gamma^{2(n + k -1)}}$, this implies that there is some $i \in [2(n + k -1)]$ such that $|\int_{J''} x^i f(x) dx|  \geq \Omega(\gamma^{c'(n + k - 1)})$ for some constant $c'$ that does not depend on $n, k$.

Then using the bound of $O(\gamma^{-i - 2(n + k -1)} e^{-\frac{1}{8 \gamma^2}})$ for $E_{\Re - I} [p(x) f(x)]$, for sufficiently small $\gamma$ this implies that $|E_{F}[x^i] - E_{F'}[x^i]| \geq \Omega(\gamma^{c''' ( n + k -1)})$
\end{proof}

Unfortunately, the transformation $\FFF_{\alpha}$ does not preserve the statistical distance between two distributions.  However, we show that it, at least roughly, preserves (up to a polynomial) the disparity in low-order moments of the distributions.

\begin{lemma}\label{prop:momentchange}[Lemma 6 from~\cite{2Gs}.]
Suppose that each constituent Gaussian in $F$ or $F'$ has variances in the interval $[\alpha, 1]$. Then
$$\sum_{i=1}^k |M_i\left(\FFF_{\alpha}(F)\right) - M_i\left(\FFF_{\alpha}(F')\right)| \le \frac{(k+1)!}{\lfloor k/2\rfloor !} \sum_{i=1}^k |M_i(F) - M_i(F')|,$$
\end{lemma}
The proof of the above lemma follows easily from the observation that the moments of $F$ and $\FFF_\alpha(F)$ are related by a simple linear transformation, which can also be viewed as a recurrence relation for Hermite polynomials.

We now put the pieces together:

\begin{prevproof}{Theorem}{thm:identifiable}
The base case for our induction is when $n=k=1,$ and follows from the fact that given parameters $\mu,\mu',\sigma^2,\sigma'^2,$ such that $\sigma^2,\sigma'^2 \le 1,$ and $|\mu-\mu'|+|\sigma^2-\sigma'^2| \ge \epsilon,$ then one of the first two moments of $\cN(\mu,\sigma^2)$ differs from that of $\cN(\mu',\sigma'^2)$ by at least $\epsilon/2.$

For the induction step, assume that for all pairs of $\epsilon$-standard mixtures of $n,$ and $k$ Gaussians, respectively, one of the first $2(n+k-1)$ moments differ by at least $f(\epsilon,n+k)$.  Consider $\epsilon$-standard mixtures $F,F',$ mixtures of $n',k'$ Gaussians, respectively, where either $n'=n+1,$ or $k'=k+1,$ and either $n'=n$ or $k'=k.$  Assume without loss of generality that $\sigma_1^2$ is the minimal variance in the mixtures, and that it occurs in mixture $F$.

We first consider the case that there exists a component of $F'$ whose mean, variance, and weight match $\mu_1,\sigma_1^2,w_1$ to within an additive $x$, where $x$ is chosen so that each of the first $2(n+k-1)$ moments of any pair of Gaussians whose parameters are within $x$ of each other, differ by at most $f(\epsilon/2,n+k-1)/2;$ specifically, letting $q(y)$ be the polynomial (dependent on $n,k$) of Lemma~\ref{lemma:same_moments} bounding the discrepancy in the first $2(n+k-1)$ moments of Gaussians whose parameters differ by $y$, we set $x$ so that $q(x) = f(\epsilon/2,n+k-1)/2.$ Note that for fixed $n,k$, $x$ will be polynomial in $\epsilon.$  Since Lemma~\ref{lemma:same_moments} requires that $\sigma_1^2 \ge \sqrt{x},,$ if this is not the case, we convolve the pair of mixtures by $\cN(0,\eps),$ which by Lemma~\ref{prop:momentchange} changes the disparity in low-order moments by a polynomial amount, and proceed with the chosen value of $x$ and the transformed pair of GMMs.

Now, consider the mixtures $H,H',$ obtained from $F,F'$ by removing the two nearly-matching Gaussian components, and rescaling the weights so that they still sum to 1.   The pair $H,H'$ will now be mixtures of $k'-1$ and $n'-1$ components, and will still be $(\epsilon-\eps^2)$-standard, and the discrepancy in their first $2(n'+k'-1)$ moments is at most $f(\epsilon/2,n+k-1)/2$ different from the discrepancy in the pair $F,F'$.  By our induction hypothesis, there is a discrepancy in one of the first $2(n'+k'-3)$ moments of at least $f(\epsilon/2,n+k-3)$ and thus the original pair $F,F'$ will have discrepancy in moments at least half of this, which is still $poly(\epsilon),$ for any fixed $n,k$.

In the case that there is no component of $F'$ that matches $\mu_1,\sigma_1^2,w_1,$ to within the desired accuracy $x$, we can apply Lemma~\ref{prop:L1} with $\gamma=x$, and thus by Lemma~\ref{lemma:momDisp} there exists some $\alpha$ such that in the transformed mixtures $\FFF_{\alpha}(F),\FFF_{\alpha}(F'),$ there is a $poly(x)=\poly(\eps)$ disparity in the first $2(k+n-1)$ moments.  By Lemma~\ref{prop:momentchange}, this disparity in the first $2(k+n-1)$ moments is polynomially related to the disparity in these first moments of the original pair of mixtures, $F,F'$.
\end{prevproof}

\section{The Basic Univariate Algorithm}\label{appendix:basicUnivariate}

In this section we formally state the {\sc Basic Univariate Algorithm}, and prove its correctness.  In particular, we will prove the following corollary to the polynomially robust identifiability of GMMs (Theorem~\ref{thm:identifiable}).

\begin{theorem1}{Corollary}{cor:basicuni}
Suppose we are given access to independent samples from a GMM $$\sum_{i=1}^k w_i \cN(\mu_i,\sigma_i^2)$$ with mean 1 and variance in $[1/2,2],$ where $w_i \geq \epsilon$, and $|\mu_i - \mu_j| + |\sigma_i^2 - \sigma_j^2| \ge \epsilon$.  The Basic Univariate Algorithm, for any fixed $k$, has runtime at most $poly_k(\frac{1}{\epsilon},\frac{1}{\delta})$ samples and with probability at least $1 - \delta$ will output mixture parameters $\hat{w}_i,\hat{\mu}_i,\hat{\sigma_i}^2$, so that there is a permutation $\pi:[k] \rightarrow [k]$ and
$$|w_i - \hat{w}_{\pi(i)}|\leq \eps, \quad |\mu_i-\hat{\mu}_{\pi(i)}| \leq \eps, \quad |\sigma_i^2 - \hat\sigma_{\pi(i)}^2| \leq \eps \text{ for each $i=1,\ldots,k$ }.$$
\end{theorem1}

\begin{figure}
\begin{center}
\myalg{alg:basunivariate}{Basic Univariate Algorithm}{
Input: $k$, $\epsilon$, $\alpha \le \eps,$ $\delta$, sample oracle $\EX(F),$ where $F=\sum_{i}w_i \cN(\mu_i,\sigma_i^2)$ is a mixture of $k$ Gaussians, where the mixture has mean is 0 and variance at most $2$, and whose components have weights and pairwise parameter distances are at least $\epsilon.$

Output: $(\hat{w_1},\hat{\mu_1},\hat{\sigma_1}^2,\ldots,\hat{w_k},\hat{\mu_k},\hat{\sigma_k}^2),$ s.t. with probability at least $1-\delta$ over the random samples, satisfies $$ \alpha \ge \min_{\pi} \sum_{i=1}^k\left(|w_i-\hat{w_i}|+|\mu_i-\hat{\mu_i}|+|\sigma^2_i-\hat{\sigma_i}^2|\right).$$

\begin{enumerate}

\item Set $\alpha \le \eps^{O{k}+2k},$ where the $O(k)$ is $2k$ more than the exponent is from Theorem~\ref{thm:identifiable}.

\item Take $\alpha \eps^{-8k} \delta^{-2}$ samples from $\EX(F),$ and compute the first $4k-2$ sample moments, $\hat{m_1},\ldots,\hat{m_{4k-2}}.$

\item Let $\gamma = O(\alpha^{4k-1}),$ and we will iterate through the entire set of candidate parameter vectors of the form $\tilde{F} = (\tilde{w_1},\tilde{\mu_1},\tilde{\sigma_1}^2,\ldots,\tilde{w_k},\tilde{\mu_k},\tilde{\sigma_k}^2)$ satisfying:
\begin{itemize}
\item All the elements are multiples of $\gamma,$
\item $\tilde{w_i} \ge \eps/2,$ and $\sum_i \tilde{w_i} = 1$
\item each pair of components has parameter distance at least $\eps/2$.
\item $|\tilde{\mu_i}|,|\tilde{\sigma_i}^2| \le 2/\eps.$
\end{itemize}

\item \qquad Compute the first $4k-2$ moments of mixture $\tilde{F}$, $\tilde{m_1},\ldots,\tilde{m_{4k-2}}$.

\item \qquad \qquad If for all $i \in \{1,\ldots, 4k-2\},$  $|\tilde{m_i} - \hat{m_i}|\le \alpha,$  then RETURN $\tilde{F}.$
\end{enumerate}

\smallskip \noindent
}
\end{center}
\caption{\small{The Univariate Algorithm.}\label{fig:basunivariate}
}
\end{figure}

Our proof of the above Corollary will consist of three parts; first, we will show that for any $\alpha \le \eps,$ a there is some polynomial $p$ such that $p(\alpha,\eps)$ samples suffices to guarantee that with probability at least $1-\delta,$ the first $4k-2$ sample moments will all be within $\alpha$ of the corresponding true moments.  Next, we show that it suffices to perform brute-force search over a polynomially-fine mesh of parameters in order to ensure that at least  one point $(\hat{w_1}, \hat{\mu_1},\hat{\sigma_1}^2,\ldots, \hat{w_k}, \hat{\mu_k},\hat{\sigma_k}^2)$ in our parameter-mesh will have the first $4k-2$ moments that are each within $\alpha$ from the true moments.  Finally, we will use Theorem~\ref{thm:identifiable} to conclude that the recovered parameter set $(\hat{\mu_1},\hat{\sigma_1}^2,\ldots, \hat{\mu_k},\hat{\sigma_k}^2)$ must be close to the true parameter set, because the first $4k-2$ moments nearly agree.  We now formalize these pieces.

\begin{lemma}\label{lem:momentconcentrate}
Let $x_1,x_2,\ldots,x_m$ be independent draws from a univariate GMM $F$ that is in isotropic position, and each of whose components has weight at least $\eps$. With probability $\geq 1-\delta$,
$$\left|\frac{1}{m}\sum_{i=1}^m x_i^k - \E_{x \sim F}[x^k]\right| \leq  \frac{1}{m \delta^2} O(\eps^{-2k}),$$ where the hidden constant on the  big-Oh depends on $k$.
\end{lemma}
\begin{proof}
By Chebyshev's inequality, with probability at most $\delta$,
$$\left(\frac{1}{m}\sum_{i=1}^m x_i^k - a\right)^2 \leq \frac{1}{\delta^2}\E\left[\left(\frac{1}{m}\sum_{i=1}^m x_i^k - \E_{x \sim F}[x^k]\right)^2\right].$$
We now bound the right hand side.  Clearly, $\E_{x_1,\ldots,x_m}\left[\frac{1}{m}\sum_{i=1}^m x_i^k - \E_{x \sim F}[x^k] \right]=0$.  Using the fact that the variance of a sum of independent random variables is the sum of the variances,
\begin{align*}
\E\left[\left(\frac{1}{m}\sum_{i=1}^m x_i^k - a\right)^2\right] &= \frac{1}{m} \E_{x \sim F}\left[\left(x^k-\E_{x \sim F}[x^k]\right)^2\right]
\\& \leq \frac{1}{m} \E_{x \sim F}[x^{2k}].
\end{align*}
To conclude, we give a very crude upper bound on the $j^{th}$ moment of $F$; since $F$ is in isotropic position and each Gaussian component has weight at least $\epsilon,$ the mean and variance of each component has magnitude at most $1/\eps.$  Thus $\E_{x \sim F}[x^j]$ can be bounded by $(2/\eps)^j + T_j,$ where $T_j \ge \max_{\sigma^2 \le 1/\eps} \left(\int_{|x-\mu| \ge 1/\eps} \cN(\mu,\sigma^2,x) dx \right),$ which, by Corollary~\ref{cor:tailbd} is at most $O(1/\eps^j),$ from which the lemma follows.
\end{proof}

We now argue that a polynomially-fine mesh suffices to guarantee that there is some parameter set in our mesh whose first $4k-2$ moments are all close to the corresponding true moments.

\begin{lemma}~\label{lemma:polymesh}
  Given a univariate mixture $F$ of $k$ Gaussians centered at 0 with variance at most $2$, each of whose weights are at least $\eps$, such that each pair of components has parameter distance at least $\eps,$ and a target accuracy $\alpha \le \eps,$ there exists a $\gamma = poly(\alpha),$ and set of parameters $(\hat{w_1}, \hat{\mu_1},\hat{\sigma_1}^2,\ldots, \hat{w_k}, \hat{\mu_k},\hat{\sigma_k}^2)$ such that each parameter is a multiple of $\gamma,$ each is bounded by $2/\eps$, each weight is at least $\eps/2,$ each pair of components has parameter distance at least $\eps/2$, and the first $4k-2$ moments of $F$ are within $\alpha$ of the corresponding moments of $\hat{F},$ the mixture corresponding to the recovered parameters.
\end{lemma}
\begin{proof}
  Consider the parameter set obtained by rounding the true parameter set, excluding the weights, to the nearest multiple of $\gamma.$  For each weight $w_i$, we set $\hat{w_i}$ to be either the multiple of $\gamma$ just above, or just below $w_i$, ensuring that $\sum_i \hat{w_i} = 1,$ which can clearly be down.  That the rounded mixture has component weights at least $\eps/2,$ pairwise parameter distances at least $\eps/2,$ and values bounded in magnitude by $2/\eps$ is obvious.  We now analyze how much the rounding has effected the moments.

  From Claim~\ref{claim:momentcoefbd}, the $i^{th}$ moment of each component is just some polynomial in $\mu, \sigma^2,$ which is a polynomial of degree at most $i$, with coefficients bounded in magnitude by $(i+2)!$  Thus changing the mean or variance by at most $\gamma$ will change the $i^{th}$ moment by at most $$(i+2)! i\left((2/\eps+\gamma)^i-(2/\eps)^i \right) \le (i+2)! i (2/\eps)^{i} \left((1+\gamma \eps/2)^i-1 \right) \le (i+2)! i (2/\eps)^{i} (i \gamma \eps)= i^2 (i+2)! 2^i \eps^{-i+1} \gamma.$$  Thus if we used the true mixing weights, the error in each moment of the entire mixture would be at most $k$ times this.  To conclude, note that for each mixing weight $|w_j-\hat{w_j}|\le \gamma,$ and since, as noted in the proof of the previous lemma, each moment is at most $O(\eps^{-i})$ (where the hidden constant depends on $i$), thus the rounding of the weight will contribute at most an extra $O(\gamma \eps^{-i}).$  Adding these bounds together, we get that each of the first $4k-2$ moments of $\hat{F}$ can be off from the true ones by at most $k (O(\gamma \eps^{-4k+2})+ 2 (4k-2)^2 (4k)! \eps^{-4k+3} \gamma = O(\gamma \eps^{-4k+2}),$ where the hidden constant depends on $k$.  Thus letting $\gamma = c_k \alpha^{4k-1},$ where the constant $c_k$ depends on $k$ suffices to ensure that all  moments are within $\alpha$ of their true values.
\end{proof}

We now piece together the above two lemmas to prove Corollary~\ref{cor:basicuni}.

\begin{prevproof}{Corollary}{cor:basicuni}
Given a desired moment accuracy $\alpha \le \eps,$ by applying a union bound to Lemma~\ref{lem:momentconcentrate}, $O(\alpha \eps^{-8k}\delta^{-2})$ samples suffices to guarantee that with probability at least $1-\delta,$ the first $4k-2$ sample moments are within $\alpha$ from the true moments.  Thus with at least probability $1-\delta,$ by Lemma~\ref{lemma:polymesh}, our polynomial mesh of parameters suffices to recover a set of parameters $(\hat{w_1}, \hat{\mu_1},\hat{\sigma_1}^2,\ldots, \hat{w_k}, \hat{\mu_k},\hat{\sigma_k}^2)$ whose weights and pairwise parameter-distances are at least $\eps/2,$ and whose first $4k-2$ sample moments will all be within $2 \alpha$ from the sample moments, and hence within $3\alpha$ from the true moments.

To conclude, note that the pair of mixtures $F,\hat{F},$ after rescaling by at most $(\eps/2)^{1/2}$ so as to ensure each component in the mixture has variance at most 1 (which scales the $k^{th}$ moments by $(\eps/2)^{k/2}$), satisfies the first three conditions of being $\eps/2$-standard, and thus, if the first $4k-2$ moments (after rescaling) agree to within $(\eps/2)^{2k-1}\cdot (\eps/2)^{O(k)},$ Theorem~\ref{thm:identifiable} guarantees that the recovered parameters must be accurate to within $\eps$ (where the first $O(k)$ in the exponent is from Theorem~\ref{thm:identifiable}).  Thus setting $3 \alpha \le (\eps/2)^{O(k)} = poly_k(\eps)$ will ensure that with the desired high probability, the recovered parameters are $\eps/2$ accurate.
\end{prevproof}

\section{The {\sc General Univariate Algorithm}}\label{sec:agenuni}

\subsection{Composing Subdivisions}

\begin{lemma}~\label{lemma:composition}
Suppose that $F$, $G$ and $H$ are GMM of $k_1 \leq k_2 \leq k_3$ Gaussians respectively. If $(G, \pi_1) \in \cD_{\eps}(F)$ and $(H, \pi_2) \in \cD_{\eps}(G)$, then $(H, \pi_2 \circ \pi_1) \in \cD_{O(k_1) \eps}(F)$.
\end{lemma}

\begin{proof}
Note that $\pi_1: [k_1] \rightarrow [k_2]$ and $\pi_2: [k_2] \rightarrow [k_3]$. Consider $\pi_3: [k_1] \rightarrow [k_3] = \pi_2 \circ \pi_1$. This function $\pi_3$ is onto, because both $\pi_1$ and $\pi_2$ are both onto.

Also consider any $j \in \pi_3^{-1}(h)$ (for some $h \in [k_3]$). In fact, let $i \in \pi_2^{-1}(h)$ and $j \in \pi_1^{-1}(i)$. Then because parameter distance is a distance (i.e. satisfies triangle-inequality):
$$D_p(F_j, H_h) \leq D_p(F_j, G_{i}) + D_p(G_{i}, H_h) \leq 2 \eps$$
because $(G, \pi_1) \in \cD_{\eps}(F)$ and $(H, \pi_2) \in \cD_{\eps}(G)$ and $\pi_2(i) =h$ and $\pi_1(j) = i$. We write $w^F_j$ for the weight of the $j^{th}$ component of $F$ to simplify notation, and similarly for $G, H$.  Then using this notation:
$$|\sum_{j \in \pi_3^{-1}(i)} w^F_j - w^H_i| \leq \sum_{\ell \in \pi_2^{-1}(i)} | \sum_{j \in \pi_1^{-1}(\ell)} w^F_j - w^G_{\ell}| + |\sum_{j \in \pi_2^{-1}(i)} w^G_j - w^H_i| \leq k_2 \eps + \eps \leq (k_1 +1) \eps.$$
\end{proof}

\begin{fact}~\label{fact:close}
\begin{eqnarray*}
 \|\cN(\mu,\sigma^2)-\cN(\mu,\sigma^2(1+\delta))\|_1 &\leq& 10 \delta\\
  \|\cN(\mu,\sigma^2)-\cN(\mu+\sigma \delta,\sigma^2)\|_1 &\leq& 10 \delta
\end{eqnarray*}
\end{fact}

\begin{corollary}~\label{cor:close}
If $\sigma_1^2 = \Theta(1)$, then $$D_p(\cN(\mu_1, \sigma_1^2), \cN(\mu_2, \sigma_2^2)) = \Theta(D(\cN(\mu_1, \sigma_1^2), \cN(\mu_2, \sigma_2^2)))$$
\end{corollary}

\begin{claim}~\label{claim:preconv}
Convolving two Gaussians $F_1, F_2$ by the same Gaussian $\cN(\mu, \sigma^2)$ preserves the parameter distance between $F_1$ and $F_2$. Also, given an estimate $\hat{F}_i$ which is within $D$ in parameter distance from $\cN \circ F_i$, by subtracting $\mu$ from the mean of $\hat{F}_i$ and $\sigma^2$ from the variance of $\hat{F}_i$, we obtain an estimate for $F_i$ which is within $D$ in parameter distance from $F_i$.
\end{claim}

\begin{lemma}~\label{lemma:closesub}
Suppose $(\hat{F}, \pi) \in \cD_{\eps}(F)$ and that each Gaussian $F_i$ in the mixture $F$ has variance at least $\frac{1}{2}$. Then $D(F, \hat{F}) \leq  O(k') \eps$, where $k'$ is the number of components in the GMM $\hat{F}$.
\end{lemma}

\begin{proof}
Let $k$ be the number of components in $F$. Then
$$D(F, \hat{F}) \leq \frac{1}{2} \sum_{i \in [k']} \| \hat{w}_i \hat{F}_i - \sum_{j \in \pi^{-1}(i)} w_j F_j \|_1$$
And for each $i \in [k']$:
$$ \| \hat{w}_i \hat{F}_i - \sum_{j \in \pi^{-1}(i)} w_j F_j \|_1 \leq |  \hat{w}_i - \sum_{j \in \pi^{-1}(i)} w_j | + \min(\sum_{j \in \pi^{-1}(i)} w_j , \hat{w}_i) \max_{j \in \pi^{-1}(i)} \| \hat{F}_i - F_j \|_1$$
We can then apply Fact~\ref{fact:close} and the assumption that each Gaussian has variance at least $\frac{1}{2}$ (and if $\eps << 1$) implies that  $\| \hat{F}_i - F_j \|_1 = O(D_p(\hat{F}_i, F_j)) = O(\eps)$ for all $j \in \pi^{-1}(i)$. And so
$D(F, \hat{F}) \leq O(k') \eps$
\end{proof}

\subsection{Windows}

Here we define the notion of a Window. Suppose we run the {\sc Basic Univariate Algorithm} with target precision of $\eps$ (and an error parameter $\delta$). Then {\sc Basic Univariate Algorithm} uses at most some polynomial in $\frac{1}{\eps}$ and $\frac{1}{\delta}$ number of samples.

Here that we assume the {\sc Basic Univariate Algorithm} run with precision $\eps$ and an error parameter $\delta$ requires some polynomial in $\frac{1}{\eps}$ and $\frac{1}{\delta}$ samples. We in fact assume that the number of samples is at most $C_{B} (\eps \delta)^{-c_B}$ for some universal constants $C_B, c_b > 0$. Then we denote $Q(\eps, \delta)$ as $\frac{1}{C_B} (\eps \delta)^{c_B}$.

\begin{definition}
Let $Q(\eps, \delta)$ be the inverse of the number of samples needed by the {\sc Basic Univariate Algorithm} when given a target precision $\eps$ (and an error parameter $\delta$).
\end{definition}

We would like to define a Window to be the range of values from $Q(\eps, \delta)$ to $\eps$ so that if all pairs of Gaussians either have parameter distance at least $\eps$, or statistical distance at most $Q(\eps, \delta)$ then the we can just run the {\sc Basic Univariate Algorithm} and assume that the algorithm behaves as if each pair of Gaussians that is extremely close is replaced with a single (appropriately) chosen Gaussian. However, we will need some slack, and so we make the Window wider so that we can take union bounds over many different runs of the algorithm, and compose different subdivisions.

\begin{definition}
Let $R(\eps, \delta) = \frac{Q(\eps, \delta)}{C_1 k^4}$ and let $S(\eps, \delta) = \frac{R(\eps, \delta)}{C_2 k^4}$ for some sufficiently large constants $C_1, C_2$.
\end{definition}

\begin{definition}
Given a target precision $\eps$, we define the Window $W(\eps)$ at $\eps$ as the range of values $[ R(\eps, \delta), \eps]$.
\end{definition}

\begin{definition}
Given a mixture of Gaussians $F$, we will say that a Window $W(\eps)$ is good if for all $i \neq j$, $D_p(F_i, F_j) \notin W(\eps)$.
\end{definition}

We give a number of claims that will be useful in the case in which we have a good Window $W(\eps)$. So suppose that the Window $W(\eps)$ is good

\begin{claim}~\label{claim:equiv}
The set of Gaussians at parameter distance at most $R(\eps, \delta)$ is an equivalence class.
\end{claim}

\begin{proof}
Consider Gaussians $F_1, F_2$ and $F_3$ such that $F_1$ and $F_2$ are at parameter distance at most $R(\eps, \delta)$ and $F_2$ and $F_3$ are also at parameter distance at most $R(\eps, \delta)$. $D_p(F_1, F_3) \leq D(F_1, F_2) + D(F_2, F_3) \leq 2 R(\eps, \delta) << \eps$ and since there is no pair of Gaussians with parameter distance inside the Window $W(\eps)$, this implies that $D_p(F_1, F_3) \leq R(\eps, \delta)$.
\end{proof}

We will let $\cE = \{\cE_1, \cE_2, ... \cE_{k'}\}$ be the equivalence class of Gaussians at parameter distance at most $R(\eps, \delta)$. We let $\pi_{\cE}: [k] \rightarrow [k']$ be the mapping function that maps a Gaussian $F_j$ to the corresponding equivalence class $\cE_i$ (i.e. $\pi_{\cE}(j) = i$). From this equivalence class and this mapping function, we can define a natural $R(\eps, \delta)$-correct subdivision of $F$.

\begin{definition}
We define the natural $R(\eps, \delta)$-correct subdivision $\hat{F}^{\cE}$ as a mixture of $k'$ Gaussians in which $\hat{F}^{\cE}_j$ is an arbitrarily chosen representative from $\cE_{i}$ ($\pi_{\cE}(j) = i$), and $\hat{w}^{\cE}_i = \sum_{j \in \pi_{\cE}^{-1}(i)} w_j$.
\end{definition}

Clearly, $(\hat{F}^{\cE}, \pi_{\cE}) \in \cD_{R(\eps, \delta)}(F)$, and $\hat{F}^{\cE}$ actually is an $R(\eps, \delta)$-correct subdivision.

\begin{claim}~\label{claim:refinement}
Let $(\hat{F}, \pi) \in \cD_{R(\eps, \delta)}(F)$, then $\hat{F}^{\cE} \in \cD_{O(k) R(\eps, \delta)}(\hat{F} )$.
\end{claim}

\begin{proof}
Let $k', k''$ be the number of Gaussians in the GMMs  $\hat{F}^{\cE}$ and $\hat{F}$ respectively. Consider any two Gaussians $F_i, F_j$ that are not mapped to the same equivalence class - i.e. $\pi_{\cE}(i) \neq \pi_{\cE}(j)$. Since  the Window $W(\eps)$ is good, this implies that $D_p(F_i, F_j) \geq \eps$. So in order for $\hat{F}$ to be an $R(\eps, \delta)$-correct subdivision, it must be the case that $\pi(i) \neq \pi(j)$.

This means that $\pi$ as a partition is a refinement of the partition $\pi_{\cE}$. Formally, there must be some function $\pi_{int}: [k''] \rightarrow [k']$ such that $\pi_{\cE} = \pi_{int} \circ \pi$. Then it follows that $(\hat{F}^{\cE}, \pi_{\int}) \in \cD_{O(k) R(\eps, \delta)}(\hat{F} )$: Consider any $i \in [k']$. $$|\hat{w}^{\cE}_i - \sum_{j \in \pi_{int}^{-1}(i)} \hat{w}_j| \leq |\hat{w}^{\cE}_i - \sum_{h \in \pi_{\cE}^{-1}(i)} w_h| + |\sum_{h \in \pi_{\cE}^{-1}(i)} w_h - \sum_{j \in \pi_{int}^{-1}(i)} \hat{w}_j|$$
And $\sum_{h \in \pi_{\cE}^{-1}(i)} w_h = \sum_{j \in \pi_{int}^{-1}(i)} \sum_{h \in \pi^{-1}(j)} w_h$ so this implies that
$$|\hat{w}^{\cE}_i - \sum_{j \in \pi_{int}^{-1}(i)} \hat{w}_j| \leq |\hat{w}^{\cE}_i - \sum_{h \in \pi_{\cE}^{-1}(i)} w_h|  + \sum_{j \in \pi_{int}^{-1}(i)} | \hat{w}_j - \sum_{h \in \pi^{-1}(j)} w_h| \leq (k'' + 1) R(\eps, \delta) $$
And similarly for any $j \in \pi_{int}^{-1}(i)$ let $h = \pi^{-1}(j)$, $$D_p(\hat{F}^{\cE}_i, \hat{F}_j) \leq D_p(\hat{F}^{\cE}_i, F_h) + D_p(F_h, \hat{F}_i) \leq 2R(\eps, \delta)$$
where the last line follows because $\pi_{\cE}(h) = i$.
\end{proof}

\begin{lemma}~\label{lemma:good}
Suppose we are given a mixture of Gaussians $F = \sum_{i=1}^k w_i \cN(\mu_i,\sigma_i^2,x)$ that is in near isotropic position, where $w_i \geq \eps$ and the Window $W(\eps)$ is good and suppose further that $\sigma_i^2 \geq \frac{1}{2}$. Let $(\hat{F}, \pi) \in \cD_{R(\eps, \delta)}(F)$. Then with probability at least $1 - 2\delta$, the output of the {\sc Basic Univariate Algorithm} is a GMM $N$ such that $N \in \cD_{O(k) \eps}(\hat{F} )$.
\end{lemma}

\begin{proof}
Let $\cE = \{\cE_1, \cE_2, ... \cE_{k'}\}$ be the equivalence class of Gaussians at parameter distance at most $R(\eps, \delta)$ (see Claim~\ref{claim:equiv}), and let $\hat{F}^{\cE}$ and $\pi_{\cE}$ be the natural $R(\eps, \delta)$-correct subdivision for $F$ and corresponding mapping function.

Let $k''$ be the number of components in $\hat{F}$. Then we can apply Claim~\ref{claim:refinement} and this implies that $\hat{F}^{\cE}$ is an $O(k)R(\eps, \delta)$-correct subdivision for $\hat{F}$.

Using Lemma~\ref{lemma:closesub}, this implies that $D(F, \hat{F}^{\cE})+ D(\hat{F}^{\cE}, \hat{F}) \leq O(k^2) R(\eps, \delta)$. So this implies that given $\frac{1}{Q(\eps, \delta)}$ samples taken from $F$ when running the {\sc Basic Univariate Algorithm}, with probability at least $$1- \frac{O(k^2) R(\eps, \delta)}{Q(\eps, \delta)} \geq 1 - \delta$$ we can assume that all samples came from $\hat{F}^{\cE}$ (because there is an approximate between $F$ and $\hat{F}^{\cE}$ that fails with probability at $D(F, \hat{F}^{\cE})$ and with probability at least $1 - \delta$ this coupling will never fail, given the number of samples obtained from $F$).

When the {\sc Basic Univariate Algorithm} is run on $\hat{F}^{\cE}$, the constraints of the {\sc Basic Univariate Algorithm} are met because for all $i \neq j$, $D_p(\hat{F}^{\cE}_i, \hat{F}^{\cE}_j) \geq \eps$ because the Window $W(\eps)$ is good. So with probability at least $1 - \delta$, the {\sc Basic Univariate Algorithm} (when run on $\hat{F}^{\cE}$) will return an $\eps$-correct subdivision $N$ of $\hat{F}^{\cE}$ (in fact, a stronger guarantee is true because the {\sc Basic Univariate Algorithm} will actually return an estimate $N$ that has $k'$ components, which matches the number of components in $\hat{F}^{\cE}$). Then we can apply Lemma~\ref{lemma:composition}, and $N$ must then be an $O(k)\eps$-correct subdivision for $\hat{F}$.
\end{proof}

\subsection{Reaching a Consensus}

\begin{figure}
\begin{center}
\myalg{alg:genunivariate}{General Univariate Algorithm}{
Input: $\epsilon$, $k,$ sample oracle $\EX(F),$ where $F$ is a mixture of at most $k$ Gaussians, is $\eps$-statistically learnable and is in isotropic position.

Output: $\hat{F}$ which is a mixture of at most $k$ Gaussians, and is an $\eps$-correct subdivision of $F$

\begin{enumerate}

\item Set $\eps_1 = \frac{\eps}{C k^{k^2}}, \eps_2 = S(\eps_1, \delta), ... \eps_{i+1} = S(\eps_i, \delta), ... \eps_{k^2 + 1} = S(\eps_{k^2}, \delta)$

\item Let $\EX(F') = \cN(0, \frac{1}{2}) \circ \EX(F)$

\item For all $i \in [k^2]$, $\hat{F}^i \leftarrow ${\sc Basic Univariate Algorithm}$(\eps_i, \delta, \EX(F'))$

\item For all $T \subset [k^2]$

\item \qquad if $|T| > \frac{k^2}{2}$ and the $T$-sequence of estimates is an $O(k)\eps_1$ correct chain

\item \qquad \qquad Output $\hat{F}^i \circ \cN(0, - \frac{1}{2}) $, where $i = \min_{j \in T} j$

\item \qquad end

\item end

\item Output FAIL

\end{enumerate}

\smallskip \noindent

}
\end{center}
\caption{\small{General Univariate Algorithm.}\label{fig:genunivariate}
}
\end{figure}

\begin{definition}
We call a sequence of GMMs, $F^1, F^2, ... F^r$ an $\eps$-correct chain if for all $i \in [r-1]$, $F^{i+1} \in \cD_{\eps}(F^i)$
\end{definition}

\begin{theorem}~\label{thm:generaluni}
Suppose we are given a mixture of Gaussians $F = \sum_{i=1}^k w_i \cN(\mu_i,\sigma_i^2,x)$ that is in isotropic position, where $w_i \geq \eps$. Then the {\sc General Univariate Algorithm} will return a GMM of $k' \leq k$ Gaussians $\hat{F}$ such that $\hat{F}$ is an $\eps$-correct subdivision of $F$.
\end{theorem}

\begin{proof}
Given $\eps$, we first define a sequence of parameters where $$\eps_1 = \frac{\eps}{C k^{k^2}}, \eps_2 = S(\eps_1, \delta), ... \eps_{i+1} = S(\eps_i, \delta), ... \eps_{k^2 + 1} = S(\eps_{k^2}, \delta)$$

Suppose first that each Gaussian in $F$ has variance at least $\frac{1}{2}$. Then in this case, the idea is to run the {\sc Basic Univariate Algorithm} for a number of different precisions, each of which corresponds to a particular Window. We will choose parameters so that these Windows are disjoint, and then because a Window is bad iff there is some pair of Gaussians with parameter distance contained inside the Window, at most ${ k \choose 2} < \frac{k^2}{2}$ Windows can be bad. So this will guarantee that a strict majority of the computations are correct.

To formalize this, given the sequence of parameters $\eps_1, \eps_2, ... \eps_{k^2 + 1}$ we define a sequence of Windows $\cW = W(\eps_1), W(\eps_2), ... W(\eps_{k^2 + 1})$.

\begin{claim}
The sequence of Windows $\cW$ is disjoint
\end{claim}

\begin{proof}
If we consider the Window $W(\eps_i)$, the largest value contained in any Window $W(\eps_j)$ for $j > i$ is the largest value contained in the Window $W(\eps_{i + 1})$ which is $\eps_{i + 1}$. Yet $\eps_{i + 1} = S(\eps_i, \delta)$ and the lower bound for the Window $W(\eps_i)$ is $R(\eps_i, \delta)$ and $R(\eps_i, \delta) >> S(\eps_i, \delta)$. Similarly, the smallest value in $W(\eps_j)$ for $j \leq i$ is the smallest value in $W(\eps_{i })$. So this implies that for any $i$, the set of Windows $W(\eps_1), W(\eps_2), ... W(\eps_i)$ are separable from the set of Windows $W(\eps_{i + 1}), W(\eps_{i+2}), ... W(\eps_{k^2 + 1})$ and this implies the claim.
\end{proof}

Suppose running the {\sc Basic Univariate Algorithm} on Window $W(\eps_i)$ returns an estimate $\hat{F}^i$.

\begin{definition}
Given any subset of indices $T \subset [k^2 + 1]$, let $i_1 > i_2 > ...i_j > ... i_{|T|}$ be the indices in $T$ arranged in decreasing order. We can generate a sequence of estimates $\hat{F}_T^1, \hat{F}_T^2, ... \hat{F}_T^{|T|}$ in which $\hat{F}_T^j = \hat{F}^{i_j}$. Also let $prec(\hat{F}_T^j) = \eps_{i_j}$, which corresponds to the precision of the Window that returned the estimate $\hat{F}_T^j = \hat{F}^{i_j}$. We call this sequence the $T$-sequence of estimates.
\end{definition}

Note that this sequence of estimates $\hat{F}_T^1, ... \hat{F}_T^{|T|}$ is arranged in order of coarsening precision - i.e. $prec(\hat{F}_T^i) << prec(\hat{F}_T^{i+1})$.

\begin{claim}
$S(prec(\hat{F}_T^j), \delta) \geq prec(\hat{F}_T^{j-1})$
\end{claim}

\begin{proof}
Let $i_1 > i_2 > ...i_j > ... i_{|T|}$ be the indices in $T$ arranged in decreasing order. So $i_{j-1} > i_j$. Then $ S(prec(\hat{F}_T^j) , \delta) =S(\eps_{i_j}, \delta) = \eps_{i_j + 1}$. And because $i_{j -1} \geq i _j + 1$, it implies that $\eps_{i_{j-1}} \leq \eps_{i_j + 1}$, and this yields the claim.
\end{proof}

Let $G \subset [k^2 + 1]$ be the set of indices of Windows that are good - i.e. $W(\eps_i)$ is good iff $i \in G$. Then let $\hat{F}_G^1, \hat{F}_G^2, ... \hat{F}_G^{|G|}$ be the $G$-sequence of estimates. Because the sequence of Windows $\cW$ is disjoint, and each pair of Gaussians (and the corresponding parameter distance) can only make a single Window bad, the set $G$ is a strict majority - i.e. $|G| > |[k^2 + 1] - G|$.

\begin{claim}
The $G$-sequence of estimates is an $O(k)\eps_1$-correct chain, and $\hat{F}_G^1$ is an $O(k)\eps_1$-correct subdivision for $F$.
\end{claim}

\begin{proof}
Let $\eps'_1<< \eps'_2 << ... \eps'_{|G|}$ be the sequence of precisions given by $prec(\hat{F}_G^1), prec(\hat{F}_G^2), ... prec(\hat{F}^{|G|})$.

Using Lemma~\ref{lemma:good}, $\hat{F}_G^1 \in \cD_{O(k) \eps'_1 }(F)$. Because $O(k) \eps'_1 \leq O(k) S(\eps'_2, \delta) \leq R(\eps'_2, \delta)$ (using the above claim) this implies that $\hat{F}_G^1 \in \cD_{R(\eps'_2, \delta)}(F)$ and so we can apply Lemma~\ref{lemma:good} again and $\hat{F}^2 \in \cD_{O(k) \eps'_2}(\hat{F}^1)$. Continuing this argument, for all $i$, $\hat{F}^{i+1} \in \cD_{O(k) \eps'_i}(\hat{F}^i)$. Since $O(k) \eps'_i \leq O(k) \eps'_{|G|} \leq O(k) \eps_1$, the sequence $\hat{F}^1, \hat{F}^2, ... \hat{F}^{|G|}$ is an $O(k)\eps_1$-correct chain.
\end{proof}

Given a subset $G' \subset [k^2 + 1]$, we can check if the $G'$-sequence of estimates is an $O(k) \eps_1$-correct chain because this property is only a function of the estimates. Then if we consider all sets in $2^{[k^2 + 1]}$, we will find some set $G' \subset [k^2 + 1]$ that is a strict majority (i.e. $|G'| > |[k^2 + 1] - G'|$) and the $G'$-sequence of estimates is an $O(k)\eps_1$-correct chain. Because $G'$ is a strict majority, and a strict majority $G$ of the Windows are good, $G \cap G' \neq \emptyset$. Suppose that $g \in G \cap G'$, and let $j$ the value such that $g$ is the $j^{th}$ largest index in $G'$.

Given the $G'$-sequence of estimates, we can take the sequence $\cS = F, \hat{F}_{G'}^j, \hat{F}_{G'}^{j+1}, ... \hat{F}_{G'}^{|G'|}$. Since the index $g$ corresponds to a good Window ($W(\eps_g)$ is good), the computation $\hat{F}_{G'}^j$ (which corresponds to the estimate $\hat{F}^g$) is at least an $O(k)\eps_1$-correct subdivision of $F$. So the sequence $\cS$ is an $O(k)\eps_1$-correct chain. So we can apply Lemma~\ref{lemma:composition}, and this implies that $\hat{F}_{G'}^{|G'|}$ (i.e. the last estimate in the sequence $\cS$) is an $(Ck)^{k^2 + 1} \eps_1$-correct subdivision for $F$. Since $(Ck)^{k^2 + 1} \eps_1 \leq \eps$, this implies that $\hat{F}_{G'}^{|G'|}$ is an $\eps$-correct subdivision for $F$.

However, we have assumed thus far in the proof of this theorem that each Gaussian has variance at least $\frac{1}{2}$. So given samples from $F$, we can add random noise to each sample. We add Gaussian noise of variance $\frac{1}{2}$ and mean $0$, and this corresponds to convolving the original distribution $F$ by $\cN(0, \frac{1}{2})$ to obtain a new distribution $F'$. Then this distribution $F'$ has each Gaussian with variance at least $\frac{1}{2}$ and is also in nearly isotropic position - because the original mixture $F$ was in isotropic position, and convolving by $\cN(0, \frac{1}{2})$ just adds $\frac{1}{2}$ to the variance of the mixture ($var(F') = \frac{1}{2} + var(F)$).

Using the above argument, we can recover an estimate $\hat{F}_{G'}^{|G'|}$ that is an $\eps$-correct subdivision for $F'$. We can subtract $\frac{1}{2}$ from the variance of each component in $\hat{F}_{G'}^{|G'|}$, and then using Claim~\ref{claim:preconv} this resulting mixture $N$ will be an $\eps$-correct subdivision for $F$.

\end{proof}

\section{Exponential Dependence on $k$ is Inevitable}\label{sec:aexpDep}

We restate the main proposition that we prove in this section:

\begin{theorem1}{Proposition}{prop:LB}
There exists a pair $D_1,D_2$ of $1/(4k^2+2)$-standard distributions that are each mixtures of $k^2+1$ Gaussians such that $$||D_1-D_2||_1 \le 11 k e^{-k^2/24}.$$
\end{theorem1}

The following lemma will be helpful in the proof of correctness of our construction.

\begin{lemma}~\label{lemma:infC}
Let $F_k(x) = c_k \sum_{i=-\infty}^{\infty} \frac{1}{\sqrt{\pi}} e^{-(i/k)^2} \cN(i/k,1/2, x),$ where $c_k$ is a constant chosen so as to make $F_k$ a distribution.  $$||F_k(x),\cN(0,1, x)||_1 \le 10 k e^{-k^2/24}.$$
\end{lemma}

\begin{proof}
The probability density function $F_k(x)$ can be rewritten as $F_k(x) = \left(c_k C_{1/k}(x) \cN(0,1/2,x)\right) \circ \cN(0,1/2, x),$ where $C_{1/k}(x)$ denotes the infinite comb function, consisting of delta functions spaced a distance $1/k$ apart, and $\circ$ denotes convolution.  Considering the Fourier transform, we see that $$\mathcal{F}(F_k)(s) = c_k k \left(C_{k}(s) \circ \cN(0,2, s) \right) \cN(0,2, s).$$

It is now easy to see that why the lemma should be true, since the transformed comb has delta functions spaced at a distance $k$ apart, and we're convolving by a Gaussian of variance 2 (essentially yielding nonoverlapping Gaussians with centers at multiples of $k$) , and then multiplying by a Gaussian of variance 2.  The final multiplication will nearly kill off all the Gaussians except the one centered at 0, yielding a Gaussian with variance $1$ centered at the origin, whose inverse transform will yield a Gaussian of variance 1, as claimed.

To make the details rigorous, observe that the total Fourier mass of $\mathcal{F}(F_k)$ that ends up within the interval $[-k/2,k/2]$ contributed by the delta functions aside from the one at the origin, even before the final multiplication by $\cN(0,2),$ is bounded by the following:
\begin{eqnarray*}
 2 c_k k \sum_{i=1}^{\infty} \int_{(i-1/2)k}^{\infty} \cN(0,2,x) dx & = & 2 c_k k \sum_{i=1}^{\infty}\int_{(i-1/2)k/\sqrt{2}}^{\infty} \cN(0,1,x) dx \\
 & \le & 2 c_k k \sum_{i=1}^{\infty} \frac{1}{\sqrt{\pi} (i-1/2)k} e^{-(i-1/2)^2 k^2/2}\\
 & \le & 4 c_k e^{-k^2/8} \le 4 e^{-k^2/8}.
\end{eqnarray*}
Additionally, this $L_1$ fourier mass is an upper bound on the $L_2$ Fourier mass.  The total $L_1$ Fourier mass (which bounds the $L_2$ mass) outside the interval $[-k/2,k/2]$ contributed by the delta functions aside from the one at the origin is bounded by
\begin{eqnarray*} 2 c_k \int_{k/2}^{\infty} 2 \max_y(\cN(0,2,y)) \cN(0,2,x) dx & \le & 4 c_k \int_{k/2}^{\infty} \cN(0,2,x) dx \\ & \le & 4 c_k \int_{k/(2 \sqrt{2})}^{\infty} \cN(0,1,x) dx \\ & \le & 4 c_k \frac{2}{k \sqrt{\pi}} e^{-k^2/8} \le 4  \frac{2}{k \sqrt{\pi}} e^{-k^2/8}. \end{eqnarray*}


Thus we have that $$||\mathcal{F}(F_k)-c_k k \cN(0,2) \cN(0,2)||_2 = ||\mathcal{F}(F_k)-c_k k \frac{1}{2 \sqrt{2 \pi}} \cN(0,1)||_2 \le 4 e^{-k^2/8}+4 \frac{2}{k \sqrt{\pi}} e^{-k^2/8}.$$
From Plancherel's Theorem: $F_k$, the inverse transform of $\mathcal{F}(F)$, is a distribution, whose $L_2$ distance from a single Gaussian (possibly scaled) of variance 1  is at most $8 e^{-k^2/8}.$  To translate this $L_2$ distance to $L_1$ distance, note that the contributions to the $L_1$ norm from outside the interval $[-k,k]$ is bounded by $4 \int_{k}^{\infty} \cN(0,1,x) dx \le 4 \frac{1}{k \sqrt{2 \pi}} e^{-k^2/2}.$  Since the magnitude of the derivative of $F_k-c_k k \frac{1}{2 \sqrt{2 \pi}} \cN(0,1)$, is at most 2 and the value of $F_k(x)-c_k k \frac{1}{2 \sqrt{2 \pi}} \cN(0,1,x)$ is close to $0$ at the endpoints of the interval $[-k, k]$, we have $$\left(\max_{x \in [-k, k]} (|F_k(x)-c_k k \frac{1}{2 \sqrt{2 \pi}} \cN(0,1, x)|)\right)^3/(4 \cdot 3) \le \int_{-k}^k |F_k(x)-c_k k \cN(0,1, x)|^2 dx,$$ which, combined with the above bounds on the $L_2$ distance, yields $\max_{x \in [-k, k]} (|F_k(x)-c_k k \frac{1}{2 \sqrt{2 \pi}} \cN(0,1, x)|) \le (72 e^{-k^2/8})^{1/3}.$  Thus we have $$||F_k(x)-c_k k \frac{1}{2 \sqrt{2 \pi}} \cN(0,1, x)||_1 \le 4 \frac{1}{k \sqrt{2 \pi}} e^{-k^2/2} + (2k) (72 e^{-k^2/8})^{1/3}.$$ The lemma follows from the additional observation that $$||\cN(0,1)-c_k k \frac{1}{2 \sqrt{2 \pi}} \cN(0,1)||_1 = \min_{p(x)}(||c_k k \frac{1}{2 \sqrt{2 \pi}} \cN(0,1)-p(x)||_1),$$ where the minimization is taken to be over all functions that are probability density functions.
\end{proof}

\begin{prevproof}{Proposition}{prop:LB}
We will construct a pair of $1/(4k^2 +2)$-standard distributions, $D_1,D_2$, that are mixtures of $k^2+1$ Gaussians, whose statistical distance is inverse exponential in $k$.  Let \begin{eqnarray*}D_1 & = & \frac{1}{2}\cN(0,1/2) + \frac{1}{2(2k^2+1)} \sum_{i=-k^2}^{k^2} \cN(i/k,1/2), \\ D_2 & = & \frac{1}{2} c_k' \sum_{i=-k^2}^{k^2} \cN(0,1/2,i/k) \cN(i/k,1/2) +\frac{1}{2(k^2+1)}\sum_{i=-k^2}^{k^2} \cN(i/k,1/2),
\end{eqnarray*}
where $c_k'$ is a constant chosen so as to make $c_k' \sum_{i=-k^2}^{k^2} \cN(0,1/2,i/k) \cN(i/k,1/2)$ a distribution.  Clearly the pair of distributions is $1/(4k^2+2)$-standard, since all weights are at least $1/(4k^2+2),$ and the Gaussian component of $D_1$ centered at $0$ can not be paired with any component of $D_2$ without having a discrepancy in parameters of at least $1/2k.$

We now argue that $D_1,D_2$ are statistically close.  Let $D_2' = c_k' \sum_{i=-k^2}^{k^2} \cN(0,1/2,i/k) \cN(i/k,1/2).$  Note that $\int_{k}^{\infty} F_k(x) dx \le \int_{k}^{\infty} \cN(0,1/2,x) 2 \max_y(\cN(0,1/2,y)) dx \le \frac{2\sqrt{2}}{k \sqrt{\pi}}e^{-k^2} \le 2 e^{-k^2},$ and thus $||D_2'-F_k||_1 \le 8 e^{-k^2},$ and our claim follows from Lemma~\ref{lemma:infC}.
\end{prevproof}

\section{Partition Pursuit}\label{sec:apartitionpursuit}

\subsection{Paired Estimates}

We first need to ensure that if we consider two directions $r, r_{x, y}$ that are $\eps_2$-close, the parameters of a component in $P_u[F]$ cannot change too much as we vary $u$ from $r$ to $r_{x, y}$.

\begin{claim}~\label{claim:sweep}
Given a mixture of $k$ $n$-Dimensional Gaussians $F = \sum_i w_i F_i$ that is in isotropic position and is $\epsilon$-statistically learnable, for all $i$, $\|\mu_i\|, \|\Sigma_i\|_2 \leq \frac{1}{\eps}$.
\end{claim}

\begin{proof}
For all $i, j$ s.t. $\|\mu_i - \mu_j \| \leq \frac{1}{\eps}$ because if we project onto the direction $\frac{\mu_1 - \mu_2}{\|\mu_1 - \mu_2\|}$ the variance of the mixture $F$ is $1$ and is also at least $w_i w_j \| \mu_i - \mu_j\|^2 $, and this implies that $\|\mu_i - \mu_j\| \leq \frac{1}{\eps}$. Yet the convex hull of $\mu_i$ for all $i$ contains the origin and so $$\|\mu_i - \vec{0}\| \leq \max_{j} \|\mu_i - \mu_j\| \leq \frac{1}{\eps}$$ Similarly, for any $i \in [k]$, if we choose $u$ corresponding to the direction of the maximum eigenvector of $\Sigma_i$, $$1 = var(P_u(F)) \geq w_i u^T \Sigma_i u = w_i \|\Sigma_i\|_2$$ and so $\| \Sigma_i \|_2 \leq \frac{1}{\eps}$.
\end{proof}

Suppose $F$ is an $n$-dimensional GMM that is $\eps$-statistically learnable.

\begin{definition}
Let $\hat{F}^u, \hat{F}^v$ be univariate mixtures of Gaussians. Then we call components $\hat{F}^u_a, \hat{F}^v_b$ paired estimates if there is some $\pi_u, \pi_v$ and $i \in [k]$ such that $\pi_u(i) = a, \pi_v(i) = b$ and $(\hat{F}^u, \pi_u) \in \cD_{\eps_1}(P_u[F])$ and $(\hat{F}^v, \pi_v) \in \cD_{\eps_1}(P_v[F])$.
\end{definition}

\begin{claim}~\label{claim:ispaired}
Let $(\hat{F}^u, \pi_u) \in \cD_{\eps_1}(P_u[F])$ and $(\hat{F}^v, \pi_v) \in \cD_{\eps_1}(P_v[F])$, then for every component $\hat{F}^u_a$ in $\hat{F}^u$, there is some component $\hat{F}^v_b$ such that the components $\hat{F}^u_a, \hat{F}^v_b$ are paired estimates.
\end{claim}

\begin{proof}
This follows because $\pi_u$ is onto.
\end{proof}

Suppose $u, v$ are $\eps_2$-close (i.e. $\| u - v\| \leq \eps_2$), and let $\hat{F}^u_a,$ and $\hat{F}^v_b$ be paired estimates.

\begin{claim}~\label{claim:paired}
$D_p(\hat{F}^u_a, \hat{F}^v_b) \leq 2 \eps_1 + \frac{4 \eps_2}{\eps}$.
\end{claim}

\begin{proof}
$\pi_u, \pi_v$ and $i \in [k]$ such that $\pi_u(i) = a, \pi_v(i) = b$ and $(\hat{F}^u, \pi_u) \in \cD_{\eps_1}(P_u[F])$ and $(\hat{F}^v, \pi_v) \in \cD_{\eps_1}(P_v[F])$. Then
$$D_p(\hat{F}^u_a, \hat{F}^v_b)  \leq D_p(\hat{F}^u_a, P_u[F_i]) + D_p(P_u[F_i], P_v[F_i]) + D_p(P_v[F_i],  \hat{F}^v_b) \leq 2\eps_1 +  D_p(P_u[F_i], P_v[F_i])$$
And we can write:
 $$ D_p(P_u[F_i], P_v[F_i]) =  | \mu_i^T (u - v)| + |u^T \Sigma_i u - v^T \Sigma_i v|$$
 Note that $$|u^T \Sigma_i u - v^T \Sigma_i v| = | (v + u -v)^T \Sigma_i (v + u - v) - v^T \Sigma_i v|  \leq 2 \|\Sigma_i\|_2 \eps_2 + \|\Sigma_i\|_2 \eps_2^2$$
 Then this implies that $D_p(P_u[F_i], P_v[F_i]) \leq \|\mu_i\| \|u - v\| +  2 \|\Sigma_i\|_2 \eps_2 + \|\Sigma_i\|_2 \eps_2^2$ and if we apply Claim~\ref{claim:sweep}, this is at most $\frac{4 \eps_2}{\eps}$, and this implies the claim.
\end{proof}

\subsection{Reconstruction}\label{sec:arec}

\begin{figure}
\begin{center}

\myalg{alg:solve}{{\sc Solve}}{
Input: $n\geq 1$,  $\eps_2 > 0$, basis $B=(b_1,\ldots,b_n) \in \reals^{n \times n}$, means and variances $m^0,v^0,$ and $m^{ij},v^{ij} \in \reals$ for each $i,j \in [n]$.\\
Output: $\hat\mu \in \reals^n$, $\hat\Sigma \in \reals^{n\times n}$.
\begin{enumerate} \itemsep 0pt
\item Let $v^i=\frac{1}{n} \sum_{j=1}^n v^{ij}$ and $v=\frac{1}{n^2}\sum_{i=1}^n v^{ij}$.

\item For each $i \leq j \in [n]$, let $$V_{ij} = \frac{\sqrt{n}(v-v^i-v^j)}{(2\eps_2 +\sqrt{n})2\eps_2^2}-\frac{v^{ii}+v^{jj}}{(2\eps_2+\sqrt{n})4\eps_2}-\frac{v^0}{2\eps_2\sqrt{n}}
    +\frac{v^{ij}}{2\eps_2^2}.$$

\item For each $i > j \in [n]$, let $V_{ij}=V_{ji}$.  (* So $V \in \reals^{n \times n}$ *)

\item Output $$\hat\mu=\sum_{i=1}^n \frac{m^{ii}-m^0}{2\eps_2}b_i,\quad \hat\Sigma = B\left(\mathop{\arg\min}_{M \succeq 0} \|M-V\|_F\right)B^T.$$
\end{enumerate}
}
\end{center}
\caption{\small{{\sc Solve} \label{fig:solve}.}
}
\end{figure}

\begin{theorem1}{Lemma}{lemma:solve} \cite{2Gs}
Let $\eps_2,\eps_1>0$.  Suppose $|m^0-\mu\cdot r|$,$|{m}^{ij}-\mu\cdot r^{ij}|$, $|v^0-r^T \Sigma r|$,$|v^{ij}-(r^{ij})^T\Sigma r^{ij}|$ are all at most $\eps_1$.  Then {\sc Solve} outputs $\hat\mu \in \reals^n$ and $\hat\Sigma \in \reals^{n \times n}$ such that $\|\hat\mu-\mu\|<\frac{\eps_1 \sqrt{n}}{\eps_2}$, and $\|\hat{\Sigma}-\Sigma\|_F \leq \frac{6n \eps_1}{\eps_2^2}$.  Furthermore, $\hat\Sigma \succeq 0$ and $\hat\Sigma$ is symmetric.
\end{theorem1}

We will again need the notion of a Window:

\begin{definition}
Given a target additive error $\eps$, we call a Window $W = (\eps_1, \eps_2, \eps_3, \eps_4)$ well-separated if the following conditions hold:

\begin{enumerate}

\item $\max(\frac{\eps_1 \sqrt{n}}{\eps_2}, \frac{6n \eps_1}{\eps_2^2}) \leq \eps$

\item $\frac{\eps_2}{\eps} + \eps_1 << \eps_3$

\item $\frac{\eps_2}{\eps} + \eps_1 + \eps_3 << \eps_4$

\end{enumerate}
\end{definition}

\begin{definition}
We say that a univariate estimate $\hat{F} = \sum_\ell \hat{w}_\ell F_\ell$ (strongly) satisfies a Window $(\eps_1, \eps_2, \eps_3, \eps_4)$ if for all pairs $\hat{F}_i, \hat{F}_j$, the parameter distance is either at most $\eps_1$ or at least $\eps_4$. We say instead that the estimate (weakly) satisfies the Window if all pairwise parameter distances are at most $\eps_1$ or at least $\eps_3$.
\end{definition}

\begin{claim}
Given any univariate estimate $\hat{F}$ that (weakly) satisfies a Window $W= (\eps_1, \eps_2, \eps_3, \eps_4)$, the set of components of $\hat{F}$ with parameter distance at most $\eps_1$ is an equivalence class.
\end{claim}

Let $u, v$ be two directions that are $\eps_2$-close - i.e. $\|u - v \| \leq \eps_2$. Suppose that $(\hat{F}^u, \pi_u) \in \cD_{\eps_1}(P_u[F])$ and $(\hat{F}^v, \pi_v) \in \cD_{\eps_1}(P_v[F])$. Suppose further that $\hat{F}^u$ and $\hat{F}^v$ (weakly) satisfy the Window $(\eps_1, \eps_2, \eps_3, \eps_4)$. Let $\cE^u = \{\cE^u_1, \cE^u_2, ... \cE^u_{k'}\}$ and $\cE^v = \{ \cE^v_1, \cE^v_2, ... \cE^v_{k''}\}$ be the equivalence classes of components of $\hat{F}^u, \hat{F}^v$ respectively at parameter distance at most $\eps_1$.

\begin{lemma}~\label{lemma:bij}
Then $k' = k''$, and there is a permutation $\pi_{u,v}: [k'] \rightarrow [k'']$ such that $P_u[F_j]$ is mapped to the equivalence class $\cE^u_h$ by the mapping $\pi_u$ iff $P_v[F_j]$ is mapped to the equivalence $\cE^v_{\pi_{u, v}(h)}$ by the mapping $\pi_v$. Also we can construct $\pi_{u, v}$ from the estimates $\hat{F}^u, \hat{F}^v$.
\end{lemma}

\begin{proof}
To establish this claim, consider two distinct equivalence classes $\cE^u_a, \cE^u_b$, and let $\hat{F}^u_{a'}, \hat{F}^u_{b'}$  be arbitrary representative. For each component $\hat{F}^u_{a'}$ in $\hat{F}^u$, there is some component $P_u[F_i]$ in $P_u[F]$ that is mapped by $\pi_u$ to $\hat{F}^u_{a'}$. Then let $P_u[F_i], P_u[F_j]$ be mapped to $\hat{F}^u_{a'}, \hat{F}^u_{b'}$ respectively - i.e. $\pi_u(i) = a', \pi_v(j) = b'$. Then since $\hat{F}^u$ (weakly) satisfies the Window $W$, we have that $D_p(\hat{F}^u_{a'}, \hat{F}^u_{b'}) \geq \eps_3$.

Suppose that $P_v[F_i], P_v[F_j]$ are are mapped to $\hat{F}^v_{c'}, \hat{F}^v_{d'}$ and these two components are in the same equivalence class in the mixture $\hat{F}^v$. Then $D_p(\hat{F}^v_{c'}, \hat{F}^v_{d'}) \leq \eps_1$. Yet $\hat{F}^v_{a'}, \hat{F}^v_{c'}$ are paired estimates so using Claim~\ref{claim:paired}, $D_p(\hat{F}^v_{a'}, \hat{F}^v_{c'}) \leq 2 \eps_1 + \frac{4\eps_2}{\eps}$, and similarly for $\hat{F}^v_{b'}, \hat{F}^v_{d'}$. Then $D_p(\hat{F}^u_{a'}, \hat{F}^u_{b'}) \leq \eps_1 + 4 \eps_1 + \frac{8\eps_2}{\eps}$ using the triangle inequality, but this contradicts the above implication that $D_p(\hat{F}^u_{a'}, \hat{F}^u_{b'}) \geq \eps_3$ because $\eps_3 >> \eps_1 + \frac{\eps_2}{\eps}$.

This implies that every every two components in $\hat{F}^u$ that are in a different equivalence classes must be each paired to to two components in $\hat{F}^v$ that are also in a different equivalence class. The claim is symmetric w.r.t. $u, v$, so this implies that $\hat{F}^u, \hat{F}^v$ have the same number of equivalence classes.

And also consider any component $\hat{F}^u_a$. Using Claim~\ref{claim:ispaired}, there is some component $\hat{F}^v_b$ so that $\hat{F}^u_a, \hat{F}^v_b$ are paired estimates. Then using Claim~\ref{claim:paired}, $D_p(\hat{F}^u_a, \hat{F}^v_b) \leq 2 \eps_1 + \frac{4 \eps_2}{\eps}$. Yet for any component $\hat{F}^v_c$ that is not in the same equivalence class as $\hat{F}^v_b$, $$D_p(\hat{F}^u_a, \hat{F}^v_c) \geq D_p(\hat{F}^u_b, \hat{F}^v_c) - D_p(\hat{F}^u_a, \hat{F}^v_b) \geq \eps_3 - 2 \eps_1 - \frac{4 \eps_2}{\eps}$$ where the last line follows because $\hat{F}^v$ (weakly) satisfies the Window $W$. So we can construct $\pi_{u, v}$ given just $\hat{F}^u, \hat{F}^v$ because for any pair of equivalence classes $\cE^u_i, \cE^v_j$, if there is a pair of Gaussians that are paired estimates, the parameter distance between \emph{any} representative from $\cE^u_i$ to \emph{any} representative from $\cE^v_j$ must be at most $4 \eps_1 + \frac{4 \eps_2}{\eps}$. Yet if there is no such pair of Gaussians, one from each equivalence class, that are paired estimates, the parameter distance between \emph{any} representative from $\cE^u_i$ to \emph{any} representative from $\cE^v_j$ is at least $\eps_3 - 2 \eps_1 - \frac{4 \eps_2}{\eps}$ so we can distinguish these two cases because $\eps_3 >> \eps_1 + \frac{\eps_2}{\eps}$.
\end{proof}

Let $W= (\eps_1, \eps_2, \eps_3, \eps_4)$ be a well-separated window. Suppose for some root direction $r$, and $n^2$ $\eps_2$-close-by directions $r_{x, y}$ as in the {\sc Partition Pursuit Algorithm}, we run the {\sc General Univariate Algorithm} with precision $\eps_1$ and for each run we get an estimate $\hat{F}^{x,y}$ that (weakly) satisfies the Window $W$. Then suppose we run {\sc Solve} given the directions $r, r_{x,y}$ and the estimate $\hat{F}^{x,y}$.

\begin{claim}~\label{claim:nrecover}
{\sc Solve} returns an $n$-dimensional estimate $\hat{F}$ that is an $\eps$-correct subdivision of $F$.
\end{claim}

\begin{proof}
We can apply Lemma~\ref{lemma:bij} and find a partition of all equivalence classes that arise in any estimate in any direction, into sets $\cE^h = \{\cE^h_1, \cE^h_2, ... \cE^h_{n^2}\}$ with the property that for all $F_i$, there is some $h$ such that in each direction $r_{x,y}$, $F_i$ is mapped some equivalence class in $\cE^h$. Suppose in direction $r_{x, y}$, $F_i$ is mapped to the equivalence class $\cE^h_j$. Then we can take an arbitrary $\hat{F}^h_j$ in this set, and use these parameters as an estimate for the projected mean and projected variance of $P_{r_{x, y}}[F_i]$ and these estimates will be $2 \eps_1$ close in parameter distance to the actual values. So we can apply Lemma~\ref{lemma:solve}, and the component $\hat{F}_h$ of the estimate $\hat{F}$ output from {\sc Solve} that has parameter distance at most $\eps$ to $F_i$. So for every component $F_i$, there will be some estimate $\hat{F}_h$ output from {\sc Solve} that has parameter distance at most $\eps$ to $F_i$. Additionally, for every set of equivalence classes $\cE^h$, there is some component $F_i$ with the property that in each direction $r_{x,y}$, $F_i$ is mapped some equivalence class in $\cE^h$. So the mapping from a component $F_i$ to an estimate $\hat{F}_h$ that is $\eps$-close in parameter distance, will be onto. Lastly, given any partition into sets $\cE^1, \cE^2, ... \cE^{k'}$, we can choose the weight $\hat{w}_h$ to be the sum of the estimated weights in any equivalence class $\cE^h_j$ in the set, and because the {\sc General Univariate Algorithm} returns an $\eps_1$-correct subdivision, this aggregate weight $\hat{w}_h$ will be within an additive $k\eps_1$ of the actual aggregate weight of the components $F_i$ that are $\eps$-close in parameter distance to $\hat{F}_h$.
\end{proof}

\subsection{Observed Components}

\begin{definition}
Given precision $\eps_1$ (given to the {\sc General Univariate Algorithm}), we say that the number of observed pairs in the estimate $\hat{F}$ returned is the maximum value of ${k'' \choose 2}$ such that there is a subset of $k''$ components of $\hat{F}$ with the property that every pair is at parameter distance $> \eps_1$. And we will say that the number of observed components is $k''$.
\end{definition}

Suppose we are given any well-separated Window $W= (\eps_1, \eps_2, \eps_3, \eps_4)$, and an estimate $\hat{F}$ that (weakly) satisfies the Window $W$. Suppose further that the set of equivalence classes $\cE_1, \cE_2, ... \cE_{k'}$ (of components in $\hat{F}$ at parameter distance at most $\eps_1$)has $k'$ elements.

\begin{claim}~\label{claim:obs1}
The number of observed components is $k'$.
\end{claim}

So let $u, v$ be two directions that are $\eps_2$-close (i.e. $\| u - v\|\leq \eps_2$), and let $\hat{F}^u, \hat{F}^v$ be the estimates returned by the {\sc General Univariate Algorithm} when given target precision $\eps_1$, for the directions $u$, $v$ respectively. Suppose further that $\hat{F}^u$ (strongly) satisfies the Window $W$.

\begin{claim}~\label{claim:increase}
Then the estimate $\hat{F}^v$ will either (weakly) satisfy the Window $W = (\eps_1, \eps_2, \eps_3, \eps_4)$, or the number of observed pairs in $\hat{F}^v$ is strictly more than the number observed in $\hat{F}^u$.
\end{claim}

\begin{proof}
Since the estimate $\hat{F}^u$ (strongly) satisfies the Window $W = (\eps_1, \eps_2, \eps_3, \eps_4)$, it also (weakly) satisfies this Window. So we can apply Claim~\ref{claim:obs1} and this implies that there are $k'$ observed components in the estimate $\hat{F}^u$ (if there are $k'$ equivalence classes of components in $\hat{F}^u$ at parameter distance at most $\eps_1$).

Let $\hat{F}^v_c, \hat{F}^v_d$ be two arbitrary components in $\hat{F}^v$. We can apply Claim~\ref{claim:ispaired} to get two components  $\hat{F}^u_a, \hat{F}^u_b$ in $\hat{F}^u$ such that $\hat{F}^u_a$ and $\hat{F}^v_c$ are paired estimates, and similarly $\hat{F}^u_b$ and $\hat{F}^v_d$ are also paired estimates.

Suppose $\hat{F}^u_a, \hat{F}^u_b$ are not in the same equivalence class in $\hat{F}^u$. This implies that $D_p(\hat{F}^u_a, \hat{F}^u_b) \geq \eps_4$ because $\hat{F}^u$ (strongly) satisfies the Window $W$. Using Claim~\ref{claim:paired}, we get that $$D_p(\hat{F}^v_c, \hat{F}^v_d) \geq \eps_4 - 4\eps_1 -  \frac{8 \eps_2}{\eps} >> \eps_3$$ so this implies that the parameter distance $D_p(\hat{F}^v_c, \hat{F}^v_d)$ does not contribute to $\hat{F}^v$ not (weakly) satisfying $W$.

So suppose $\hat{F}^u_a, \hat{F}^u_b$ are in the same equivalence class in $\hat{F}^u$. Then using Claim~\ref{claim:paired}, we get that $$D_p(\hat{F}^v_c, \hat{F}^v_d) \leq \eps_1+ 4\eps_1 +  \frac{8 \eps_2}{\eps} << \eps_3$$ because $D(\hat{F}^u_a, \hat{F}^u_b) \leq \eps_1$.

This implies that the only way that the Window $W$ could be not (weakly) satisfied if there is some pair $\hat{F}^v_c, \hat{F}^v_d$ for which the paired estimates of each are in the same equivalence class in $\hat{F}^u$, and yet $D_p(\hat{F}^v_c, \hat{F}^v_d) > \eps_1$. So for each other equivalence class in $\hat{F}^u$ (other than the one that $\hat{F}^u_a, \hat{F}^u_b$ are in), we can select a representative component $\hat{F}^u_e$, and for each one we apply Claim~\ref{claim:ispaired} and find a corresponding component $\hat{F}^v_{e'}$. If we take this set, and $\hat{F}^v_c, \hat{F}^v_d$ this is a set of $k' + 1$ components, and using the above argument all pairs of distances are at least $\eps_3 >> \eps_1$, except for the pair $D_p(\hat{F}^v_c, \hat{F}^v_d)$ which is still $> \eps_1$, so we have $k' + 1$ observed components in $\hat{F}^v$ if $\hat{F}^v$ does not (weakly) satisfy the Window $W$.
\end{proof}

\subsection{{\sc Partition Pursuit}}

\begin{figure}
\begin{center}
\myalg{alg:partitionpursuit}{Partition Pursuit}{
Input: $\epsilon$, $k,$ sample oracle $\EX(F),$ where $F$ is a mixture of at most $k$ Gaussians, is $\eps$-statistically learnable and is in isotropic position.

Output: $\hat{F}$ which is a mixture of at most $k$ Gaussians, is an $\eps$-correct subdivision of $F$ and if $F$ has more than one component, $\hat{F}$ also has more than one component.

\begin{enumerate}

\item Set $\eps_4 = \frac{\eps^5 \delta^2}{100 n^2}$

\item Choose $r$ uniformly at random

\item Let $W = (\eps_1, \eps_2, \eps_3, \eps_4)$ be a well-separated Window

\item $<retry>:$ Choose a basis $B = (b_1, b_2, ..., b_n) \in \Re^{n \times n}$ uniformly at random among all bases for which $r = \sum_{i = 1}^n \frac{b_i}{\sqrt{n}}$

\item $\hat{F}^r \leftarrow${\sc General Univariate Algorithm}$(\eps_1, r^T \EX(F), \delta, k)$

\item While $\hat{F}^r$ does not strongly satisfy $W$

\item \qquad Shift Window $W$: $(\eps_1, \eps_2, \eps_3, \eps_4) \leftarrow (\eps_1', \eps_2', \eps_3', \eps_1)$ where $(\eps_1', \eps_2', \eps_3', \eps_1)$

\qquad is a well-separated Window

\item \qquad $\hat{F}^r \leftarrow${\sc General Univariate Algorithm}$(\eps_1, r^T \EX(F), \delta, k)$

\item end

\item Set $r^{i,j} = r + \eps_2 b_i + \eps_2 b_j$

\item For $i, j \in [n]$

\item \qquad $\hat{F}^{i,j} \leftarrow${\sc General Univariate Algorithm}$(\eps_1, (r^{i,j})^T \EX(F), \delta, k)$

\item \qquad if $\hat{F}^{i,j}$ does not weakly satisfy $W$

\item \qquad \qquad Set $r \leftarrow r^{i,j}$

\item \qquad \qquad Shift Window $W$: $(\eps_1, \eps_2, \eps_3, \eps_4) \leftarrow (\eps_1', \eps_2', \eps_3', \eps_1)$ where $(\eps_1', \eps_2', \eps_3', \eps_1)$

\qquad \qquad is a well-separated Window

\item \qquad \qquad jump to $<retry>$

\item \qquad end

\item end

\item $\hat{F} \leftarrow ${\sc Solve}$(\{\hat{F}^{i,j}\}_{i,j}, \hat{F}^r, \eps_1, \eps_2, \{r^{i,j}\}_{i,j}, r)$

\item Output $\hat{F}$

\end{enumerate}

\smallskip \noindent

}
\end{center}
\caption{\small{The Partition Pursuit Algorithm.}\label{fig:partitionpursuit}
}
\end{figure}

\begin{theorem1}{Theorem}{thm:partitionpursuit}
Given an $\eps$-statistically learnable GMM $F$ in isotropic position, the {\sc Partition Pursuit Algorithm} will recover an $\eps$-correct sub-division $\hat{F}$ and if $F$ has more than one component, $\hat{F}$ also has more than one component.
\end{theorem1}

\begin{proof}
Given an $\eps$-statistically learnable GMM $F$ in $n$ dimensions (and in isotropic position), we can project onto a direction $r$ chosen uniformly at random. Using Lemma~\ref{lemma:projsep}, we can instantiate the  {\sc Partition Pursuit Algorithm}  with a Window $W= (\eps_1, \eps_2, \eps_3, \eps_4)$ with $\eps_4 = poly(\eps, \frac{1}{n})$ so that there is at least one pair of Gaussians (with high probability) that when projected onto $r$ are at parameter distance at least $\eps_4$. So when we run the {\sc General Univariate Algorithm} after projecting onto the direction $r$, the estimate returned $\hat{F}^r$ will have at least two components in order for it to be an $\eps_1$ correct subdivision for $P_r[F]$.

If the estimate $\hat{F}^r$ returned by the {\sc General Univariate Algorithm} does not (strongly) satisfy the Window $W$, we can perform a shifting operation on the Window $W$ to obtain a new Window $W'= (\eps_1', \eps_2', \eps_3', \eps_1)$ so that $W'$ is also well-separated and the number of pairwise components observed has strictly increased. So eventually we can find a Window $W'= (\eps'_1, \eps'_2, \eps'_3, \eps'_4)$ such that the estimate $\hat{F}^r$ returned by {\sc General Univariate Algorithm} run with target precision $\eps_1$ (strongly) satisfies the Window. Because the number of observed components strictly increases each time we perform a shifting operation, the number of times that we must slide the Window is at most $k$. And each time we slide a Window, the parameters of the new Window are polynomially related to the parameters in the old Window. So the precision $\eps'_1$ of this Window will be some polynomial in the original precision $\eps_1$.

So the total number of times that we need to slide the Window is at most $k$, and this implies that the parameters we need remain polynomially lower-bounded in $\eps, \frac{1}{n}$. And when we need to perform no more slides, we have reached a root direction $r$ such that the estimate returned by the {\sc General Univariate Algorithm} is (strongly) consistent with the Window $W'$, and for each direction $r_{i, j}$ the estimate returned by the {\sc General Univariate Algorithm} (weakly) satisfies the Window $W'$ as well.

Using Claim~\ref{claim:nrecover}, this implies that the output of our algorithm is an $n$-dimensional $\eps$-correct sub-division $\hat{F}$ for $F$.
\end{proof}

\section{Clustering and Recursion}\label{sec:aclustering}

\subsection{Bi-Partitions}

Suppose the estimate $\hat{F}$ returned by the {\sc Partition Pursuit Algorithm} is an $\eps_1$-correct subdivision for $F$, but is not a good estimate in terms of statistical distance. The only way that this can happen is if there is some component of $F$ which has a co-variance matrix $\Sigma_i$ that has a very small eigenvalue. In this case, we can use this direction (i.e. the eigenvector corresponding to this eigenvalue) to cluster samples from $F$ into two sets, and proceed in each set by induction.

In this section, we give some simple claims that will be useful building blocks for deciding how to cluster. Specifically, we will need to choose some clustering scheme for samples coming from $F$, so that there is some bi-partition of the components of $F$ into $S \subset [k]$ and $[k] - S$ such that any sample generated from $F_i$ ($i \in S$) has a negligible probability of being mis-clustered.

\begin{claim}~\label{claim:bipartition}
Given a set of $k$ points $x_1, x_2, ... x_k \in \Re$ on the line and the maximum distance between any pair is $\Delta$. Then there is a bi-partition $A \subset \{x_1, x_2, ... x_k\}$, $B = \{x_1, x_2, ... x_k\} - A$ such that $D(A, B) \geq \frac{\Delta}{2^{k-2}}$ (and $A, B \neq \emptyset$) and $diam(A), diam(B) \leq \Delta(1 - 2^{k-1})$.
\end{claim}

\begin{proof}
Assume that $x_1$ is at least as small as any other value in the set, and assume that $x_2$ is at least as large as any other value in the set.
Then set $A_2 = \{x_1\}, B_2 = \{x_2\}$. Clearly $D(A_2, B_2) \geq \Delta$. Consider the point $x_3$. Either $D(A_2, x_3)$ or $D(B_2, x_e)$ must be at least $\frac{\Delta}{2}$, using the triangle inequality (because $D(A_2, B_2) \geq \Delta$). Add the point $x_3$ to the side that it is closest to, and the resulting subsets $A_3, B_3$ are at distance at least $\frac{\Delta}{2}$. Iterating this procedure yields two subset $A_k, B_k$ that are disjoint, have $D(A_k, B_k) \geq \frac{\Delta}{2^{k-2}}$ and $A_k \cup B_k = \{x_1, x_2, ... x_k\}$. Also $diam(A_k) = \max_{x_i \in A_k} x_i - x_1 \leq x_2 - D(A_k, B_k) - x_1 \leq \Delta(1 - 2^{k-1})$, and similarly for $B_k$. So take $A = A_k, B= B_k$, and this implies the claim.
\end{proof}

\begin{claim}~\label{claim:bipartition2}
Given a set of $k$ points $x_1, x_2, ... x_k \in \Re^+$ on the line that are strictly positive s.t. the maximum ratio of any two points in the set is $C > 1$. Then there is a bi-partition $A \subset \{x_1, x_2, ... x_k\}$, $B = \{x_1, x_2, ... x_k\} - A$ such that for all $x_i \in A, x_j \in B$, $$\frac{x_i}{x_j} \geq C^{\frac{1}{2^k}}$$ (and $A, B \neq \emptyset$) and also for all $x_i, x_j \in A$, $\frac{x_i}{x_j} \leq C^{1 - \frac{1}{2^k}}$ and also for all $x_i, x_j \in B$, $\frac{x_i}{x_j} \leq C^{1 - \frac{1}{2^k}}$.
\end{claim}

\begin{proof}
Let $y_1, y_2, ... y_k \in \Re$ be the logarithm of each point $x_i$ - i.e. $y_i = \log x_i$. Then the maximum distance between any two points in $y_1, y_2, ... y_k$ is $\max_{i, j} \log x_i - \log x_j = \max_{i,j} \frac{x_i}{x_j} = \log C$. So let $\Delta = \log C$ and apply Claim~\ref{claim:bipartition} to the set $y_1, y_2, .. y_k$. Then we get a bipartition $A', B'$ of $y_1, y_2, ... y_k$ and let $A, B$ be the corresponding bi-partition of $x_1, x_2, ... x_k$ - i.e. $x_i \in A$ iff $y_i \in A'$.

Then $\min_{y_i \in A', y_j \in B'} y_i - y_j \geq \frac{\Delta}{2^{k-1}}$ and $y_i - y_j = \log \frac{x_i}{x_j}$. So this  implies that $$\min_{x_i \in A, x_j \in B} \frac{x_i}{x_j} \geq 2^{\frac{\log C}{2^{k-1}}} = C^{\frac{1}{2^{k-1}}} > 1$$ Also from Claim~\ref{claim:bipartition}, we have that $\max_{y_i, y_j \in A'} y_i - y_j \geq \Delta(1 - \frac{1}{2^{k-1}})$ and $y_i - y_j = \log \frac{x_i}{x_j}$ and so $$\max_{x_i, x_j \in A} \frac{x_i}{x_j} = 2^{(1 - \frac{1}{2^{k-1}})\log C} = C^{1 - \frac{1}{2^{k-1}}}$$ and similarly for $B$.
\end{proof}

Let $\hat{F}$ be a mixture of $n$-dimensional Gaussians s.t. $\hat{F}$ is an $\eps_1$-correct sub-division for $F$. Also we assume that $F$ is in isotropic position.

\begin{claim}~\label{claim:closeto}
Let $F$ be an $\eps$-statistically learnable distribution in isotropic position. Let $(\hat{F}, \pi) \in \cD_{\eps_1}(F)$. Then for any direction $r$, $var(P_r[\hat{F}]) \geq 1 - k^2 O(\frac{\eps_1}{\eps^2})$
\end{claim}

\begin{proof}
Let $\mu = \sum_i w_i \mu_i, \hat{\mu} = \sum_i \hat{w}_i \hat{\mu}_i$. We can apply Claim~\ref{claim:sweep} to get that $\|\mu - \hat{\mu}\| \leq \eps_1 + k O(\frac{\eps_1}{\eps}) = O(\frac{k \eps_1}{\eps})$. Also using Claim~\ref{claim:sweep}, we obtain $\|\Sigma_i\|_2 \leq \frac{1}{\eps}$ and $\|\hat{\Sigma}_{\pi(i)}\|_2 \leq \frac{1}{\eps} + \eps_1$.

Consider any symmetric matrix $A$: $(u+ v)^T A (u + v)  = u^T A u + 2 v^T A u + v^T A v$. And so $$(u + v)^T A (u + v)  \leq u^T A u + 2 \|v\| \|u\| \|A\|_2 + \|v\|^2 \|A\|_2$$ And we can apply this equation using $A = rr^T$, $u = \hat{\mu}_{\pi(i)} - \hat{\mu}$ and $v = \mu_i - \mu - u$ and note that $\|A\|_2 = 1, \|u\| \leq O(\frac{1}{\eps} + \frac{k \eps_1}{\eps}) = O(\frac{1}{\eps})$ and $\|v\| \leq O(\frac{k \eps_1}{\eps})$. Then this implies that $(r^T (\mu_i - \mu))^2 \leq (r^T(\hat{\mu}_{\pi(i)} - \hat{\mu}))^2 + O(k\frac{\eps_1}{\eps^2})$. Then if we take $\Delta$ to be the discrete distribution $r^T \mu_i$ with probability $w_i$, and similarly $\hat{\Delta}$ to be the discrete distribution $r^T \hat{\mu}_i$ with probability $\hat{w}_i$, $var(\hat{\Delta}) \geq var(\Delta) - O(k^2 \frac{\eps_1}{\eps^2})$.

Also $|r^T (\Sigma_i - \hat{\Sigma}_{\pi(i)}) r| \leq \|\Sigma_i - \hat{\Sigma}_{\pi(i)} \|_F \leq \frac{1}{\eps}$.
These facts are enough to be able to apply Fact~\ref{fact:1dvar} to get that
$var(P_r[\hat{F}]) \geq var(P_r[F]) - O(k^2 \frac{\eps_1}{\eps^2})$
\end{proof}

\subsection{How to Cluster}\label{sec:ahtc}

\begin{definition}
We will call $A, B \subset \Re^n$ a clustering scheme if $A \cap B = \emptyset$
\end{definition}

\begin{definition}
For $A \subset \Re^n$, we will write $P[F_i, A]$ to denote $Pr_{x \sim F_i}[x \in A]$ - i.e. the probability that a randomly chosen sample from $F_i$ is in the set $A$.
\end{definition}

Let $(\hat{F}, \pi) \in \cD_{\eps_1}(F)$. Suppose also that $\hat{F}$ is a mixture of $k'$ components.

\begin{theorem1}{Lemma}{lemma:line}
Suppose that for some direction $v$, for all $i$: $v^T \hat{\Sigma}_i v \leq \eps_2$, for $\eps_1 \leq \frac{\sqrt{\eps_2}}{2 \eps_3}$. If there is some bi-partition $S \subset [k']$ s.t. $\forall_{i \in S, j \in [k'] - S} |v^T \hat{\mu}_i - v^T \hat{\mu}_j| \geq  \frac{3\sqrt{\eps_2}}{\eps_3}$ then there is a clustering scheme $(A, B)$ (based only on $\hat{F}$) so that for all $i \in S, j \in \pi^{-1}(i)$, $P[F_i, A] \geq 1 - \eps_3$ and for all $i \notin S, j \in \pi^{-1}(i)$, $Pr[F_i, B] \geq 1 - \eps_3$.
\end{theorem1}

\begin{proof}
For each $i$, consider the interval $I_i =  [ v^T \hat{\mu}_i - \frac{\sqrt{\eps_2}}{\eps_3}, v^T \hat{\mu}_i +\frac{\sqrt{\eps_2}}{\eps_3}  ]$. Then we will choose $A =\{x \in \Re^n | v^T x \in  \cup_{i \in S} I_i\}$ and similarly we choose $B =\{x \in \Re^n | v^T x \in  \cup_{i \notin S} I_i\}$.

We first demonstrate that $A \cap B = \emptyset$. Because of how $A, B$ are defined, this condition is equivalent to the condition that $A_i = \cup_{i \in S} I_i$ and $B_i = \cup_{i \notin S} I_i$ be disjoint. ($A_i, B_i \subset \Re$ and $A_i \cap B_i = \emptyset$). So consider any two intervals $I_i, I_j$ for $i \in S, j \notin S$. Then because $i, j$ are on different sides of the bipartition $S, [k'] - S$, we get that $ |v^T \hat{\mu}_i - v^T \hat{\mu}_j| \geq \frac{3\sqrt{\eps_2}}{\eps_3}$ so $I_i, I_j$ are in fact disjoint. This implies $A_i, B_i$ are disjoint, and this implies that $A, B$ are disjoint.

Since the standard deviation of $F_j$ in the direction of $v$ is at most $\sqrt{2\eps_2}$, points outside $I_{\pi(j)}$ are at least $1/(2\eps_3)$ standard deviations from their true mean.  Using the fact that, for a one-dimensional Gaussian random variable, the probability of being at least $s$ standard deviations from the mean is at most $2e^{-s^2/2}/(\sqrt{2\pi}s)\leq 1/s$, we get that the probability that $x$ sampled from $F_i$ is outside the range $ [ v^T \hat{\mu}_i - \frac{\sqrt{\eps_2}}{\eps_3}, v^T \hat{\mu}_i +\frac{\sqrt{\eps_2}}{\eps_3} ]$ is at most $\eps_3$. And this implies the lemma.
\end{proof}

Let $(\hat{F}, \pi) \in \cD_{\eps_1}(F)$. Suppose also that $\hat{F}$ is a mixture of $k'$ components.

\begin{theorem1}{Lemma}{lemma:line2}
Suppose that for some direction $v$ and some $i \in [k']$ such that: $v^T \hat{\Sigma}_i v \leq \eps_m$, for $\eps_m >> \eps_1$. If there is some bi-partition $S \subset [k']$ s.t. $$\frac{\min_{i \in S} v^T \hat{\Sigma}_i v}{\max(\max_{j \notin S} v^T \hat{\Sigma}_j v, \eps_m)} \geq \frac{1}{\eps_t}$$ (and $\eps_t << \eps_3^3$) then there is a clustering scheme $A, B$ such that  for all $i \in S, j \in \pi^{-1}(i)$, $P[F_i, A] \geq 1 - \eps_3$ and for all $i \notin S, j \in \pi^{-1}(i)$, $Pr[F_i, B] \geq 1 - \eps_3$.
\end{theorem1}

\begin{proof}
Let $T = [k'] - S$. Let $\hat{\sigma}_{S} = \min_{i \in S} v^T \hat{\Sigma}_i v$, $\hat{\sigma}_{T} = \max_{j \in T} v^T \hat{\Sigma}_j v$. So we are given that $\frac{\hat{\sigma}_{S}}{\max(\hat{\sigma}_{T}, \eps_m)} \geq \frac{1}{\eps_t}$.

Let $B_v = \cup_{i \in T} I_i$ where $I_i =  [ v^T \hat{\mu}_i - \frac{\sqrt{\max(\hat{\sigma}_{T}, \eps_m)}}{\eps_3} - \eps_1, v^T \hat{\mu}_i +\frac{\sqrt{\max(\hat{\sigma}_{T}, \eps_m)}}{\eps_3}  + \eps_1 ]$.

Let $F_j$ be a component in $F$ s.t. $\pi(j) = i \in T$. Then the variance of $F_j$ in the direction $v$ is at most $\hat{\sigma}_T + \eps_1 \leq 2 \max(\hat{\sigma}_{T}, \eps_m)$ where here we have used the condition that $\eps_m >> \eps_1$. So any point $x$ outside the interval $I_{\pi(j)}$ is at least $1/(2\eps_3)$ standard deviations from their true mean. Using the fact that, for a one-dimensional Gaussian random variable, the probability of being at least $s$ standard deviations from the mean is at most $2e^{-s^2/2}/(\sqrt{2\pi}s)\leq 1/s$, we get that the probability that $v^Tx$  (when $x$ is sampled from $F_j$) is outside the range $B_v$ is at most $\eps_3$.

We will we take as our clustering algorithm $B = \{x \in \Re^n | v^T x \in B_v\}$ and and $A = \Re^n - B$, then clearly $A \cap B = \emptyset$. So the above statement implies that $Pr[F_j, B] \geq 1 - \eps_3$ for any $i \notin S, j \in \pi^{-1}(i)$.

Also, for any $F_j$ with $\pi(j) \in S$, the variance when projected onto $v$ is at least $\hat{\sigma}_S - \eps_1$. So the probability that a point $v^Tx $ (where $x$ is sampled from $F_j$) is inside the range $B_v$ is at most the measure of $B_v$ times the maximum density of $P_v[F_j]$. This is at most
$$(2 \frac{\sqrt{\max(\hat{\sigma}_{T}, \eps_m)}}{\eps_3} + 2\eps_1) \times \sqrt{\frac{1}{\hat{\sigma}_S - \eps_1}} \leq 2 \sqrt{\frac{\eps_t}{\eps_3}} << \eps_3$$ where the last line follows because $\hat{\sigma}_{S} >> \eps_m >> \eps_m$ because $\eps_{\bar{S}} \geq \eps_1$ and the ratio  $\frac{\hat{\sigma}_{S}}{\max(\hat{\sigma}_{T}, \eps_m)} \geq \frac{1}{\eps_t}$ is large, and because $\eps_t << \eps_3^3$.

So we also have that $Pr[F_j, B] \leq \eps_3$ for all $i \in S, j \in \pi^{-1}(i)$. So $Pr[F_j, A] \geq 1 - \eps_3$, and this implies the lemma.
\end{proof}

\subsection{Making Progress when there is a Small Variance}

\begin{theorem1}{Lemma}{lemma:dens} \cite{2Gs}
Suppose $\|\hat{\mu}_i - \mu_i\| \leq \eps_1$, $\|\hat{\Sigma}_i - \Sigma_i\|_F \leq \eps_1$, and $| \hat{w}_i - w_i| \leq \eps_1$, if either $\|\Sigma^{-1}_i\|_2 \leq \frac{1}{2\eps_m}$ or $\|\hat{\Sigma}^{-1}_i\|_2 \leq \frac{1}{2\eps_m}$ then $$D(\hat{F}_i, F_i)^2 \leq \frac{2n \eps_1}{\eps_m} + \frac{\eps_1^2}{2 \eps_m}$$
\end{theorem1}

Now we can describe the idea behind the hierarchical clustering. Suppose the entire algorithm on $k-1$ Gaussians requires $m$ samples. Then choose $\eps_3 = \frac{\eps \delta}{m}$ so that if we take $\frac{m}{\eps}$ samples in total, then each side in the bipartition that results from clustering would get at least $m$ samples and none of the samples obtained from the oracle are mis-clustered. Then we can run the $k-1$ Gaussian algorithm on each side of the bi-partition in order to get a statistically good estimate for the original mixture of $k$ Gaussians.

Given $\eps_3$, choose $\eps_2$ s.t. $\frac{\sqrt{\eps_2}}{\eps_3} \leq \frac{1}{2^{k+1}}$. Also choose $\eps_m << \eps_2$ s.t. $(\frac{\eps_m}{\eps_2})^{\frac{1}{2^k}} << \eps_3^3$. Then choose $\eps_1 << \eps_m$.

\begin{definition}
We call the set of parameters $\eps_1 << \eps_m << \eps_2 << \eps_3$ good if
\begin{enumerate}

\item $\frac{2 n \eps_1}{\eps_m} + \frac{\eps_1^2 }{\eps_m} \leq \eps^2$

\item $k^2 \frac{\eps_1}{\eps^2} = o(1)$

\item $\eps_1 \leq \frac{\sqrt{\eps_2}}{2 \eps_3}$

\item $\frac{3\sqrt{\eps_2}}{ \eps_3} = o(2^{-k})$

\item $(\frac{\eps_m}{\eps_2})^{\frac{1}{2^k}} << \eps_3^3$

\end{enumerate}
\end{definition}

Suppose we choose a set of good parameters $\eps_1 << \eps_m << \eps_2 << \eps_3$. Then the {\sc Hierarchical Clustering Algorithm} will either return an $\eps$-close statistical estimate $\hat{F}$ for $F$ or make progress by returning a clustering scheme.

\begin{figure}
\begin{center}
\myalg{alg:hierarchical}{Hierarchical Clustering Algorithm}{
Input: $\epsilon, \eps_3$, $k,$ sample oracle $\EX(F),$ where $F$ is a mixture of at most $k$ Gaussians, is $\eps$-statistically learnable and is in isotropic position.

Output: \begin{itemize}

\item EITHER: $\hat{F}$ which is a mixture of at most $k$ Gaussians, is $\eps$-close to $F$ and if $F$

\item OR: A clustering scheme $(A, B)$ s.t. there is some partition $S, T$ of the Gaussians in $F$ ($S, T \neq \emptyset$) and for all $i \in S$, $Pr_{x \sim F_i}[x_i \in A] \geq 1 - \eps_3$, and $j \in T$, $Pr_{x \sim F_i}[x_i \in B] \geq 1 - \eps_3$

\end{itemize}

\begin{enumerate}

\item Choose a good set of parameters $\eps_1, \eps_m, \eps_2, \eps_3$ ($\eps_3$ is already fixed)

\item $\hat{F} \leftarrow${\sc Partition Pursuit}$(\eps_1, \EX(F), \delta, k)$

\item If $\hat{F}$ has only one component

\item \qquad Output $\hat{F}$

\item end

\item If no component in $\hat{F}$ has a co-variance matrix $\hat{\Sigma}_i$ with $\lambda_{min}(\hat{\Sigma}_i) \leq \eps_m$

\item \qquad Output $\hat{F}$

\item Else let $\hat{F}_i = \cN(\hat{\mu}_i, \hat{\Sigma}_i)$ and $v^T \hat{\Sigma}_i v \leq \eps_m$

\item \qquad If for all $h \neq i$, $v^T \hat{\Sigma}_h v \leq \eps_2$ and there is some $j \neq i$ s.t. $| v^T(\hat{\mu}_i - \hat{\mu}_j)| = \Omega(1)$

\item \qquad \qquad Find a $\frac{3 \sqrt{\eps_2}}{\eps_3}$-mean separated partition $S', T'$ of components in $\hat{F}$

\item \qquad \qquad Let $I_h =  [ v^T \hat{\mu}_h - \frac{\sqrt{\eps_2}}{\eps_3}, v^T \hat{\mu}_h +\frac{\sqrt{\eps_2}}{\eps_3}  ]$

\item \qquad \qquad Let $A = \{x \in \Re^n | v^T x \in \cup_{h \in S'} I_h\}$, $B = \{x \in \Re^n | v^T x \in \cup_{h \in T'} \}$

\item \qquad \qquad Output $(A, B)$

\item \qquad Else

\item \qquad \qquad Find a $(\eps_3^3, \eps_m)$-variance separated partition $S', T'$ of components in $\hat{F}$.

\item \qquad \qquad Let $T'$ be the set of smaller-variance components.

\item \qquad \qquad Let $\hat{\sigma}_{S'}, \hat{\sigma}_{T'}$ be the smallest and largest variances in $S', T'$ respectively

\item \qquad \qquad Let $I_h =  [ v^T \hat{\mu}_h - \frac{\sqrt{\max(\hat{\sigma}_{T}, \eps_m)}}{\eps_3} - \eps_1, v^T \hat{\mu}_h +\frac{\sqrt{\max(\hat{\sigma}_{T}, \eps_m)}}{\eps_3}  + \eps_1 ]$

\item \qquad \qquad Set $B = \{x \in \Re^n | v^T x \in \cup_{h \in T'}\}$, $A = \Re^n - B$.

\item \qquad \qquad Output $(A, B)$

\item \qquad end

\item end

\end{enumerate}

\smallskip \noindent

}
\end{center}
\caption{\small{The Hierarchical Clustering Algorithm.}\label{fig:hierarchical}
}
\end{figure}

\begin{theorem1}{Theorem}{thm:clustering}
The {\sc Hierarchical Clustering Algorithm} either returns an $\eps$-close statistical estimate $\hat{F}$ for $F$, or returns a clustering scheme $A, B$ such that there is some bipartition $S \subset [k]$ such that for all $i \in S, j \in \pi^{-1}(i)$, $P[F_i, A] \geq 1 - \eps_3$ and for all $i \notin S, j \in \pi^{-1}(i)$, $Pr[F_i, B] \geq 1 - \eps_3$. And also $S, [k] -S$ are both non-emtpy.
\end{theorem1}

\begin{proof}
We analyze the output of the {\sc Hierarchical Clustering Algorithm} via a case analysis:

\begin{itemize}
\item \textbf{Case 1:} Suppose that no Gaussian $\hat{F}_i$ has any variance (i.e. in any direction) that is at most $\eps_m$.
\end{itemize}

Suppose that no Gaussian $\hat{F}_i$ has any variance (i.e. in any direction) that is at most $\eps_m$. Then we can apply Lemma~\ref{lemma:dens} and because $\frac{2 n \eps_1}{\eps_m} + \frac{\eps_1^2 }{\eps_m} \leq \eps^2$, and this will imply that the estimate $\hat{F}$ is statistically close to the actual mixture $F$.

\begin{itemize}
\item \textbf{Case 2:} So suppose there is a Gaussian $\hat{F}_i$ which has a variance of at most $\eps_m$ on some direction $v$.
\end{itemize}

Then using Claim~\ref{claim:closeto}, $var(P_v[\hat{F}]) \geq 1 - O(k^2 \frac{\eps_1}{\eps^2})$. Because the parameters are good, we know that $k^2 \frac{\eps_1}{\eps^2} = o(1)$ and so $var(P_v[\hat{F}]) = \Omega(1)$. Suppose that for all $\hat{F}_j$, $D_p(P_v[\hat{F}_i], P_v[\hat{F}_j]) = o(1)$. In this case, we could apply Fact~\ref{fact:1dvar} and $\sum_j \hat{w}_j v^T\hat{\Sigma}_j v = o(1)$ and similarly $var(\hat{\Delta})$ (where $\hat{\Delta}$ is the discrete distribution on $\Re$ which takes value $v^T \hat{\mu}_j$ with probability $\hat{w}_j$) will be upper bounded by $\max_{j} (v^T(\hat{\mu}_i - \hat{\mu}_j))^2 = o(1)$. So if  for all $\hat{F}_j$, $D_p(P_v[\hat{F}_i], P_v[\hat{F}_j]) = o(1)$, we would have $var(P_v[\hat{F}]) = o(1)$ which is not possible, hence there must be some other Gaussian $\hat{F}_j$ s.t. $D_p(P_v[\hat{F}_i], P_v[\hat{F}_j]) = \Omega(1)$.

\begin{itemize}
\item \textbf{Case 2a:} Suppose that each Gaussian $\hat{F}_h$ has projected variance $v^T \hat{\Sigma}_h v \leq \eps_2$, and there is a Gaussian $\hat{F}_j$ s.t. the difference in projected means $| v^T (\hat{\mu}_i - \hat{\mu}_j)| = \Omega(1)$.
\end{itemize}

In this case, we can apply Claim~\ref{claim:bipartition} to get a bipartition $S' \subset [k']$ (let $T' = [k'] - S'$) such that $S', T' \neq \emptyset$ and such that for all $i \in S', j \in T'$, $|v^T (\hat{\mu}_i - \hat{\mu}_j)| \geq \Omega(2^{-k})$. Because the parameters are good, we have that $\frac{3\sqrt{\eps_2}}{ \eps_3} = o(2^{-k})$. Then we can apply Lemma~\ref{lemma:line} to obtain a clustering so that each successive point sampled from the oracle has probability at most $\eps_3$ of being mis-clustered, as desired. And since both $S', T'$ are non-empty, this clustering scheme returned by Lemma~\ref{lemma:line} has the property that for either side of the clustering scheme, there is some component $F_i$ in the original mixture that is mapped to that side w.h.p.

\begin{itemize}
\item \textbf{Case 2b:} Either there is some Gaussian $\hat{F}_h$ which has projected variance $v^T \hat{\Sigma}_h v \geq \eps_2$, or for all Gaussians $\hat{F}_j$ ($j \neq i$) the difference in projected means $| v^T (\hat{\mu}_i - \hat{\mu}_j)| = o(1)$.
\end{itemize}

Either case implies that there is some Gaussian $\hat{F}_h$ such that when projected onto $v$, $\hat{F}_h$ has variance at least $\eps_2$. In the first case, this is directly true. In the second case, (if we let $\hat{\Delta}$ be the discrete distribution on $\Re$ which takes value $v^T \hat{\mu}_j$ with probability $\hat{w}_j$), $var(\hat{\Delta}) = o(1)$. And using Claim~\ref{claim:closeto} and Fact~\ref{fact:1dvar}, then there must be some component $\hat{F}_h$ with $v^T \hat{F}_h v = \Omega(1) >> \eps_2$.

So let $\hat{F}_h$ be the component for which $v^T \hat{F}_h v$ is the largest (and is at least $\eps_2$).

We can do the following: Let $A_1 \subset [k'] = \{i \in [k'] | v^T \hat{\Sigma}_i v \leq \eps_m\}$. Let $B_1 = [k'] - A_1$, which is necessarily non-empty because $h \in B_1$. Then take $B_2 = \{\eps_m\} \cup \{v^T \hat{\Sigma}_i v | i \in B_1 \}$ and we can apply Claim~\ref{claim:bipartition2} to get a bi-partition $A_3, B_3$ of $B_2$ with the property that $\eps_m \in A_3$, both $A_3, B_3$ are non-empty and (choosing $C = \frac{\eps_2}{\eps_m}$ in Claim~\ref{claim:bipartition2} and $C^{\frac{1}{2^k}} >>\frac{1}{\eps_3^3}$) the ratio $\frac{\min(B_3)}{\max(A_3)} \geq \frac{1}{\eps_t} >> \frac{1}{\eps_3^3}$.

Then every projected variance $v^T \hat{\Sigma}_i v$ is in the set $A_1 \cup A_3 \cup B_3$. So we can take $A$ to be the set of indices $i \in [k']$ such that $v^T \hat{\Sigma}_i v \in A_1 \cup A_3$ and similarly we take $B$ to be the set of indices $i \in [k']$ such that $v^T \hat{\Sigma}_i v \in B_3$. Then $A, B$ is a bipartition of $[k']$.

Also $\frac{\min_{i \in B} v^T \hat{\Sigma}_i v}{ \max(\eps_m, \max_{i \in A} v^T \hat{\Sigma}_i v)} = \frac{\min(B_3)}{\max(A_3)} \geq \frac{1}{\eps_t} >> \frac{1}{\eps_3^3}$. And then we can apply Lemma~\ref{lemma:line2} and this yields a clustering so that each successive point sampled from the oracle has probability at most $\eps_3$ of being mis-clustered, as desired. Note that $i \in A$, and $h \in B$, so both of the sides of this clustering scheme are non-empty (for either side of the clustering scheme, there is some component $F_i$ in the original mixture that is mapped to that side w.h.p.).

\end{proof}

This completes the description of the {\sc Hierarchical Clustering Algorithm}.

\subsection{Recursion}\label{sec:arecursion}

\begin{figure}
\begin{center}
\myalg{alg:highdimiso}{High Dimensional Isotropic Algorithm}{
Input: $k$, $\epsilon$, sample oracle $\EX(F),$ which is a mixture of at most $k$ Gaussians which are $\eps$-statistically learnable and in isotropic position

Output: An estimate $\hat{F}$ that is $\eps$-close to $F$

\begin{enumerate}

\item Let $\eps_{k -1} = H_a(\frac{\eps}{2}, \delta, k-1)$, $\eps_3  =\frac{\eps}{2}  \eps_{k-1} \delta$

\item $OUT \leftarrow ${\sc Hierarchical Clustering Algorithm}$(\eps, \delta, \eps_3, k)$

\item If $OUT$ is an estimate $\hat{F}$

\item \qquad Output $\hat{F}$

\item Else $OUT$ is a clustering scheme $A, B$

\item \qquad Take $m \frac{\delta}{\eps_3} $ total samples $x_1, x_2, ..., x_m$ from $\EX(F)$

\item \qquad Let $X_S, X_T$ be the samples from $x_1, x_2, ..., x_m$ that are in $A, B$ respectively

\item \qquad $\hat{F}_A \leftarrow${\sc High Dimensional Anisotropic Algorithm}$(\frac{\eps}{2}, \delta, k-1, X_S)$

\item \qquad $\hat{F}_B \leftarrow${\sc High Dimensional Anisotropic Algorithm}$(\frac{\eps}{2}, \delta, k-1, X_T)$

\item \qquad Output $\hat{F} = \frac{|X_S|}{m} \hat{F}_A + \frac{|X_T|}{m} \hat{F}_B$

\item end

\end{enumerate}

\smallskip \noindent
}
\end{center}
\caption{\small{The High Dimensional Isotropic Algorithm.}\label{fig:highdimiso}
}
\end{figure}

\begin{figure}
\begin{center}
\myalg{alg:highdimani}{High Dimensional Anisotropic Algorithm}{
Input: $k$, $\epsilon$, sample oracle $\EX(F),$ which is a mixture of at most $k$ Gaussians which are $\eps$-statistically learnable

Output: An estimate $\hat{F}$ that is $\eps$-close to $F$

\begin{enumerate}

\item Let $\eps_{k } = \delta H_a(\frac{\eps}{2}, \delta, k)$

\item Take $m = O(\frac{n^4 \ln \frac{k}{\delta}}{\eps_{k}^{3}})$ samples $x_1, x_2, ..., x_m$ from $\EX(F)$

\item Compute the transformation $\hat{T}$ that places $x_1, x_2, ..., x_m$ in exactly isotropic position

\item $\hat{F} \leftarrow$ {\sc High Dimensional Isotropic Algorithm}  $(\frac{\eps}{2}, \delta, k, \hat{T}(\EX(F)))$

\item Output $\hat{F}$

\end{enumerate}

\smallskip \noindent
}
\end{center}
\caption{\small{The High Dimensional Anisotropic Algorithm.}\label{fig:highdimani}
}
\end{figure}

\section{The {\sc Isotropic Projection Lemma} for $k$ Gaussians}\label{sec:aprojlem}

\begin{theorem1}{Lemma}{lemma:projsep} [Isotropic Projection Lemma]
Given a mixture of $k$ $n$-Dimensional Gaussians $F = \sum_i w_i F_i$ that is in isotropic position and is $\epsilon$-statistically learnable, with probability $\geq 1 - \delta$ over a randomly chosen direction $u$, there is some pair of Gaussians $F_i, F_j$ s.t. $D_p(P_u[F_i], P_u[F_j]) \geq \frac{\eps^5 \delta^2}{50 n^2}$.
\end{theorem1}

\begin{proof}
Let $\eps_1 = \frac{\eps^5 \delta^2}{100 n^2}$, and $\eps_2 = \frac{4 \eps_1}{\eps}$

Let $t=2\eps_1\sqrt{n}/\delta$.

{\bf Case 1:} $\|\mu_i-\mu_j\|>t$ for some $i, j \in [k]$.  In this case, by Lemma \ref{lem:far-means}, with probability $\geq 1-\delta$, $|u \cdot (\mu_i-\mu_j)|\geq \delta t/\sqrt{n} = 2\eps_1$, as desired.

{\bf Case 2:} $\|\mu_i-\mu_j\| \leq t$ for all $i, j \in [k]$. By Lemma \ref{lem:close-means}, with probability $\geq 1-\delta$, for some $h$,
\begin{equation}\label{eq1}
u^T \Sigma_h u \leq  1-\frac{\eps\delta^2(\eps^3-t^2)}{12n^2}\leq 1-\frac{\eps\delta^2(\eps^3/2)}{12n^2} \leq 1-\eps_2
\end{equation}
If $|u \cdot (\mu_i-\mu_j)| \geq 2\eps_1$, then we are done. If not, then $|u \cdot (\mu_i-\mu_j)| \leq 2\eps_1$ for all $i, j \in [k]$. Then using Fact~\ref{fact:1dvar},
$var(\Delta) + \sum_j w_j u^T\Sigma_j u = 1$
where $\Delta$ is the discrete distribution on points in $1$-dimension which is $u^T \mu_j$ with probability $w_j$. The variance of this mixture $\Delta$ is upper bounded by $\max_{i, j} |u^T \mu_i - u^T \mu_j|^2$ which is at most $4\eps_1^2 \leq 2 \eps_1$.

So this implies
$\sum_j w_j u^T \Sigma_j u \geq 1 - 2 \eps_1$.
Then we get that $\sum_{j \neq h} w_j u^T \Sigma_j u \geq 1 - w_\ell + w_h \eps_2 - 2\eps_1$
and $w_h \eps_2 - 2\eps_1 \geq 2 \eps_1 \geq 2(\sum_{j \neq h} w_j )\eps_1$.  So, finally, we obtain
$\sum_{j \neq h} w_j u^T \Sigma_j u \geq ((\sum_{j \neq h} w_j )) (1 + 2 \eps_1)$. So there is some $j \neq h$ s.t. $u^T \Sigma_j u \geq 1 + 2 \eps_1$, and for this $j$, $D_p(P_u[F_j], P_u[F_h]) \geq  2 \eps_1$ and this yields the lemma.
\end{proof}

\begin{lemma}\label{lem:close-means}
Let $\eps,\delta>0$, $t\in (0,\eps^2)$.  Let $F$ be an $\eps$-statistically learnable distribution in isotropic position. Suppose for all $i, j \in [k]$ that $\|\mu_i-\mu_j\|\leq t$.
Then, for uniformly random $r$,
$$\P_{r \in \bS_{n-1}}\left[\min_i\{r^T \Sigma_i r\} > 1-\frac{\eps\delta^2(\eps^3-t^2)}{12n^2}\right] \leq \delta.$$
\end{lemma}

\begin{proof}
We can apply Lemma~\ref{lemma:mineig} and then apply Lemma~\ref{lemma:range}. So with probability at least $1 - \delta$, there is some $i$ s.t. $u^T \Sigma_i u \notin [1-c,1+c]$ for $c = =\frac{\delta^2 a}{4n}, a = \frac{\eps^3-t^2}{3n}$. If $u^T \Sigma_i u < 1 - c$ then we are done. If instead $u^T \Sigma_i u > 1 + c$, we can apply Fact~\ref{fact:1dvar} which implies that $\sum_j w_j u^T \Sigma_j u \leq 1$ and we have that $w_i u^T \Sigma_i u > w_i (1 + c)$. We can apply Claim~\ref{lemon} which implies that there is then some $j \neq i$ s.t. $u^T \Sigma_j u < 1 - \eps c$ which implies the lemma.
\end{proof}

\begin{claim}\label{lemon}
Suppose $w_1(1+\alpha)+w_2 (1-\beta) \leq 1$, $w_1,w_2 \geq \eps\geq 0$, $w_1+w_2=1$ and $\alpha >0$.  Then,
$\beta \geq \eps \alpha$.
\end{claim}

\begin{lemma}\label{lem:far-means}
For any $\mu_i,\mu_j \in \reals^n, \delta>0$, over uniformly random unit vectors $u$,
$$\P_{u \in \bS_{n-1}}\left[ |u \cdot \mu_i-u \cdot \mu_j| \leq \delta\|\mu_i-\mu_j\|/\sqrt{n} \right] \leq \delta.$$
\end{lemma}

\begin{lemma}\cite{2Gs}~\label{lemma:range}
Suppose $\|\Sigma_i^{-1}\|_2 \geq 1+a$, then
$\Pr_{u\in \bS_{n-1}}\left[~u^T\Sigma_iu\in [1-c,1+c]~  \right] < \delta, \quad c=\frac{\delta^2 a}{4n}.$
\end{lemma}

\begin{lemma}~\label{lemma:mineig}
Suppose the mixture $F = \sum_i w_i F_i$ is in isotropic position and is $\epsilon$-statistically learnable, and that for all $i, j \in [k]$, $\|\mu_i - \mu_j \| \leq t$. Then
$\max_i\{~\|\Sigma_i^{-1}\|_2~\} \geq 1+a, \quad a=\frac{\eps^3-t^2}{3n}.$
\end{lemma}

\begin{proof}
By Fact \ref{fact:gvar}, the squared variation distance between $F_i$ and $F_j$ is,
\begin{align*}
\eps^2 &\leq (D(F_i,F_j))^2 \leq \frac{1}{2}\sum_{i=1}^n (\lambda_i+\frac{1}{\lambda_i}-2)+(\mu_1-\mu_2)^T\Sigma_i^{-1}(\mu_1-\mu_2).
\end{align*}
Where $\lambda_1,\ldots,\lambda_n> 0$ are the eigenvalues of $\Sigma_i^{-1}\Sigma_j$. Suppose $(\mu_1-\mu_2)^T\Sigma_i^{-1}(\mu_1-\mu_2) \geq \frac{t^2}{\epsilon}$, then this implies $\|\Sigma_i^{-1}\|_2 \geq \frac{1}{\epsilon}$ because $\|\mu_1-\mu_2\| \leq t$, and we would be done in this case. If not, then from the above equation we get
$$\eps^2 \leq \frac{1}{2}\sum_{i=1}^n (\lambda_i+\frac{1}{\lambda_i}-2) + \frac{t^2}{\eps}.$$
In particular, there must be some eigenvalue $\lambda$, such that,
$\lambda +1/\lambda - 2 \geq \frac{2}{n}\ (\eps^2-\frac{t^2}{\eps} )=\frac{6a}{\eps^2}.$
Let $v$ be a unit (eigen)vector corresponding to $\lambda$, i.e., $v=\lambda\Sigma_i^{-1}\Sigma_jv$.  Then we have that $v^T\Sigma_iv =\lambda v^T\Sigma_jv$ and this yields
$$\Big(\frac{v^T\Sigma_iv}{v^T\Sigma_jv}-1\Big )+\Big (\frac{v^T\Sigma_jv}{v^T\Sigma_iv}-1\Big )=\lambda+\frac{1}{\lambda}-2 \geq \frac{6a}{\eps^2}$$
Since one of the two terms in parentheses above must be at least $3a/\eps^2$, WLOG, we can take $\frac{v^T\Sigma_iv}{v^T\Sigma_jv}\geq 1+3a/\eps^2$.  This means that the numerator or denominator is bounded from 1.  We can break this into two cases.

{\bf Case 1:} $v^T\Sigma_jv<1/(1+a)$.  This establishes the lemma immediately.

{\bf Case 2:} $v^T\Sigma_iv \geq (1+3a/\eps^2)/(1+a) = 1+(3/\eps^2-1)a/(1+a)\geq 1+(3/\eps^2-1)a/2$.  By Claim~\ref{lemon}, since $\sum_h w_h v^T \Sigma_h v \leq 1$, we have there is some $g \in [k] $, $g \neq i$ such that $$v^T \Sigma_g v \leq 1-\frac{\eps}{2}\left(\frac{3}{\eps^2}-1\right)a\leq 1-a.$$
This means that $\|\Sigma_g^{-1}\|_2 \geq 1/(1-a) \geq 1+a$.

\end{proof}

\section{Approximate Isotropic Position}

\begin{theorem} \cite{2Gs} ~\label{thm:isosingle}
Let $F_1 = \cN(\mu_1, \Sigma_1)$. Let $m = O(\frac{n^4 \ln \frac{1}{\delta}}{\eps^{4 c}})$. Then given $m$ samples from $F_1$, $x_1, x_2, ... x_m$ compute $\hat{\mu}_1 = \frac{1}{m} \sum_i x_i$ and $\hat{\Sigma}_1 = \frac{1}{m} \sum_i x_i x_i^T - \hat{\mu}_1 \hat{\mu}_1^T$. Let $\hat{F}_1 = \cN(\hat{\mu}_1, \hat{\Sigma}_1)$. Then with probability at least $1 - \delta$, $D(F_1, \hat{F}_1) \leq O(\eps^c)$.
\end{theorem}

Then, suppose we are given access to an $\eps$-statistically learnable distribution $F$ on $k$ components, which is not necessarily in isotropic position. Suppose additionally that our sample oracle gives us the labeling (corresponding to which component each sample came from) and we are given $m = O(\frac{n^4 \ln \frac{k}{\delta}}{\eps^{4 c + 1}})$ samples and labels $(x_1, \ell_1), (x_2, \ell_2), ... (x_m, \ell_2)$, where each $\ell_i \in [k]$.

Then suppose, from these samples, we construct an empirical distribution $\hat{F}$. Consider each component $\hat{F}_i$. We take $\hat{\mu}_i = \frac{1}{|\{ j | \ell_j = i\}|} \sum_{j \mbox{ s.t. } \ell_j = i } x_j$ and we similarly take $\hat{\Sigma}_i = \frac{1}{|\{ j | \ell_j = i\}|} \sum_{j \mbox{ s.t. } \ell_j = i } x_j x_j^T - \hat{\mu}_i \hat{\mu}_i^T$. And further, take $\hat{w}_i = \frac{|\{ j | \ell_j = i\}|}{m}$

\begin{corollary}~\label{cor:isomany}
For $m = O(\frac{n^4 \ln \frac{k}{\delta}}{\eps^{4 c + 1}})$, $D(F, \hat{F}), \max_i D(F_i, \hat{F}_i), \max_i |w_i - \hat{w}_i| \leq O(\eps^c)$, with probability at least $1 - \delta$.
\end{corollary}

\begin{proof}
First, consider any $i$ and let $m_i = |\{ j | \ell_j = i\}|$. Then we can apply Hoeffding's bound and $$Pr[|\frac{m_i}{m} - w_i| \geq \frac{\eps^c}{4k}] \leq 2 e^{-2m \frac{e^{2c}}{16}} \leq \frac{\delta}{4k}$$ because $m \geq \Omega(\frac{\log \frac{k}{\delta}}{\eps^{2c}})$.

So each $i$ receives at least $\Omega(\eps m) = \Omega(\frac{n^4 \ln \frac{k}{\delta}}{\eps^{4 c }})$ samples, so using Theorem~\ref{thm:isosingle}, $D(F_i, \hat{F}_i) \leq O(\eps^c)$ with probability at least $1 - \frac{\delta}{4k}$.

Then $D(F, \hat{F}) \leq \max_i D(F_i, \hat{F}_i) +  \sum_i |w_i - \hat{w}_i| = O(\eps^c)$ and the total probability of any bad event occurring is at most $\delta$ so this implies the corollary.
\end{proof}

\begin{claim}
$E_{x \sim \hat{F}} = \frac{1}{m} \sum_i x_i$ and $E_{x \sim F} [x x^T] = \frac{1}{m} \sum_i x_i x_i^T$
\end{claim}

\begin{proof}
$$E_{x \sim \hat{F}} = \sum_i \hat{w}_i \hat{\mu}_i = \sum_i \frac{|\{ j | \ell_j = i\}|}{m} \frac{1}{|\{ j | \ell_j = i\}|} \sum_{j \mbox{ s.t. } \ell_j = i } x_j = \frac{1}{m} \sum_i \sum_{j \mbox{ s.t. } \ell_j = i } x_j = \frac{1}{m} \sum_i x_i$$
And also
$$E_{x \sim \hat{F}} x x^T = \sum_i \hat{w}_i E_{x \sim \hat{F}_i} x x^T = \sum_i \hat{w}_i \hat{\Sigma}_i + \hat{\mu}_i \hat{\mu}_i^T = \sum_i \hat{w}_i \frac{1}{|\{ j | \ell_j = i\}|} \sum_{j \mbox{ s.t. } \ell_j = i } x_j x_j^T = \frac{1}{m} \sum_i x_i x_i^T$$
\end{proof}

The transformation $\hat{T}$ that puts $\hat{F}$ in isotropic position is only a function of $E_{x \sim \hat{F}}$ and $E_{x \sim F} [x x^T]$, and these quantities are computable without the labels $\ell_i$. So this implies

\begin{theorem}~\label{thm:neariso}
Given an $\eps'$-statistically learnable distribution (for $\eps' \geq \eps$) $F$, given $m = O(\frac{n^4 \ln \frac{k}{\delta}}{\eps^{5}})$ samples from $F$, one can compute a transformation $\hat{T}$ such that there is $\eps' - O(\eps)$-statistically learnable distribution $\hat{F}$ s.t. with probability at least $1 - \delta$

\begin{itemize}
\item computing an $\gamma$-close estimate for $\hat{F}$ is also an $\gamma + O(\eps)$-close statistical estimate for $F$
\item a transformation $\hat{T}$ places $\hat{F}$ in exactly isotropic position
\item $\hat{T}$ can be computed from just the sample points $x_1, x_2, ... x_m$
\item $D(F, \hat{F}) \leq O(\eps)$
\end{itemize}
\end{theorem}

\section{Basic Properties of Gaussians}~\label{sec:gaussian_properties}

In this section we state many useful basic facts about univariate Gaussian distributions that are used throughout this paper.

\begin{definition}
Given a discrete distribution on points in $1$-dimension, $\Delta$, we will define $var(\Delta)$ to be the variance of this distribution.
\end{definition}

\begin{fact}~\label{fact:1dvar}
Given a GMM of $1$-dimensional Gaussians, $F = \sum_i w_i \cN(\mu_i, \sigma_i^2)$, $$var(F) = var(\Delta) +  \sum_i w_i \sigma_i^2$$
where $\Delta$ is the discrete distribution on points in $1$-dimension corresponding to selecting each $\mu_i$ with probability $w_i$.
\end{fact}

\begin{proof}
$$E_{x \sim F}[x^2] = \sum_i w_i E_{x \sim \cN(\mu_i, \sigma_i^2)}[x^2] = \sum_{i} w_i (\sigma_i^2 + \mu_i^2) = \sum_i w_i \sigma_i^2 + \sum_i w_i \mu_i^2 =E_{x \sim \Delta}[x^2]+  \sum_i w_i \sigma_i^2 $$

Also
$E_{x \sim F}[x] = \sum_i w_i \mu_i = E_{x \sim \Delta}[x]$
and combining these equations yields:
$$var(F) = E_{x \sim F}[x^2]  - (E_{x \sim F}[x])^2 =E_{x \sim \Delta}[x^2] - (E_{x \sim \Delta}[x])^2 + \sum_i w_i \sigma_i^2 =  var(\Delta) +  \sum_i w_i \sigma_i^2$$
\end{proof}

\begin{fact}~\label{fact:gvar}
Let $F_1=\cN(\mu_1,\Sigma_1)$ and $F_2=\cN(\mu_2,\Sigma_2)$ be two $n$-dimensional Gaussian distributions. Let $\lambda_1,\ldots,\lambda_n>0$ be the eigenvalues of $\Sigma_1^{-1}\Sigma_2$. Then the variation distance between them satisfies,
$$(D(F_1,F_2))^2 \leq \sum_{i=1}^n (\lambda_i+\frac{1}{\lambda_i}-2)+(\mu_1-\mu_2)^T\Sigma_1^{-1}(\mu_1-\mu_2).$$
\end{fact}

\begin{fact}~\label{fact:shiftmean}
$$\max_{\sigma^2} \cN(0,\sigma^2,\gamma) = \frac{1}{\gamma \sqrt{2 \pi e}}.$$
\end{fact}
\begin{proof}
  It is easy to verify that $\text{argmax}_{\sigma^2}\cN(0,\sigma^2,\gamma) = \gamma^2,$ from which the fact follows.
\end{proof}

\begin{corollary}~\label{cor:diffZ}
$$\max_{\mu,\sigma^2: \mu+\sigma^2 \ge \gamma} \cN(\mu,\sigma^2,0) \le \max\left(\frac{2}{\gamma \sqrt{2 \pi e}}, \frac{1}{\sqrt{\pi \gamma}}\right).$$
\end{corollary}

\begin{proof}
Either $\mu \ge \gamma/2,$ or $\sigma^2 \ge \gamma/2$.  In the first case, using Fact~\ref{fact:shiftmean}, $$\max_{\mu \ge \gamma/2} \cN(\mu,\sigma^2, 0) =  \max_{\sigma^2} \cN(0,\sigma^2,\gamma/2) = \frac{2}{\gamma \sqrt{2 \pi e}}.$$
In the second case, we have $$\max_{x,\sigma^2 \ge \gamma/2} \cN(0,\sigma^2,x)=\cN(0,\gamma/2,0) = \frac{1}{\sqrt{\pi \gamma}}.$$
\end{proof}

\begin{lemma}~\label{lemma:tailbd}[Lemma 29 from ~\cite{2Gs}]
   Given $\sigma^2 \le 2,$  $$\int_{|x| \ge 1/\eps} |x|^i \cN(0, \sigma^2,x) dx \le O\left(\eps^{-i} e^{-\frac{1}{8\eps^2}} \right).$$
\end{lemma}

\begin{corollary}~\label{cor:tailbd}
$$\int_{|x - \mu| \ge \sigma/\eps} |x|^i \cN(\mu, \sigma^2,x) dx \le O\left(\max(|\mu|, \frac{\sigma}{\eps})^i e^{-\frac{1}{8 \eps^2}} \right).$$
\end{corollary}

\begin{proof}
Using Lemma~\ref{lemma:tailbd}, the above bound follows by a change of variables and induction. Note that the constant inside the $O()$ depends (exponentially) on $i$.
\end{proof}

\begin{lemma}~\label{lemma:same_moments}
Given $\mu,\mu',\sigma^2,\sigma'^2$ such that $|\mu|,|\mu'| < c$ and $\eps^{1/3} \le \sigma^2,\sigma'^2 \le 2$ and $|\mu-\mu'|+|\sigma^2-\sigma'^2| \le \epsilon,$ (and we also assume that $\eps c^2 = o(1)$ and $c \geq 1$) then $$|\int x^i \cN(\mu,\sigma^2,x) dx - \int x^i \cN(\mu',\sigma'^2,x) dx| \le O(c^{i + 2} \eps^{1/6} + c^i e^{-\frac{c^2}{8}})$$
\end{lemma}
\begin{proof}
Consider the interval $I = [-2c, 2c]$. Then in order to bound  $\max(|\cN(\mu,\sigma^2,x)-\cN(\mu',\sigma'^2,x)|)$ over $I$, we first bound $\max(|\cN(\mu,\sigma^2,x)-\cN(\mu',\sigma^2,x)|)$ over $I$ and next we bound $\max(|\cN(\mu',\sigma^2,x)-\cN(\mu',\sigma'^2,x)|)$ over $I$.

\begin{claim}~\label{claim:same_moments}
$$\max_{x \in I}(|\cN(\mu',\sigma^2,x)-\cN(\mu',\sigma'^2,x)|) = O(c^2\eps^{1/6})$$
\end{claim}

\begin{proof}
We prove this claim in two parts: first we bound $\max_{x \in I} |\cN(\mu,\sigma^2,x)-\cN(\mu',\sigma^2,x)|$:
\begin{eqnarray*}
\max_{x \in I} |\cN(\mu,\sigma^2,x)-\cN(\mu',\sigma^2,x)| &=& \frac{1}{\sqrt{2 \pi \sigma^2}} e^{- \frac{(x - \mu')^2}{2 \sigma^2}} | 1 - e^{- \frac{- 2 x(\mu' - \mu) + (\mu' - \mu)^2}{2 \sigma^2}}| \\
&\leq& \frac{1}{\sqrt{\sigma^2}} | 1 - e^{- \frac{- 2 x(\mu' - \mu) + (\mu' - \mu')^2}{2 \sigma^2}}| \\
& \leq &O(\frac{|x| |\mu' - \mu| + (\mu' - \mu)^2}{\sigma^3} ) = O(c \sqrt{\eps})
\end{eqnarray*}

Next, we bound the term $\max_{x \in I} (|\cN(\mu',\sigma^2,x)-\cN(\mu',\sigma'^2,x)|)$. We accomplish this by bounding both $\max_{x \in I} (\cN(\mu',\sigma^2,x)-\cN(\mu',\sigma'^2,x)$ and $\max_{x \in I}(\cN(\mu',\sigma'^2,x) - \cN(\mu',\sigma^2,x))$. Assume that $\sigma^2 \geq \sigma'^2$. Then it follows that:
$\max(\cN(\mu',\sigma'^2,x) - \cN(\mu',\sigma^2,x)) = \cN(\mu',\sigma'^2,\mu') - \cN(\mu',\sigma^2,\mu') = \frac{1}{\sqrt{2\pi}}[ \frac{1}{\sigma'} - \frac{1}{\sigma}]$
because $\cN(\mu',\sigma'^2,x)$ decreases at a faster rate than $\cN(\mu',\sigma^2,x)$ whenever $\cN(\mu',\sigma'^2,x) > \cN(\mu',\sigma^2,x)$. Also using the restriction that $\sigma'^2, \sigma^2 \geq \eps^{1/3}$ yields
$[ \frac{1}{\sigma'} - \frac{1}{\sigma}] \leq \frac{1}{\sigma} O(\frac{\eps}{\sigma^2}) \leq O(\sqrt{\eps})$.

Lastly, we bound the term $\max_{x \in I} \cN(\mu',\sigma^2,x)-\cN(\mu',\sigma'^2,x)$:

\begin{eqnarray*}
\cN(\mu',\sigma^2,x)-\cN(\mu',\sigma'^2,x) &\leq& \frac{1}{\sqrt{2 \pi \sigma'^2}} [ e^{- \frac{(x - \mu)^2}{2 \sigma^2}} - e^{- \frac{(x - \mu)^2}{2 \sigma'^2}}] \leq \frac{1}{\sqrt{2 \pi \sigma'^2}} [ e^{- \frac{(x - \mu)^2}{2 \sigma^2}} - e^{- \frac{(x - \mu)^2}{2 \sigma^2 - 2 \eps}}] \\
&\leq&\frac{1}{\sqrt{2 \pi \sigma'^2}} [ e^{- \frac{(x - \mu)^2}{2 \sigma^2}} - e^{- \frac{(x - \mu)^2}{2 \sigma^2 } (1 + O(\frac{\eps}{\sigma^2})) }] \\
&\leq& O(\frac{c^2 \eps}{\sigma^5}) = O(c^2 \eps^{1/6})
\end{eqnarray*}
Thus these bounds imply that $\max_{x \in I}(|\cN(\mu',\sigma^2,x)-\cN(\mu',\sigma'^2,x)|) = O(c^2\eps^{1/6})$
\end{proof}

So we can use the Claim~\ref{claim:same_moments} to conclude that

$$|\int_{x \in I} x^i \cN(\mu,\sigma^2,x) dx - \int_{x \in I} x^i \cN(\mu',\sigma'^2,x) dx| \leq \int_{x \in I} |x|^i |\cN(\mu,\sigma^2,x) - \cN(\mu',\sigma'^2,x)| dx = O(c^{i + 2} \eps^{1/6})$$

And we can use Corollary~\ref{cor:tailbd} to conclude that

$$|\int_{x \notin I} x^i \cN(\mu,\sigma^2,x) dx - \int_{x \notin I} x^i \cN(\mu',\sigma'^2,x) dx| \leq |\int_{x \notin I} x^i \cN(\mu,\sigma^2,x) dx| + |\int_{x \notin I} x^i \cN(\mu',\sigma'^2,x) dx| \leq O(c^i e^{-\frac{c^2}{8}})$$
\end{proof}

\begin{claim}~\label{claim:momentcoefbd}
  The $k^{th}$ raw moment of a univariate Gaussian, $M_k(\cN(\mu,\sigma^2)) = \sum_{i=0}^k c_i \mu^i \sigma^{2(k-i)},$ where $|c_i| \le (k+2)!.$
\end{claim}
\begin{proof}
Consider the moment generating function $M_X(t)= e^{t \mu+\sigma^2 t^2/2}.$  We claim that $\frac{d^i M-X(t)}{d t^i} = poly_i(\mu,\sigma,t) \cdot M_X(t),$ where $poly_i(\mu,\sigma,t)$ is a polynomial of $\mu,\sigma^2,t$, whose degree when viewed as a polynomial over $t$ is at most $i$, whose degree when viewed as a polynomial over $\mu,\sigma^2$ is at most $i$, and whose  coefficients are bounded in magnitude by $i!.$  We prove this by induction, with the base case $i=1$ being trivial.  Assuming the statement holds for some value $i \ge 1,$ we have
\begin{eqnarray*} \frac{d^i M-X(t)}{d t^i} & = & poly_i(\mu,\sigma,t) \cdot \frac{d M_X(t)}{dt}+\frac{d poly_i(\mu,\sigma,t)}{d t} M_X(t) \\
& = & \left(poly_i(\mu,\sigma,t) (2 \sigma^2 t +\mu) + \frac{d poly_i(\mu,\sigma,t)}{d t}\right) M_X(t) \end{eqnarray*}
Thus $poly_{i+1}(\mu,\sigma,t) = poly_i(\mu,\sigma,t) (2 \sigma^2 t +\mu) + \frac{d poly_i(\mu,\sigma,t)}{d t}.$   Clearly $deg_t(poly_{i+1}(\mu,\sigma,t)) = i+1,$ and the degree in terms of $\mu$ and $\sigma^2$ increases by at most one.  To get from $poly_i$ to $poly_{i+1},$ each coefficient is multiplied by 2 in the first product, and multiplied by at most $i$ in the second term because of the differentiation.  Thus if $c$ is the maximum magnitude of a coefficient of $poly_i,$ the maximum magnitude of a coefficient of $poly_{i+1}$ will be at most $(2+i)c,$ from which the claim follows.
\end{proof}

\end{document}